%% file: main.tex
\pdfoutput=1

\documentclass[english,11pt]{article}
\include{macros}

\begin{document}

\title{A Stability Principle for Learning under Non-Stationarity}\blfootnote{Author names are sorted alphabetically.}

\author{Chengpiao Huang\thanks{Department of IEOR, Columbia University. Email: \texttt{chengpiao.huang@columbia.edu}.}
	\and Kaizheng Wang\thanks{Department of IEOR and Data Science Institute, Columbia University. Email: \texttt{kaizheng.wang@columbia.edu}.}
}

\date{This version: May, 2025}

\maketitle

\begin{abstract}
We develop a versatile framework for statistical learning in non-stationary environments. In each time period, our approach applies a stability principle to select a look-back window that maximizes the utilization of historical data while keeping the cumulative bias within an acceptable range relative to the stochastic error. Our theory showcases the adaptivity of this approach to unknown non-stationarity. We prove regret bounds that are minimax optimal up to logarithmic factors when the population losses are strongly convex, or Lipschitz only. At the heart of our analysis lie two novel components: a measure of similarity between functions and a segmentation technique for dividing the non-stationary data sequence into quasi-stationary pieces. We evaluate the practical performance of our approach through real-data experiments on electricity demand prediction and hospital nurse staffing.
\end{abstract}

\input{main_intro}

\input{main_setup}

\input{main_method}

\input{main_theory}

\input{main_theory_general}

\input{main_theory_lower}

\input{main_experiments}

\input{main_discussions}

\section*{Acknowledgement}
Chengpiao Huang and Kaizheng Wang's research is supported by an NSF grant DMS-2210907, and
a startup grant and a Data Science Institute seed grant SF-181 at Columbia University.

\newpage 
\appendix

\input{appendix_proof_sketch}

\input{appendix_proof_general}

\input{appendix_proof}

\input{appendix_proof_lower}

\input{appendix_technical}

\input{appendix_experiments}

\newpage

{
\bibliographystyle{ims}
\bibliography{bib}
}

\end{document}

%% file: main_intro.tex
\section{Introduction}\label{sec-intro}

It has been widely observed in economics \citep{CHe01}, healthcare \citep{NMB19},  environmental science \citep{MBF08}, and many other fields that the underlying environment is constantly changing over time. The pervasive non-stationarity presents formidable challenges to statistical learning and data-driven decision making, as knowledge from the past may no longer be useful for the future and the learner needs to chase a moving target. In this paper, we develop a principled approach for adapting to unknown changes in the environment.

As a motivating example, consider the problem of hospital nurse staffing, where a hospital needs to decide how many nurses to schedule every week. The number of patient visits can vary significantly across time, due to the the seasonality of certain diseases (e.g., influenza), the outbreak of a new disease (e.g., COVID-19), or other factors. Consequently, past data on the numbers of patient visits may not be informative or can even be misleading for making future nurse staffing decisions. Indeed, as is demonstrated by our numerical studies on New York City emergency department visits \citep{NYC24} in \Cref{sec-experiments}, blindly using data from more than 6 months ago leads to a cost twice as high as using data from just the previous week. In contrast, our proposed approach adaptively chooses past data to make decisions, reducing the cost by 3\% to 64\% compared with different benchmarks. This highlights the importance and benefits of adapting to the non-stationary environment.

More generally, consider a canonical setup of online learning where, in each time period, a learner chooses a decision from a feasible set to minimize an unknown loss function, and observes a noisy realization of the loss through a batch of samples. The decision is made based on historical data and incurs an excess loss, which is the difference between the learner's loss and the loss of the optimal decision. The learner's overall performance is measured by the cumulative excess loss, which is an example of the dynamic regret in online learning \citep{Zin03}. 

In the presence of non-stationarity, the historical observations gathered at different time periods are not equally informative for minimizing the present objective. Most learning algorithms are designed for stationary settings, which can lead to sub-optimal outcomes if applied directly. A natural idea is to choose a look-back window $k$, and use the observations from the most recent $k$ periods to compute an empirical minimizer. Selecting a good window involves a bias-variance trade-off: increasing the window size typically reduces the stochastic error but may result in a larger bias. The optimal window is smaller during fluctuating periods and larger in stable eras. Unfortunately, such structural knowledge is often lacking in practice. 

We propose a \emph{stability principle} for automatically selecting windows tailored to the unknown local variability. At each time step, our method looks for the largest look-back window in which the cumulative bias is dominated by the stochastic error. This is carried out by iteratively expanding the window and comparing it with smaller ones. Given two windows, we compare the associated solutions through their performance on the data in the smaller window. If the performance gap is too large, then the environment seems to have undergone adverse changes within the larger window, and we choose the smaller window. Otherwise, the larger window is not significantly worse than the smaller one, and we choose the larger window to promote statistical stability. This idea can be extended to the general scenario with multiple candidate windows. A window is deemed \emph{admissible} if it passes pairwise tests against smaller ones. Our approach picks the largest admissible window to maximize the utilization of historical data while effectively managing bias. The window selection procedure can be succinctly summarized as ``expansion until proven guilty''.

\paragraph{Main contributions.} Our contributions are three-fold.
\begin{enumerate}
\item (Flexible method) We develop a versatile framework for statistical learning in dynamic environments based on the stability principle described above. It can be easily combined with learning algorithms for stationary settings, helping them adapt to distribution shifts over time.

\item (Adaptivity guarantees in common settings) We provide sharp regret bounds for our method when the population losses are strongly convex and smooth, or Lipschitz only. We also prove matching minimax lower bounds up to logarithmic factors. Our method is shown to achieve the optimal rates while being agnostic to the non-stationarity. We further evaluate its practical performance through real-data experiments on electricity demand prediction and hospital nurse staffing.

\item (A general theory of learning under non-stationarity) We derive regret bounds based on a unified characterization of non-stationarity. We propose a novel measure of similarity between functions: two functions $f,g:\Omega\to \RR$ are said to be \emph{$(\varepsilon,\delta)$-close} if for all $\btheta\in\Omega$, it holds that
\begin{align*}
& g ( \btheta ) - \inf_{ \btheta' \in \Omega } g(\btheta') \leq e^{\varepsilon}   \bigg(  f ( \btheta ) - \inf_{ \btheta' \in \Omega } f (\btheta') + \delta
\bigg) , \\[4pt]
& f ( \btheta ) - \inf_{ \btheta' \in \Omega } f(\btheta') \leq e^{\varepsilon}   \bigg(  g ( \btheta ) - \inf_{ \btheta' \in \Omega } g (\btheta') + \delta
\bigg) .
\end{align*}
The closeness relation behaves nicely under common operations, providing a powerful tool for analyzing sample average approximation with non-i.i.d.~data. We further develop a segmentation technique that partitions the whole data sequence into quasi-stationary pieces. 
\end{enumerate}

\paragraph{Related works.} We give a review of the most relevant works, which is by no means exhaustive. Existing approaches to learning under non-stationarity can be broadly divided into \emph{model-based} and \emph{model-free} ones. Model-based approaches use latent state variables to encode the underlying distributions and directly model the evolution. Examples include regime-switching and seasonality models \citep{Ham89,CWW23}, linear dynamical systems \citep{Kal60,MJS22}, Gaussian processes \citep{SUp08}, and autoregressive processes \citep{CGB23}. While they have nice interpretations, the prediction powers can be impaired by model misspecification \citep{DSa99}. Such issue may mislead models to use data from past environments that are substantially different from the present one.

In contrast, model-free approaches focus on the most recent data to ensure relevance. A popular tool is rolling window, which has seen great success in non-stationary time series \citep{FYa03}, PAC learning \citep{MMM12}, classification \citep{HKY15}, inventory management \citep{KMS23}, distribution learning \citep{MUp23}, and so on. Our approach belongs to this family, with wider applicability and better adaptivity to unknown changes. It draws inspiration from Lepskii's method for adaptive bandwidth selection in nonparametric estimation \citep{Lep91}. Both of them identify admissible solutions through pairwise tests. In Lepskii's method, each test compares the distance between two candidate solutions with a threshold determined by their estimated statistical uncertainties. However, it is not suitable when the empirical loss does not have a unique minimizer. Our approach, on the other hand, compares candidate solutions by their objective values. This is applicable to any loss function defined on an arbitrary domain that may not have a metric. Related ideas were also used by \cite{Spo09} to estimate volatilities in time series, by \cite{LWA18} to design algorithms for contextual bandits, and by \cite{MUp23} for window selection in distribution learning.

There has also been a great number of model-free approaches in the area of non-stationary online convex optimization (OCO) \citep{Haz16}. Given access to noisy gradient information, one can modify standard first-order OCO algorithms using carefully chosen restarts \citep{BGZ15,CWW19} and learning rate schedules \citep{YZJ16,CDH23,FJM23}. The updating rules are much simpler than those of rolling window methods. However, they require knowledge about certain \emph{path variation}, which is the summation of changes in loss functions or minimizers between consecutive times.
Adaptation to the unknown variation is usually achieved by online ensemble methods \citep{HSe09,ZLZ18,BWa22,BZZ22,BNR23,ZZZ24}. 
Our measure of non-stationarity gives a more refined characterization than the path variations, especially when the changes exhibit temporal heterogeneity. Moreover, our general results imply minimax optimal regret bounds with respect to path variations. The bounds show explicit and optimal dependence on the dimension of the decision space, while existing works usually treat it as a constant. On the other hand, some works on non-stationary OCO studied robust utilization of side information such as noisy forecast of the loss gradient or the data distribution before each time period \citep{HWi13,JRS15,JLZ25}. They measured the problem complexity using the sum of forecast errors, similar to the path variation. It would be interesting to extend our non-stationarity measure to that scenario. 

Full observation of the noisy loss function or its gradient is not always possible. Instead, the learner may only receive a noisy realization of the function value at the decision. This motivated recent works on OCO with bandit feedback \citep{BGZ15,CWW19,Wan25}, which reduced the problem to first-order OCO through gradient estimation. Their settings are more difficult than ours, as it is harder to detect non-stationarity from single-point observations. In contrast, our noisy observation of the whole loss function facilitates evaluation and comparison of solutions associated with different look-back windows so as to select the optimal one. Another line of research investigated dynamic pricing \citep{KZe17,ZJY23} and various bandit problems \citep{LWA18,AGO19,CLL19,WLu21,CSZ22,SKp22,FGG23,JXK23,LVX23,MRu23}, where the learner needs to strike a balance between exploration and exploitation in the presence of non-stationarity. 

\paragraph{Outline.} The rest of the paper is organized as follows. \Cref{sec-setup} describes the problem setup. \Cref{sec-method} introduces the stability principle and the methodology. \Cref{sec-theory} presents regret bounds in common settings. \Cref{sec-theory-general} develops a general theory of learning under non-stationarity. \Cref{sec-theory-lower} provides minimax lower bounds to prove the adaptivity of our method. \Cref{sec-experiments} conducts simulations and real-data experiments to test the performance of our algorithm. Finally, \Cref{sec-discussions} concludes the paper and discusses future directions.

%% file: main_setup.tex
\section{Problem Setup}\label{sec-setup}

In this section, we formally describe the problem of statistical learning in non-stationary environments and its main challenge.

\begin{problem}[Online statistical learning under non-stationarity]\label{problem-online}
Let $\cZ$ be a sample space, $\Omega$ a decision set, and $\ell: \Omega\times\cZ \to \RR$ a known loss function. At each time $n = 1, 2, ...$, the environment is represented by an unknown data distribution $\cP_n$ over $\cZ$. A learner chooses a decision $\btheta_n \in \Omega$ based on historical information to minimize the (unknown) \emph{population loss} 
\[
F_n(\btheta) = \EE_{\bz\sim\cP_n} \left[ \ell(\btheta, \bz) \right],\quad\forall \btheta\in\Omega,
\]
and collects a batch of $B\in\ZZ_+$ i.i.d.~samples $\cD_n = \{ \bz_{n, j} \}_{j=1}^{B}$ from $\cP_n$. 
Assume that $\{ \cD_n \}_{n=1}^{\infty}$ are independent. 
\end{problem}

The data $\cD_i = \{ \bz_{i, j} \}_{j=1}^{B}$ at time $i$ defines an \emph{empirical loss}
\begin{align}
f_i(\btheta) = \frac{1}{B} \sum_{j=1}^B \ell(\btheta, \bz_{i,j}),\quad\forall \btheta\in\Omega,
\label{eqn-empirical-loss}
\end{align}
which is an unbiased estimator of $F_i$. At each time $n$, given noisy observations $\{ f_i \}_{i=1}^{n-1}$, we look for $\btheta_{n}$ that will be good for minimizing the upcoming loss function $F_{n}$.
The \emph{excess risk} in period $n$ is $F_n(\btheta_n)-\inf_{\btheta_n'\in\Omega}F_n(\btheta_n')$. Our performance measure is the cumulative excess risk, also known as the \emph{dynamic regret}:
\begin{equation*}
\sum_{n=1}^N\left[F_n(\btheta_n) - \inf_{\btheta_n'\in\Omega}F_n(\btheta_n') \right].
\end{equation*}
Here the horizon $N$ may not be known \textit{a priori}.

To minimize $F_n$, it is natural to choose some \emph{look-back window} $k \in [n - 1 ]$ and approximate $F_n$ by the pre-average $f_{n, k} = \frac{1}{k} \sum_{i = n - k}^{n-1} f_i $. Let $\widehat\btheta_{n, k} $ be an approximate minimizer of $f_{n,k}$. We will select some $\widehat{k}  \in [n - 1 ]$ and output $\btheta_n = \widehat\btheta_{n, \widehat{k}  }$. 

Choosing a good window $k$ involves a bias-variance trade-off. Increasing the window size $k$ improves the concentration of the empirical loss $f_{n, k}$ around its population version $F_{n, k}=\frac{1}{k}\sum_{i=n-k}^{n-1}F_i$ and thus reduces the stochastic error. Meanwhile, the non-stationarity can drive $F_{n,k}$ away from the target loss $F_{n}$ and induce a large approximation error (bias). Achieving a low regret requires striking a balance between the deterministic bias and the stochastic error, which is a bias-variance trade-off.

\paragraph{Notation.} Let $\ZZ_+=\{1,2,...\}$ be the set of positive integers, and $\RR_+=\{x\in\RR:x\ge 0\}$ be the set of non-negative real numbers. For $n\in\ZZ_+$, define $[n]=\{1,2,...,n\}$. For $a,b\in\RR$, define $a \wedge b = \min \{ a, b \}$ and $a \vee b = \max \{ a, b \}$. For $x \in \RR$, let $x_+ = x \vee 0$. For non-negative sequences $\{a_n\}_{n=1}^{\infty}$ and $\{b_n\}_{n=1}^{\infty}$, we write $a_n=\cO(b_n)$ if there exists $C>0$ such that for all $n\in\ZZ_+$, $a_n\le Cb_n$. We write $a_n=\widetilde{\cO}(b_n)$ if $a_n=\cO(b_n)$ up to logarithmic factors; $a_n\asymp b_n$ if $a_n=\cO(b_n)$ and $b_n=\cO(a_n)$. Unless otherwise stated, $a_n\lesssim b_n$ also represents $a_n=\cO(b_n)$. 
For $\bx \in \RR^d$ and $r \geq 0$, let $B(\bx, r) = \{ \by \in \RR^d :~ \| \by - \bx \|_2 \leq r \}$ and $B_{\infty}(\bx, r) = \{ \by \in \RR^d :~ \| \by - \bx \|_{\infty} \leq r \}$. Let $\SSS^{d-1}=\{\bx\in\RR^d:\|\bx\|_2=1\}$. The diameter of a set $\Omega \subseteq \RR^d$ is defined as $\diam (\Omega) = \sup_{ \bx, \by \in \Omega } \| \bx - \by \|_2 $. The sup-norm of a function $f:\Omega\to\RR$ is defined as $\|f\|_{\infty} = \sup_{\bx\in\Omega}|f(\bx)|$. For $\alpha \in \{ 1, 2 \}$ and a random variable $z$, define $\|z\|_{\psi_{\alpha}}=\sup_{p\ge 1} \{ p^{-1/\alpha}\EE^{1/p}|z|^p \}$, where $\|\cdot\|_{\psi_1}$ is the sub-exponential norm, and $\|\cdot\|_{\psi_2}$ is the sub-gaussian norm. For a random vector $\bv$ in $\RR^d$, define $\|\bv\|_{\psi_{\alpha}}=\sup_{\bu\in\SSS^{d-1}}\|\bu^\top\bv\|_{\psi_{\alpha}}$. The notation $N(\bmu, \bSigma)$ denotes the normal distribution with mean $\bmu$ and covariance matrix $\bSigma$. The notation $\bI_d$ denotes the $d\times d$ identity matrix.

%% file: main_method.tex
\section{A Stability Principle for Adapting to Non-Stationarity}\label{sec-method}

In this section, we propose a stability principle for adaptive selection of the look-back window under unknown non-stationarity. We will first introduce a criterion for choosing between two windows based on the idea of hypothesis testing, and then extend the approach to the general case.

\subsection{Choosing between Two Windows: To Pool or Not to Pool?}\label{sec-method-binary}

To begin with, we investigate a retrospective variant of Problem \ref{problem-online}. Imagine that at time $n$, we seek to minimize the loss $F_{n} $ based on noisy realizations $\{ f_i \}_{i=1}^{n-1}$ and $f_n$ of both the past losses and the present one. 
Suppose that $\cP_1 = \cdots = \cP_{n-1}$ but there is a possible distribution shift causing $\cP_n \neq \cP_{n-1}$. Consequently, $\{ f_i \}_{i=1}^{n-1}$ are i.i.d.~but possibly poor approximations of $F_n$. We want to decide between using the most recent observation $f_n$ and pooling all the historical data $\{f_i\}_{i=1}^n$. They lead to two candidate solutions $\widetilde\btheta_1 \in \argmin_{\btheta \in \Omega} f_n (\btheta)$ and $\widetilde\btheta_0 \in \argmin_{\btheta \in \Omega}  \frac{1}{n} \sum_{i=1}^{n} f_i (\btheta)  $, respectively. 

Our idea is to detect harmful distribution shift between $\cP_{n-1}$ and $\cP_n$, get an indicator $\test \in \{ 0, 1 \}$, and then output $\widetilde\btheta_{\test}$. We make the following observations:
\begin{enumerate}
\item If $\cP_{n-1} = \cP_n$, then $\widetilde\btheta_0$ tends to be better than $\widetilde\btheta_1$ due to its statistical stability, i.e.~$F_{n} ( \widetilde\btheta_0 ) - F_{n} ( \widetilde\btheta_1 ) \le 0$;
\item If there is a harmful distribution shift between $\cP_{n-1}$ and $\cP_n$, then $\widetilde\btheta_{0}$ will be much worse than $\widetilde\btheta_1$, i.e.~$F_{n} ( \widetilde\btheta_0 ) - F_{n} ( \widetilde\btheta_1 )$ is large.
\end{enumerate}
A faithful test should be likely to return $\test = 0$ in the first case, and $\test = 1$ in the second case. Both cases concern the performance gap $F_{n} ( \widetilde\btheta_0 ) - F_{n} ( \widetilde\btheta_1 )$. We propose to estimate it by $f_{n} ( \widetilde\btheta_0 ) - f_{n} ( \widetilde\btheta_1 )$ and compare it with some threshold $\tau > 0$. The resulting test is
\begin{align}
\test = \begin{cases}
0  , &\quad \text{if } f_{n} ( \widetilde\btheta_0 ) - f_{n} ( \widetilde\btheta_1 ) \le \tau \\[4pt]
1  , &\quad \text{if } f_{n} ( \widetilde\btheta_0 ) - f_{n} ( \widetilde\btheta_1 ) > \tau 
\end{cases}.
\label{eqn-test-binary}
\end{align}
To set the threshold $\tau$, we need some estimates on the statistical uncertainty of the test statistic $f_{n} ( \widetilde\btheta_0 ) - f_{n} ( \widetilde\btheta_1 )$ in the absence of distribution shift. As we will demonstrate in \Cref{sec-theory}, these are available in many common scenarios. 

In words, our principle can be summarized as follows:
\begin{center}
\emph{We prefer a statistically more stable solution unless it appears significantly worse.}
\end{center}

\subsection{Choosing from Multiple Windows}\label{sec-method-general}

We now develop a general framework for window selection. Recall that for any $n \geq 2$, each look-back window $k \in [n - 1]$ is associated with a loss function $f_{n,  k } =  \frac{1}{k} \sum_{i = n - k }^{n-1} f_i $ and its minimizer $\widehat\btheta_{n, k }$. Following the idea in \eqref{eqn-test-binary}, we choose positive thresholds $\{ \tau(n , i) \}_{i=1}^{n-1}$ and construct a test
\begin{align}
\test_{i, k} = \begin{cases}
0 , &\quad  \text{if } f_{n, i} ( \widehat\btheta_{n, k } ) - f_{n, i} ( \widehat\btheta_{n, i } ) \le \tau (n, i) \\[4pt]
1  , &\quad \text{if } f_{n, i} ( \widehat\btheta_{n, k } ) - f_{n, i} ( \widehat\btheta_{n, i } ) > \tau (n, i)
\end{cases}
\end{align}
for every pair of windows $i\le k$. If $\{ \cP_i \}_{i=n-k}^{n-1}$ are close and the thresholds are suitably chosen, then $\test_{1, k} = \cdots = \test_{k, k} = 0$ with high probability. Such test results give us the green light to pool $\{ \cD_i \}_{i=n-k}^{n-1}$. When $\test_{i, k} = 1$ for some $i < k$, a harmful distribution shift seems to have occurred in the last $k$ time periods, and the positive test result raises a red flag.

The pairwise tests lead to a notion of admissibility: a window size $k \in [n-1]$ is said to be \emph{admissible} if $\test_{i, k} = 0$ for all $i \in [k]$. Our stability principle suggests choosing the largest admissible window. In doing so, we maximize the utilization of historical data while keeping the cumulative bias within an acceptable range relative to the stochastic error. 
We name the procedure as \underline{S}tability-based \underline{A}daptive \underline{W}indow \underline{S}election, or SAWS for short. A formal description is given by \Cref{alg-offline-simple}.

\begin{algorithm}[t]
	{\bf Input:} Samples $\{ \cD_i \}_{i=1}^{n-1}$, non-increasing sequence of thresholds $\{ \tau ( n, k) \}_{k=1}^{n-1} \subseteq [0 , \infty) $.\\
	{\bf For $k = 1,\cdots, n-1$:}\\
	\hspace*{.6cm} Compute a minimizer $\widehat{\btheta}_{ n, k }$ of $f_{n, k} = \frac{1}{k} \sum_{i = n - k }^{n-1} f_i$, where $f_i$ is defined in \eqref{eqn-empirical-loss}.\\
	\hspace*{.6cm} Let $\test_k = 0$ if $f_{n, i } ( \widehat\btheta_{n, k}  ) - f_{n, i } ( \widehat\btheta_{n, i } ) \leq \tau ( n , i ) $ holds for all $i \in [k]$, and $\test_k = 1$ otherwise.\\
	Let $\widehat{k} =  \max \{ k \in [m]  :~ \test_k = 0 \}$.\\
	{\bf Output:} $\btheta_{n} = \widehat{\btheta}_{n, \widehat{k}  }$.
	\caption{Stability-based Adaptive Window Selection (Subroutine)}
	\label{alg-offline-simple}
\end{algorithm}

\begin{algorithm}[t]
	{\bf Input:} Thresholds $\{ \tau ( n , k) \}_{ n \in \ZZ_+, k \in [n-1] } \subseteq [0 , \infty) $.\\
	Choose any $\btheta_1\in\Omega$.\\
	{\bf For $n = 2,\cdots, N$:}\\
	\hspace*{.6cm} Run \Cref{alg-offline-simple} with samples $\{ \cD_i \}_{i=1}^{n-1}$ and thresholds $\{ \tau (n , k) \}_{k=1}^{n-1} $ to obtain $\btheta_n$.\\
	{\bf Output:} $\{ \btheta_n \}_{n = 1}^N$.
	\caption{Stability-based Adaptive Window Selection (Online Version)}
	\label{alg-online-simple}
\end{algorithm}

To tackle online learning under non-stationarity (Problem \ref{problem-online}), we apply \Cref{alg-online-simple}, which runs \Cref{alg-offline-simple} as a subroutine in each period $n\in\ZZ_+$. In \Cref{sec-theory} we will design $\tau(n,k)$ to get sharp theoretical guarantees simultaneously for all horizons $N \in \ZZ_+$. Roughly speaking, when the population losses $\{ F_n \}_{n=1}^N$ are strongly convex, we choose $\tau (n, k) \asymp \frac{d \log n}{B k}$ with $d$ being the dimension of the decision space $\Omega$; when the population losses $\{ F_n \}_{n=1}^N$ are only Lipschitz, we choose $\tau (n, k) \asymp \sqrt{ \frac{d \log n}{B k} }$.

\subsection{Efficiency Improvements}\label{sec-method-efficient}

\begin{algorithm}[t]
	{\bf Input:} Samples $\{ \cD_i \}_{i=n_0}^{n-1}$, non-increasing sequence of thresholds $\{ \tau ( n , k) \}_{k=1}^{n-1} \subseteq [0 , \infty) $, window sizes $\{ k_s \}_{s = 1}^{m} \subseteq [n-1]$ that satisfy $1 = k_1 < \cdots < k_m = n-n_0$.\\
	{\bf For $s = 1,\cdots, m$:}\\
	\hspace*{.6cm} Compute a minimizer $\widehat{\btheta}_{ n, k_s }$ of $f_{n, k_s} = \frac{1}{k_s} \sum_{i = n - k_s }^{n-1} f_i$, where $f_i$ is defined in \eqref{eqn-empirical-loss}.\\
	\hspace*{.6cm} Let $\test_s = 0$ if $f_{n, k_i } ( \widehat\btheta_{n, k_s}  ) - f_{n, k_i } ( \widehat\btheta_{n, k_i } ) \leq \tau ( n , k_i ) $ holds for all $i \in [s]$, and $\test_s = 1$ otherwise.\\
	Let $\widehat{s} =  \max \{ s \in [m]  :~ \test_s = 0 \}$.\\
	{\bf Output:} $\btheta_{n} = \widehat{\btheta}_{n, k_{\widehat{s}}   }$ and $\widehat{k} = k_{\widehat{s}}$.
	\caption{Stability-based Adaptive Window Selection (General Subroutine)}
	\label{alg-offline}
\end{algorithm}

\begin{algorithm}[t]
	{\bf Input:} Thresholds $\{ \tau ( n , k) \}_{ n \in \ZZ_+, k \in [n-1] } \subseteq [0 , \infty) $.\\
	Let $K_1 = 0$ and choose any $\btheta_1\in\Omega$.\\
	{\bf For $n = 2,\cdots, N$:}\\
	\hspace*{.6cm} Let $m = \lceil \log_2 ( K_{n-1} + 1  ) \rceil + 1$, $k_s = 2^{s - 1}$ for $s \in [m - 1]$, and $k_m = K_{n-1} + 1$.\\	
	\hspace*{.6cm} Run \Cref{alg-offline} with inputs $\{ \cD_i \}_{i=n-k_m}^{n-1}$, $\{ \tau (n , k) \}_{k=1}^{n-1} $ and $\{ k_s \}_{s=1}^m$ to obtain $\btheta_{n}$ and $\widehat{k}$.\\
	\hspace*{.6cm} Let $K_n = \widehat{k}$.\\
	{\bf Output:} $\{ \btheta_n \}_{n = 1}^N$.
	\caption{Stability-based Adaptive Window Selection (Online Version with Improved Efficiency)}
	\label{alg-online}
\end{algorithm}

In the worst case, \Cref{alg-offline-simple} solves $\cO(n)$ empirical risk minimization problems at time $n\in\ZZ_+$. Running \Cref{alg-online-simple} up to time $N$ requires solving $\cO(N^2)$ empirical risk minimization problems and storing $\cO(NB)$ samples.
To improve computational and memory efficiency, we  further develop more efficient versions of Algorithms \ref{alg-offline-simple} and \ref{alg-online-simple}, given by Algorithms \ref{alg-offline} and \ref{alg-online}, respectively. They incorporate the following two efficiency improvements.

First, \Cref{alg-offline} allows for a general collection of candidate windows $\{ k_s \}_{s=1}^m$ that is not necessarily the whole set $\{ 1, 2, \cdots, n - 1 \}$. In particular, we will use the geometric sequence $k_s=2^{s-1}$, so that at most $\cO(\log n)$ empirical risk minimization problems are solved at each time $n\in\ZZ_+$. Improving the efficiency of a search procedure by adopting a geometric candidate sequence is a standard technique in learning under non-stationarity \citep{HSe09} and beyond.

Second, \Cref{alg-online}, which runs \Cref{alg-offline} as a subroutine, employs a caching mechanism that discards irrelevant past data upon detection of non-stationarity. Specifically, at each time $n\in\ZZ_+$, \Cref{alg-online} applies \Cref{alg-offline} to obtain a window $K_n = \widehat{k}$. As the window $K_n$ indicates that a significant distribution shift has been detected at time $n-K_n$, \Cref{alg-online} discards all past data before time $n-K_n$. Thus, in the next period $n+1$, it suffices to consider look-back windows with lengths at most $K_n+1$, which leads to the candidate window sequence $\{k_s\}_{s=1}^m$ with $k_s=2^{s-1}\ \forall s\in[m-1]$ and $k_m = K_n+1$. Similar ideas are also used in multiple change-point detection \citep{NHZ16}.

Finally, we emphasize that our algorithms do not require any prior information on the non-stationarity of the underlying environment.

%% file: main_theory.tex
\section{Regret Analysis in Common Settings}\label{sec-theory}

In this section, we will provide theoretical guarantees for SAWS (\Cref{alg-online}) in two scenarios where the population losses are strongly convex and smooth, or Lipschitz only. Throughout this section, we make the following standard assumption.

\begin{assumption}[Regularity of domain]\label{assumption-bounded-domain}
The decision set $\Omega$ is a closed convex subset of $\RR^d$, and $\diam(\Omega)=M<\infty$ is a constant. 
\end{assumption}

\subsection{Strongly Convex Population Losses}\label{sec-theory-strong-cvx}

Our first study concerns the case where each population loss $F_n$ is strongly convex and thus has a unique minimizer. To set the stage, we make the following standard assumptions.

\begin{assumption}[Strong convexity and smoothness]\label{assumption-strongly-convex}
The loss function $\ell :~ \Omega \times \cZ\to \RR $ is convex and continuously differentiable with respect to its first argument. There exist constants $0<\rho\le L<\infty$ such that for every $n\in \ZZ_+$, $F_n$ is $\rho$-strongly convex and $L$-smooth: 
\begin{align*}
& F_n(\btheta') \ge F_n(\btheta) + \left\langle \nabla F_n(\btheta),~\btheta'-\btheta \right\rangle + \frac{\rho}{2} \left\| \btheta'- \btheta \right\|_2^2, \\[4pt]
& \left\|\nabla F_n(\btheta) - \nabla F_n(\btheta') \right\|_2 \le L \left\| \btheta - \btheta' \right\|_2,\quad  \forall \btheta,\btheta'\in\Omega.
\end{align*}
Moreover, for each $n\in\ZZ_+$, $F_n$ attains its minimum at an interior point $\btheta^*_n$ of $\Omega$.
\end{assumption}

\begin{assumption}[Concentration]\label{assumption-concentration-strong-cvx}
There exist constants $\sigma, \lambda > 0$ such that for all $n\in\ZZ_+$ and $\bz_n \sim \cP_n$,
\begin{align*}
&\sup_{  \btheta \in \Omega } \|\nabla \ell (\btheta , \bz_n) - \nabla F_n ( \btheta ) \|_{\psi_1}\le \sigma, \\[4pt]
&
\EE
\left[ 
\sup_{ \substack{ \btheta,\btheta' \in \Omega \\ \btheta \neq \btheta' } }
\frac{\| \nabla \ell( \btheta , \bz_n) - \nabla \ell (\btheta',\bz_n) \|_2}{\| \btheta - \btheta' \|_2}
\right]
\le 
\lambda^2 d
.  
\end{align*} 
Here the gradient of $\ell$ is taken with respect to the first argument $\btheta$.
\end{assumption}

Assumption \ref{assumption-strongly-convex} states that $F_n$ is strongly convex and smooth, and attains its minimum at some interior point of the domain. The interior minimizer assumption is common in the literature of non-stationary stochastic optimization \citep{BGZ15,Wan25}. Assumption \ref{assumption-concentration-strong-cvx} states that the empirical losses have sub-exponential tails and Lipschitz continuous gradients. Below we present canonical examples that satisfy Assumptions \ref{assumption-strongly-convex} and \ref{assumption-concentration-strong-cvx}. In these examples, $\Omega = B ( \bm{0} , M/2 )$ is a ball with diameter $M$, and $\sigma_0>0$ is a constant. We defer their verifications to \Cref{sec-eg-strongly-convex}.

\begin{example}[Gaussian mean estimation]\label{eg-Gaussian-mean}
Suppose $\cZ = \RR^d$, $\ell(\btheta,\bz) = \frac{1}{2} \| \btheta - \bz \|_2^2$, and $ \cP_n= N (\btheta_n^*,\sigma_0^2\bI_d)$ for some $\btheta_n^*$ with $\|\btheta_n^*\|_2 < M/2$. Then, Assumptions \ref{assumption-strongly-convex} and \ref{assumption-concentration-strong-cvx} hold with $\rho=L=\lambda=1$ and $\sigma=c\sigma_0$ for some universal constant $c \geq 1/2$.
\end{example}

\begin{example}[Linear regression]\label{eg-linear-regression}
Each sample $\bz_n\sim\cP_n$ takes the form $\bz_n=(\bx_n,y_n)\in\RR^d\times\RR$, where the covariate vector $\bx_n$ and the response $y_n$ satisfy $\EE(y_n|\bx_n)=\bx_n^\top\btheta_n^*$. Define the squared loss $\ell(\btheta,\bz)=\frac{1}{2}(y-\bx^\top\btheta)^2$ and the error term $\varepsilon_n=y_n-\bx_n^\top\btheta_n^*$. Suppose that $ \| \btheta_n^* \|_2 < M / 2$, $\|\bx_n\|_{\psi_2}\le\sigma_0$, $\|\varepsilon_n\|_{\psi_2}\le\sigma_0$, and $\EE ( \bx_n\bx_n^\top ) \succeq \gamma\sigma_0^2\bI_d$ for some constant $\gamma\in(0,1]$. 
Then, Assumptions \ref{assumption-strongly-convex} and \ref{assumption-concentration-strong-cvx} hold with $\sigma \asymp (M+1)\sigma_0^2$, $\lambda \asymp \sigma_0$, $\rho \asymp \gamma\sigma_0^2$, and $L \asymp \sigma_0^2$.
\end{example}

\begin{example}[Logistic regression]\label{eg-logistic-regression}
Each sample $\bz_n\sim\cP_n$ takes the form $\bz_n=(\bx_n,y_n)\in\RR^d\times\{0,1\}$, where the covariate vector $\bx_n$ and the binary label $y_n$ satisfy $\PP(y_n=1|\bx_n)=1/[1+\exp(-\bx_n^\top\btheta_n^*)]$. Define the logistic loss $\ell(\btheta,\bz)=\log [ 1+\exp(\bx^\top\btheta) ] -y\bx^\top\btheta$. Suppose that $ \| \btheta_n^* \|_2 < M / 2$, $\|\bx_n\|_{\psi_1}\le\sigma_0$, and $\EE(\bx_n\bx_n^\top)\succeq \gamma\sigma_0^2\bI_d$ for some constant $\gamma \in (0, 1]$. Then, Assumptions \ref{assumption-strongly-convex} and \ref{assumption-concentration-strong-cvx} hold with $\sigma \asymp \sigma_0$, $\lambda\asymp\sigma_0$, $L\asymp\sigma_0^2$, and $\rho=c \gamma\sigma_0^2$ for some $c>0$ determined by $M$, $\gamma$ and $\sigma_0$.
\end{example}

\begin{example}[Robust linear regression]\label{eg-robust-regression}
Each sample $\bz_n\sim\cP_n$ takes the form $\bz_n=(\bx_n,y_n)\in\RR^d\times\RR$, where the covariate vector $\bx_n$ and the response $y_n$ satisfy $y_n=\bx_n^\top \bbeta_n^* + \varepsilon_n$. Suppose $M\ge 1$, $\|\bbeta_n^*\|_2 \le M/4$, $\|\bx_n\|_{\psi_2}\le\sigma_0$, and $\EE ( \bx_n\bx_n^\top ) \succeq \gamma\sigma_0^2\bI_d$ for some constant $\gamma\in ( 0 , 1]$. Assume that the noise $\varepsilon_n$ follows the Huber contamination model \citep{Hub64}: $\varepsilon_n\sim(1-p)\cQ_n^* + p\cQ_n$ for some $p\in(0,1)$, where $\cQ_n^*$ is symmetric with respect to 0 and has a sub-gaussian norm bounded by $\sigma_0$, while $\cQ_n$ can be arbitrary and may have a nonzero mean and a heavy tail. For $u>0$, define the Huber loss
\[
h_{u}(t) =
\begin{cases}
\frac{1}{2}t^2, &\quad\text{if } |t| \le u \\[4pt]
u\left( |t| - \frac{1}{2}u \right), &\quad\text{otherwise}
\end{cases}.
\] 
Choose $\ell(\btheta,\bz) = h_{u}(y - \bx^\top\btheta)$ with $u = cM\sigma_0$ for some constant $c>0$. Then, for $(p^{-1}-1)\gamma$ sufficiently large, Assumptions \ref{assumption-strongly-convex} and \ref{assumption-concentration-strong-cvx} hold with $\sigma \asymp M\sigma_0^2$, $\lambda \asymp \sigma_0$, $\rho \asymp \gamma\sigma_0^2$, and $L \asymp \sigma_0^2$.
\end{example}

We emphasize that only the population loss $F_n$, but not the empirical loss $f_n$, is assumed to be strongly convex. This is much weaker than assuming that $f_n$ is strongly convex or exp-concave, as is commonly done in the literature \citep{HSe09,MSJ16,BWa22}. For example, $f_n$ is not strongly convex in Examples \ref{eg-linear-regression} and \ref{eg-logistic-regression} when the batch size $B$ in each time period is smaller than the dimension $d$. In \Cref{eg-robust-regression}, $f_n$ is neither strongly convex nor exp-concave due to the linearity of $h_{u}$ in $(-\infty, -u) \cup (u, \infty)$.

\begin{figure}[h]
\centering
\includegraphics[scale=0.6]{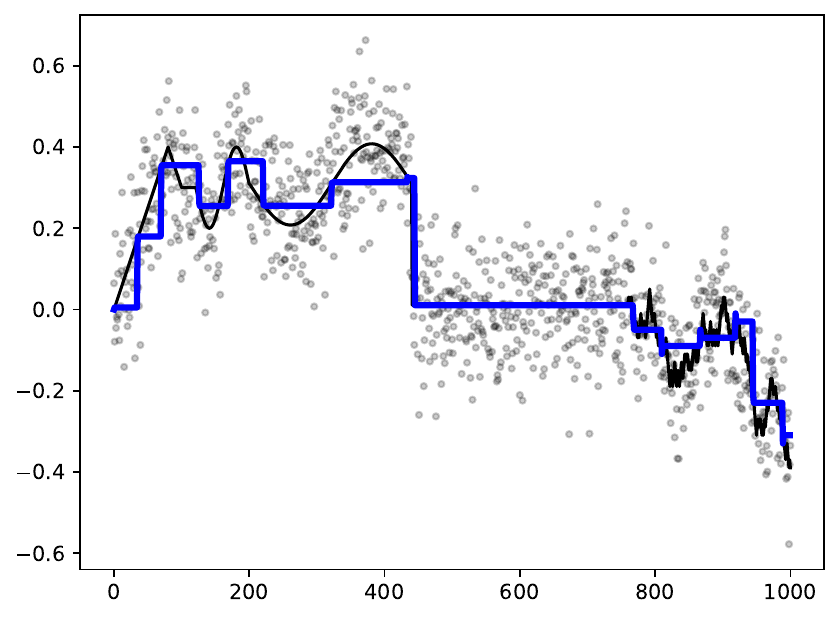}
\caption{Visualization of segmentation for \Cref{eg-Gaussian-mean}. Horizontal axis: time $n$. Vertical axis: values of $\theta_n^* \in \RR$. Black curve: trajectory of $\{\theta_n^*\}_{n=1}^N$. Gray dots: samples from $N(\theta_n^*,0.01)$. Blue curve: quasi-stationary segments of $\{\theta_n^*\}_{n=1}^N$. The sequence $\{\theta_n^*\}_{n=1}^N$ is approximated by multiple constant segments, and within each segment $\theta_n^*$ only has small variations.} \label{fig-segmentation}
\end{figure}

The regret bound of our algorithm will depend on the non-stationarity of the environment. We propose to measure it by decomposing the minimizer sequence $\{ \btheta_n^* \}_{n=1}^{N}$ into \emph{quasi-stationary segments}. Within each segment, the environment has small variations and can be treated as stationary. In this way, the non-stationarity is reflected by the number of such segments: a stationary environment is just one segment itself, while a heavily fluctuating environment needs to be divided into a large number of short segments. \Cref{fig-segmentation} provides a visualization of segmentation.

To motivate our segmentation criterion, consider the mean estimation problem in \Cref{eg-Gaussian-mean} with $d = 1$ and $\sigma_0 = 1$, and let $\Omega = \RR$ for simplicity. For any time $n \in [N-1]$ and look-back window $k \in [n-1]$, the empirical minimizer $\widehat\btheta_{n, k} = \argmin_{\btheta \in \Omega} f_{n, k} (\btheta)$ has distribution $N ( \frac{1}{k} \sum_{ i=n-k }^{n-1} \btheta_i^* , \frac{1}{Bk} )$. If $\{ \btheta_i^*\}_{i = n-k}^{n-1}$ differ by at most $\cO ( 1 / \sqrt{Bk} )$, then the bias of $\widehat\btheta_{n, k} $ in estimating $\btheta_{n-1}^*$ is at most comparable to its stochastic error, so the distribution shift over the past $k$ periods can be ignored.
In general, we treat a length-$k$ segment of $\{ \btheta_n^*\}_{n = 1 }^{N}$ as stationary if its variation does not exceed $\cO \big( \sqrt{ \frac{d}{ B k } } \big)$, which leads to the following \Cref{def-segmentation-strong-cvx}.

\begin{definition}[Segmentation]\label{def-segmentation-strong-cvx}
The minimizer sequence $\{\btheta_n^*\}_{n=1}^N$ of $\{F_n\}_{n=1}^N$ is said to consist of $J$ \textbf{quasi-stationary segments} if there exist $0 = N_0 < N_1 < \cdots < N_J = N-1$ such that
\[
\max_{N_{j - 1} < i, k \leq N_j } \| \btheta_i^* - \btheta_k^* \|_2
\le
\sqrt{ \frac{2 M \sigma }{\rho} \max\left\{\frac{\sigma}{\rho M},1\right\} \cdot \frac{d}{B ( N_j - N_{j - 1} ) } }, \qquad \forall j\in[J]
\]
\end{definition}

We can always decompose any $\{\btheta_n^*\}_{n=1}^N$ into $N-1$ quasi-stationary segments by setting $N_j = j$ for each $j \in [J]$, where each segment only contains a single time period. In what follows, we will always take a segmentation of $\{\btheta_n^*\}_{n=1}^N$ that results in the smallest $J$, so that a larger $J$ indicates greater non-stationarity. When the environment is stationary, i.e.~$\btheta_1^* = \cdots = \btheta_N^*$, we have $J = 1$. 
The lemma below bounds $J$ in terms of the \emph{path variation} (\emph{PV}) or \emph{path length} $\sum_{ n = 1 }^{N-1 } \| \btheta_{n+1}^* - \btheta_n^* \|_2$, which is a popular measure of non-stationarity \citep{Zin03,JRS15,ZLZ18}. 
The proof is deferred to \Cref{sec-lem-PV-strongly-cvx-proof}.

\begin{lemma}[From path variation to segmentation]\label{lem-PV-strongly-cvx}
Suppose $\{\btheta_n^*\}_{n=1}^N$ consists of $J$ quasi-stationary segments, and define $V = \sum_{n=1}^{N-1} \| \btheta_{n+1}^* - \btheta_n^* \|_2$. Then $J \le 1 + C(BN/d)^{1/3} V^{2/3}$, where $C>0$ is a constant depending on $M$, $\rho$ and $\sigma$.
\end{lemma}

On the other hand, \Cref{eg-nonstationary-seq-strongly-cvx} below shows that sequences with the same path variation may have very different numbers of segments. An important reason is that while the path variation tracks all the distribution shifts, our segmentation aims to capture only those that lead to significant changes in the optimal solution. As a consequence, our measure of non-stationarity is often more optimistic and refined than the path variation. Indeed, we will later see that the former yields a tighter regret bound than the latter.

\begin{figure}[h]
	\centering
	\begin{subfigure}{0.24\textwidth}
	\includegraphics[scale=1]{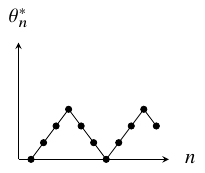}
	\caption{Large zigzags.} \label{fig-zigzag-large}
	\end{subfigure}
	\begin{subfigure}{0.24\textwidth}
	\includegraphics[scale=1]{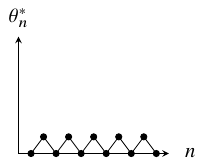}
	\caption{Small zigzags.} \label{fig-zigzag-small}
	\end{subfigure}
	\begin{subfigure}{0.24\textwidth}
	\includegraphics[scale=1]{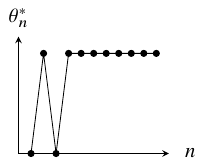}
	\caption{Uneven zigzags.} \label{fig-zigzag-uneven}
	\end{subfigure}
	\begin{subfigure}{0.24\textwidth}
	\includegraphics[scale=1]{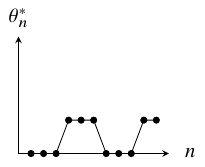}
	\caption{Alternating steps.} \label{fig-alternate-steps}
	\end{subfigure}
	\caption{Several non-stationarity patterns in \Cref{eg-nonstationary-seq-strongly-cvx}.}
\end{figure}

\begin{example}\label{eg-nonstationary-seq-strongly-cvx}
We consider several patterns of non-stationarity in the setting of \Cref{eg-Gaussian-mean}. For simplicity, we assume that $B=1$, $d=1$, $\Omega = [0, 1]$ and $N$ is large, and omit rounding a number to its nearest integer.
The following sequences $\{\theta_n^*\}_{n=1}^N$ all have path variation $V\asymp N^{1/2}$, so \Cref{lem-PV-strongly-cvx} implies $J\lesssim N^{2/3}$.
\begin{enumerate}
\item Large zigzags (\Cref{fig-zigzag-large}): For every $n\in[N]$, $|\theta_{n+1}^*-\theta_n^*|=N^{-1/2}$. Moreover, for each $k\in[N^{2/3}]$, $\theta_n^*$ is monotone on $(k-1)N^{1/3}<n\le kN^{1/3}$. Then, we can take $N_j\asymp jN^{1/3}$ and $J\asymp N^{2/3}$.
\item Small zigzags (\Cref{fig-zigzag-small}): For every $n\in[N]$, $\theta_{n+1}^*=\theta_n^*- (-1)^n N^{-1/2}$. Then, we can take $J=1$.
\item Uneven zigzags (\Cref{fig-zigzag-uneven}): For every $n\in[N^{1/2}]$, $|\theta^*_{n+1}-\theta_n^*| = 1$. Moreover, $\theta_n^*$ is constant on $N^{1/2}<n\le N$. Then, we can take $J\asymp N^{1/2}$, with $N_j=j$ for $j\in[J-1]$ and $N_{J}=N-1$.
\item Alternating steps (\Cref{fig-alternate-steps}): Choose any $u\in[N^{-1/2},N^{-1/6}]$. For  $k\in [ N^{1/2}u^{-1} ]$, the sequence $\theta_n^*$ is constant on $k N^{1/2}u< n \le (k+1) N^{1/2}u$, and $\theta^*_{k N^{1/2}u+1}=\theta^*_{ k N^{1/2}u} - (-1)^{k} u$. 
Then, each constant piece has length $N^{1/2}u$; each segment contains $N^{-1/2}u^{-3}$ constant pieces and thus has length $u^{-2}$. We can take $N_j\asymp j u^{-2}$ and $J\asymp N u^2 \in[1,N^{2/3}]$.
\end{enumerate}
\end{example}

We are now ready to present the dynamic regret of \Cref{alg-online}. See Appendix \ref{sec-thm-online-strongly-convex-proof-sketch} for a sketch of proof and \Cref{sec-thm-online-strongly-convex-proof} for a full proof. In both proofs we present a more refined bound.

\begin{theorem}[Regret bound]\label{thm-online-strongly-convex}
Let Assumptions \ref{assumption-bounded-domain}, \ref{assumption-strongly-convex} and \ref{assumption-concentration-strong-cvx} hold. Let $J_N$ be the number of quasi-stationary segments in $\{\btheta_n^*\}_{n=1}^N$.
Choose any $\alpha \in (0, 1]$. There exists a constant $\bar{C}_{\tau}>0$ such that if we choose $C_{\tau} \geq \bar{C}_{\tau}$ and run \Cref{alg-online} with
\[
\tau ( n , k ) = C_{\tau}
\frac{d }{ B k }\log  ( \alpha^{-1} + B + n  ), \quad\forall n\in\ZZ_+,~ k\in[n-1],
\]
then with probability at least $1 - \alpha$, the output of \Cref{alg-online} satisfies
\begin{equation}
\sum_{n=1}^{N} \bigg[ F_{n} ( \btheta_n ) -  F_{n} ( \btheta_n^* ) \bigg]  
\lesssim
\min \left\{ J_N \left( \frac{d}{B} + 1 \right),~ N \right\} ,\qquad\forall N\in\ZZ_+.
\label{eqn-thm-online-strongly-cvx}
\end{equation}
Here $\lesssim$ only hides a polylogarithmic factor of $B$, $N$ and $\alpha^{-1}$.
\end{theorem}

\Cref{thm-online-strongly-convex} states that the dynamic regret of \Cref{alg-online} scales linearly with the number of quasi-stationary segments $J_N$, so a less variable sequence $\{ \btheta^*_n \}_{n=1}^N$ leads to a smaller regret bound. We emphasize that \Cref{alg-online} attains this bound without any knowledge of the non-stationarity. In \Cref{sec-theory-lower-strong-cvx}, we provide a minimax lower bound that matches the regret bound \eqref{eqn-thm-online-strongly-cvx} up to logarithmic factors, showing the adaptivity of our algorithm to the unknown non-stationarity. In the stationary case where $\btheta^*_1 = \cdots = \btheta^*_N$, we have $J_N = 1$, and \Cref{alg-online} attains a logarithmic regret. We also mention that \Cref{thm-online-strongly-convex} continues to hold when $\widehat{\btheta}_{n,k}$ is only an approximate minimizer of $f_{n,k}$ satisfying $f_{n,k}(\widehat{\btheta}_{n,k}) - \min_{\btheta\in\Omega} f_{n,k}(\btheta) = \cO(\frac{d}{Bk})$.

As a corollary of a more refined version of \Cref{thm-online-strongly-convex} in \Cref{sec-thm-online-strongly-convex-proof}, we derive a near-optimal regret bound for \Cref{alg-online} in terms of the path variation $\sum_{n = 1 }^{N - 1} \| \btheta_{n + 1}^* - \btheta_n^* \|_2$. We prove \Cref{cor-segmentation-strongly-cvx} in \Cref{sec-cor-segmentation-strongly-cvx-proof}, and show its near-optimality in \Cref{sec-theory-lower} by providing a minimax lower bound that matches it up to logarithmic factors.

\begin{corollary}[PV-based regret bound]\label{cor-segmentation-strongly-cvx}
Consider the setting of \Cref{thm-online-strongly-convex} and define $V_N = \sum_{n = 1 }^{N - 1} \| \btheta_{n + 1}^* - \btheta_n^* \|_2 $. With probability at least $1-\alpha$, the output of \Cref{alg-online} satisfies
\[
\sum_{n=1}^{N} \bigg[ F_{n} ( \btheta_n ) - F_{n} ( \btheta_n^* ) \bigg]  
\lesssim
1 + \frac{d}{B} +
N^{1/3} \bigg( \frac{V_N d}{B} \bigg)^{2/3} + V_N ,\quad\forall N\in\ZZ_+.
\]
Here $\lesssim $ only hides a polylogarithmic factor of $B$, $N$ and $\alpha^{-1}$.
\end{corollary}

We now revisit \Cref{eg-nonstationary-seq-strongly-cvx} to illustrate that the segmentation-based bound in \Cref{thm-online-strongly-convex} can be much tighter than the PV-based bound in \Cref{cor-segmentation-strongly-cvx}. For the sequences in \Cref{eg-nonstationary-seq-strongly-cvx}, \Cref{thm-online-strongly-convex} gives a regret bound of $\widetilde{\cO}(J_N)$ which is often much smaller than $N^{2/3}$. In constrast, since $V_N \asymp N^{1/2}$, then \Cref{cor-segmentation-strongly-cvx} always gives a regret bound $\widetilde{\cO}(N^{2/3})$, failing to capture refined structures of non-stationarity.

\begin{remark}[Other variation metrics]\label{remark-variation-metrics}
The non-stationarity of the environment can also be quantified through other variation metrics. In the noiseless case where $f_n=F_n$ is assumed to be strongly convex, \cite{ZZh21} studied the squared path length $S_N = \sum_{n=1}^{N-1} \| \btheta_{n+1}^* - \btheta_n^* \|_2^2$ and the functional variation $\FV = \sum_{n=1}^{N-1} \| F_{n+1} - F_n \|_{\infty}$ and derived an $\cO ( \min \{ S_N , V_N , \FV \} )$ regret bound (ignoring the dependence on the dimension); \cite{BWa22} defined $C_N = \sum_{n=1}^{N-1} \| \btheta_{n+1}^* - \btheta_n^* \|_1$ as the path length and derived an $\widetilde{\cO}( d^{1/3} C_N^{2/3}N^{1/3})$ regret bound. Our results hold for more the challenging setting where $f_n$ is a random realization of $F_n$ and is not necessarily strongly convex. 
\cite{BGZ15} considered the functional variation $\FVB = \sum_{n=1}^{N-1} \sup_{\btheta \in \Omega^*} | F_{n+1} (\btheta) - F_n (\btheta) |$, where $\Omega^*$ is the convex hull of the minimizers $\{ \btheta_n^* \}_{n=1}^N$. For learning with noisy first-order feedback in constant dimension, they showed that the minimax optimal regret is $\widetilde{\cO} ( \sqrt{ \FVB  N } )$. In \Cref{sec-FV-bound-strong-cvx}, we recover the same regret bound from \Cref{thm-online-strongly-convex} by showing that the number of quasi-stationary segments $J_N$ is bounded by $1 + \cO \big( \sqrt{N \FVB  B / d} \big)$.
\end{remark}

\begin{remark}[Segmentation]\label{remark-segmentation}
The idea of quasi-stationary segments has appeared in various forms. \cite{BWa19} and \cite{CLL19} performed segmentation by comparing a certain path variation within a time interval against the stochastic error, in the settings of one-dimensional mean estimation and contextual bandits, respectively. In contrast, our segmentation uses the maximum variation between any two time periods within a segment, which can be substantially smaller than the path variation, enabling detection of more refined non-stationarity. 
In the noiseless ($f_n=F_n$) and exp-concave setting, \cite{BWa21} performed segmentation on a dynamic comparator sequence for regret analysis, which is not intrinsic to the environment. Moreover, it is not clear how their analysis can be extended to the noisy setting, where the empirical loss $f_n$ may not be strongly convex or exp-concave even if the population loss $F_n$ is. Finally, \cite{SKp22} proposed a similar concept for bandits named ``significant phases'', by comparing the dynamic regret under non-stationarity against the regret in the stationary case.
\end{remark}

\subsection{Lipschitz Population Losses}\label{sec-theory-Lipschitz}

Our second study concerns a less regular case where each $F_n$ is only assumed to be Lipschitz. The presentation parallels that of the strongly convex case in \Cref{sec-theory-strong-cvx}. We make the following assumption, which states that the empirical losses have sub-gaussian tails, and that the empirical and population losses are Lipschitz. In particular, the loss functions $\ell$ and $F_n$ need not be convex.

\begin{assumption}[Concentration and smoothness]\label{assumption-concentration-Lip}
There exist constants $\sigma, \lambda > 0$ such that for all $n\in\ZZ_+$ and $\bz_n \sim \cP_n$,
\begin{align*}
& \sup_{\btheta_1,\btheta_2\in\Omega}\big\|\ell(\btheta_1,\bz_n)- \ell(\btheta_2,\bz_n) - [F_n(\btheta_1)-F_n(\btheta_2)] \big\|_{\psi_2}\le \sigma, \\
& \sup_{\substack{\btheta_1,\btheta_2\in\Omega \\ \btheta_1\neq\btheta_2}} \frac{ |  F_n(\btheta_1)-F_n(\btheta_2)   | }{\|\btheta_1-\btheta_2\|_2}
\le 
\lambda  \qquad\text{and}\qquad \EE
\left(
\sup_{\substack{\btheta_1,\btheta_2\in\Omega \\ \btheta_1\neq\btheta_2}} \frac{ | \ell(\btheta_1,\bz_n)- \ell(\btheta_2,\bz_n)  | }{\|\btheta_1-\btheta_2\|_2}
\right)
\le 
\lambda \sqrt{d} .  
\end{align*}
\end{assumption}

Below we give several classical examples satisfying Assumption \ref{assumption-concentration-Lip}, where $\cZ = \RR^d$ and $\sigma_0>0$ is a constant. We leave their verifications to \Cref{sec-eg-Lipschitz}.

\begin{example}[Stochastic linear optimization]\label{eg-linear-opt}
Let $\Omega$ be a polytope and $\ell(\btheta,\bz)=\bz^\top\btheta$. Suppose $\sup_{\btheta \in \Omega} \| \bz^\top\btheta \|_{\psi_2} \leq \sigma_0$ and $\EE(\bz_n\bz_n^\top) \preceq \sigma_0^2\bI_d$. Then, Assumption \ref{assumption-concentration-Lip} holds with $\sigma=4\sigma_0$ and $\lambda=\sigma_0$.
\end{example}

\begin{example}[Quantile regression]\label{eg-quantile-regression}
Each sample $\bz_n\sim\cP_n$ takes the form $\bz_n=(\bx_n,y_n)\in\RR^d\times\RR$, where $\bx_n$ is the covariate vector and $y_n$ is the response. Let $\nu\in[0,1]$ and define the check loss $\rho_{\nu}(z)=(1-\nu)(-z)_+ + \nu z_+$. In quantile regression for the $\nu$-th conditional quantile of $y$ given $\bx$, we use the loss $\ell(\btheta,\bz)=\rho_{\nu}(y-\bx^\top\btheta)$. Suppose $\|\bx_n\|_{\psi_2}\le\sigma_0$. Then, Assumption \ref{assumption-concentration-Lip} holds with $\sigma\asymp M\sigma_0$ and $\lambda\asymp\sigma_0$.
\end{example}

\begin{example}[Newsvendor problem]\label{eg-newsvendor}
Let $d=1$. The sample $z_n\sim\cP_n$ represents the demand, and the decision $\theta$ represents the stocking quantity. The loss function is $\ell(\theta,z) = h(\theta-z)_+ + b(z-\theta)_+$, where $h$ is the holding/overage cost and $b$ is the backorder/underage cost. Suppose $\|z_n\|_{\psi_2}\le\sigma_0$. Then Assumption \ref{assumption-concentration-Lip} holds with $\sigma\asymp (h+b)M\sigma_0$ and $\lambda\asymp(h+b)\sigma_0$. We note that the newsvendor problem can be cast as a special case of quantile regression in \Cref{eg-quantile-regression} with $\nu = b/(h+b)$.
\end{example}

\begin{example}[Support vector machine]\label{eg-SVM}
Let $\Omega=B(\bm{0},M/2)$. Each sample $\bz_n\sim\cP_n$ takes the form $\bz_n=(\bx_n,y_n)\in\RR^d\times\{-1,1\}$, where $\bx_n$ is the feature vector and $y_n$ is the label. The loss function for the soft-margin support vector machine is given by $\ell(\btheta,\bz)=(1-y\bx^\top\btheta)_+ $. Suppose $\|\bx_n\|_{\psi_2}\le\sigma_0$. Then Assumption \ref{assumption-concentration-Lip} holds with $\sigma\asymp M\sigma_0$ and $\lambda=\sigma_0 $.
\end{example}



As in the strongly convex case in \Cref{sec-theory-strong-cvx}, we will decompose the underlying sequence $\{F_n\}_{n=1}^N$ into quasi-stationary segments. In general, the Lipschitz population loss $F_n$ does not have a unique minimizer, so the quantity $\|\btheta_i^*-\btheta_k^*\|_2$ in \Cref{def-segmentation-strong-cvx} is not well defined. Moreover, even if each $F_n$ has a unique minimizer, in the absence of strong convexity, a large distance $\|\btheta-\btheta_n^*\|_2$ does not necessarily imply a large sub-optimality gap $F_n(\btheta)-F_n(\btheta_n^*)$. Therefore, instead of the distance between minimizers, it is more suitable to measure the difference of function values. We will use $\|F_i-F_k\|_{\infty}$ to quantify the distribution shift between two periods $i$ and $k$.

To motivate the segmentation criterion, consider a one-dimensional example $(d=1)$. It is easily seen that the stochastic error $|f_{n,k}(\btheta)-F_{n,k}(\btheta)|$ is of order $1/\sqrt{Bk}$ for every fixed $\btheta\in\Omega$. If $\{F_i\}_{i=n-k}^{n-1}$ differ by at most $\cO(1/\sqrt{Bk})$, then the bias $F_{n-1}(\widehat{\btheta}_{n,k})-\min_{\btheta\in\Omega}F_{n-1}(\btheta)\lesssim \|F_{n-1}-F_{n,k}\|_{\infty}$ is at most comparable to the stochastic error. In this case, we can ignore the distribution shift over the past $k$ periods. In the general case, we think of a length-$k$ segment of $\{F_n\}_{n=1}^N$ as stationary if its variation does not exceed $\cO \big( \sqrt{\frac{d}{Bk} }  \big)$. This leads to \Cref{def-segmentation-Lip}.


\begin{definition}[Segmentation]\label{def-segmentation-Lip}
The function sequence $\{F_n\}_{n=1}^{N}$ is said to consist of $J$ \textbf{quasi-stationary segments} if there exist $0 = N_0 < N_1 < \cdots < N_J = N - 1$ such that
\[
\max_{N_{j - 1} < i, k \leq N_j } \| F_i - F_k \|_{\infty} 
\le
\frac{\sigma}{2}\sqrt{\frac{d}{B(N_j-N_{j-1})}} , \qquad \forall j \in [J] .
\]
\end{definition}

As in \Cref{sec-theory-strong-cvx}, for each sequence $\{F_n\}_{n=1}^N$, we will take a segmentation that leads to the minimal $J$. In \Cref{lem-PV-Lip} below, we bound $J$ in terms of the path variation $\sum_{n=1}^{N-1}\|F_{n+1}-F_n\|_{\infty}$. Its proof is given in \Cref{sec-lem-PV-Lip-proof}.

\begin{lemma}[From path variation to segmentation]\label{lem-PV-Lip}
Let $\{F_n\}_{n=1}^N$ consist of $J$ quasi-stationary segments, and define $V=\sum_{n=1}^{N-1}\|F_{n+1}-F_n\|_{\infty}$. Then $J\le 1 + C(BN/d)^{1/3}V^{2/3}$, where $C>0$ is a constant depending on $\sigma$.
\end{lemma}

In \Cref{thm-online-Lip}, we give a regret bound for \Cref{alg-online} in the Lipschitz case. Its proof can be found in \Cref{sec-thm-online-Lip-proof}, and contains a more refined bound.

\begin{theorem}[Regret bound]\label{thm-online-Lip}
Let Assumptions \ref{assumption-bounded-domain} and \ref{assumption-concentration-Lip} hold. Let $J_N$ be the number of quasi-stationary segments in $\{F_n\}_{n=1}^N$. Choose any $\alpha \in (0, 1]$. There exists a constant $\bar{C}_{\tau} > 0$ such that if we choose $C_{\tau} \geq \bar{C}_{\tau}$ and run \Cref{alg-online} with
\[
\tau(n,k)=C_{\tau}\sqrt{\frac{d}{Bk}\log( \alpha^{-1}+B + n )},\quad\forall n\in\ZZ_+,~ k\in[n-1],
\]
then with probability at least $1 - \alpha$, the output of \Cref{alg-online} satisfies
\begin{equation}
\sum_{n=1}^N \bigg[ F_{n} ( \btheta_n ) - \min_{ \btheta_n' \in \Omega } F_{n} ( \btheta_n' ) \bigg]
\lesssim
\min\left\{ J_N + \sqrt{J_NN\frac{d}{B}},~ N \right\} ,\qquad\forall N\in\ZZ_+.
\label{eqn-thm-online-Lip}
\end{equation}
Here $\lesssim$ only hides a polylogarithmic factor of $B$, $N$ and $\alpha^{-1}$.
\end{theorem}

\Cref{thm-online-Lip} shows that the dynamic regret of \Cref{alg-online} in the Lipschitz case is $\widetilde{\cO}(\sqrt{J_NN})$. As in the strongly convex case, the algorithm attains the bound \eqref{eqn-thm-online-Lip} without any prior knowledge of the non-stationarity. In \Cref{sec-theory-lower-Lip}, we provide a minimax lower bound that matches \eqref{eqn-thm-online-Lip} up to logarithmic factors, which shows that our algorithm automatically adapts to the unknown non-stationarity. For a stationary environment where $F_1=\cdots=F_N$, we have $J_N = 1$, which yields a regret bound of $\widetilde{\cO}(\sqrt{N})$. We remark that \Cref{thm-online-Lip} continues to hold when $\widehat{\btheta}_{n,k}$ is only an approximate minimizer of $f_{n,k}$ satisfying $f_{n,k}(\widehat{\btheta}_{n,k}) - \min_{\btheta\in\Omega} f_{n,k}(\btheta) = \cO(\sqrt{\frac{d}{Bk}})$.

As a corollary of \Cref{thm-online-Lip}, we derive the following PV-based regret bound. Its proof is deferred to \Cref{sec-cor-segmentation-Lip-proof}. 

\begin{corollary}[PV-based regret bound]\label{cor-segmentation-Lip}
Consider the setting of \Cref{thm-online-Lip} and define $V_N = \sum_{n = 1 }^{N - 1} \| F_{n+1} - F_n \|_{\infty} $. With probability at least $1-\alpha$, the output of \Cref{alg-online} satisfies
\[
\sum_{n=1}^{N} \left[ F_{n} ( \btheta_n ) - \min_{ \btheta_n' \in \Omega } F_{n} ( \btheta_n' ) \right]
\lesssim
1 + 
\sqrt{\frac{Nd}{B}}
+
N^{2/3}\left(\frac{V_Nd}{B}\right)^{1/3} + V_N, \qquad\forall N\in\ZZ_+.
\]
Here $\lesssim$ only hides a polylogarithmic factor of $B$, $N$ and $\alpha^{-1}$.
\end{corollary}

We note that the PV-based regret bound in \Cref{cor-segmentation-Lip} exhibits an $\widetilde{\cO}(V_N^{1/3}N^{2/3})$ dependence on $V_N$ and $N$, which also appears in \cite{BGZ15} for the setting of convex losses with noisy first-order feedback. In \Cref{sec-theory-lower}, we provide a minimax lower bound that matches the PV-based regret bound up to logarithmic factors.

%% file: main_theory_general.tex
\section{A General Theory of Learning under Non-Stationarity}\label{sec-theory-general}

In this section, we will develop a general framework for analyzing Algorithms \ref{alg-offline} and \ref{alg-online}. It contains as special cases the regret bounds in \Cref{sec-theory}. Our theory comprises two major components: a novel measure of similarity between functions and a general segmentation technique for dividing a non-stationary sequence into quasi-stationary pieces.

\subsection{Overview}\label{sec-theory-general-overview}

We begin with an overview of the main ideas to motivate our new notions. Recall that at time $n$, we seek to minimize $F_n$ based on noisy observations $\{ f_i \}_{i=1}^{n-1}$ of its predecessors $\{ F_i \}_{i=1}^{n-1}$. Each look-back window $k \in [n - 1]$ induces an estimated loss function $f_{n, k} = \frac{1}{k} \sum_{ i=n - k }^{n - 1} f_i$ and a candidate solution $\widehat\btheta_{ n, k} \in \argmin_{\btheta \in \Omega} f_{n, k} (\btheta) $. Since $f_{n, k}$ is an empirical approximation of a surrogate $F_{n, k} = \frac{1}{k} \sum_{ i=n - k }^{n - 1} F_i$ for $F_n$, we can apply statistical learning theory to bound their discrepancies, ensuring that any approximate minimizer of $f_{n, k}$ is also near-optimal for $F_{n, k}$, and vice versa.

Let $K \in [n-1]$ be the largest look-back window in which the environment only undergoes negligible changes. Ideally, this is the optimal window to use. However, the window $K$ depends on the unknown non-stationarity, and we wish to use data to find a window $\widehat{k}$ that is comparable to $K$. To this end, we study basic properties of the window $K$. By the definition of $K$, $F_n$ is very close to $\{ F_i \}_{i = n - K}^{n - 1}$ and thus $\{ F_{n, k} \}_{k=1}^K$. This, combined with the fact that $f_{n,k}$ is close to $F_{n,k}$, leads to the following observation.

\begin{fact}\label{fact-excess-0}
For all $k \in [K]$, any point $\btheta\in\Omega$ that is near-optimal for $f_{n, k}$ is also near-optimal for $F_{n}$, and vice versa.
\end{fact}

Since $\widehat{\btheta}_{n,K}\in\argmin_{\btheta\in\Omega} f_{n,K}(\btheta)$, Fact \ref{fact-excess-0} implies that $\widehat{\btheta}_{n,K}$ is near-optimal for $F_n$. Applying Fact \ref{fact-excess-0} again yields the following.

\begin{fact}\label{fact-excess}
For all $k \in [K]$, $\widehat{\btheta}_{n,K}$ is near-optimal for $f_{n,k}$, i.e.~$f_{n, k}  ( \widehat\btheta_{n, K }  ) - \min_{  \btheta \in \Omega } f_{n, k} (\btheta) = f_{n, k}  ( \widehat\btheta_{n, K }  ) - f_{n,k}(\widehat{\btheta}_{n,k})$ is small.
\end{fact}

\Cref{alg-offline} chooses a window $\widehat{k}$ according to a rule that mimics Fact \ref{fact-excess}. For simplicity, consider its simple version \Cref{alg-offline-simple}, which selects
\[
\widehat{k} = \max \left\{ k\in[n-1] : f_{n,i}(\widehat{\btheta}_{n,k}) - f_{n,i}(\widehat{\btheta}_{n,i}) \le \tau(i),~\forall i\in[k] \right\}.
\]
When is the performance of $\widehat{k}$ comparable to that of $K$?
\begin{itemize}
\item If $\widehat{k} \geq  K$, then the window selection rule implies
\[
f_{n, K} ( \widehat\btheta_{n, \widehat{k}}  ) - \min_{  \btheta' \in \Omega } f_{n, K} ( \btheta'  ) =
f_{n, K} ( \widehat\btheta_{n, \widehat{k}}  ) - f_{n, K} ( \widehat\btheta_{n, K }  ) \le \tau (K).
\]
In this case, we can use Fact \ref{fact-excess-0} to translate the bound above into a bound for $F_n(\widehat{\btheta}_{n,\widehat{k}})-\min_{\btheta_n'\in\Omega}F_n(\btheta_n')$.
\item If $\widehat{k} < K$, then the window selection rule implies the existence of $k \in [K - 1] $ such that
\[
f_{ n, k } ( \widehat\btheta_{n, K }  ) - \min_{  \btheta' \in \Omega } f_{n, k } ( \btheta'  ) =
f_{ n, k } ( \widehat\btheta_{n, K }  ) - f_{n, k } ( \widehat\btheta_{n, k }  ) > \tau (k).
\]
According to Fact \ref{fact-excess}, this cannot happen if the thresholds $\{ \tau(n,i) \}_{i=1}^{K-1}$ are sufficiently large.
\end{itemize}

Consequently, it is desirable to have large $\{ \tau (n,k) \}_{k=1}^{K-1}$ to keep $\widehat{k}$ from being too small, but small $\tau (n,K)$ for bounding the sub-optimality of $\widehat\btheta_{n, \widehat{k}}$. This is similar to controlling Type-I and Type-II errors in hypothesis testing. We choose $\tau (n,k)$ using simple bounds on the stochastic error of the empirical loss minimizer given by $Bk$ independent samples. In particular, $\tau (n,k) \asymp \frac{d}{Bk}$ and $\sqrt{ \frac{d}{Bk} }$ up to logarithmic factors for strongly convex and Lipschitz population losses, respectively.

To make the above analysis precise, we propose a novel notion of closeness between two functions: $f$ and $g$ with the same domain $\Omega$ are regarded as close if the sub-optimality gaps $f(\btheta) - \inf_{  \btheta' \in \Omega } f(\btheta')$ and $g (\btheta) - \inf_{  \btheta' \in \Omega } g (\btheta')$ can bound each other up to an affine transform (\Cref{defn-approx}). The slope and the intercept of the affine transform provide a quantitative measure. It will help us depict the concentration of the empirical loss $f_{n, k}$ around its population version $F_{n, k}$, as well as the discrepancy between $F_{n, k}$ and $F_n$ caused by the distribution shift over time. Moreover, it has convenient operation rules that enable the following reasoning:

\begin{itemize}
\item If $f_{n, k}$ is close to $F_{n, k}$, and if $F_{n, k}$ is close to $F_n$, then $f_{n, k}$ is close to $F_n$.
\item If $\{ F_i \}_{i=n - k}^{n - 1}$ are close to $F_n$, then the average $F_{n, k}$ is also close to $F_n$.
\end{itemize}

We have seen that \Cref{alg-offline} selects a window to maximize the utilization of historical data while keeping the cumulative bias under control. In the online setting, \Cref{alg-online} applies \Cref{alg-offline} in every time period to get a look-back window tailored to the local non-stationarity. If the whole sequence $\{ F_n \}_{n=1}^N$ consists of quasi-stationary segments, then \Cref{alg-online} is comparable to an oracle online algorithm that restarts at the beginning of each segment and treats data within the same segment as i.i.d. This observation leads to our formal notion of quasi-stationarity (\Cref{defn-quasi-stationary}) based on function closeness (\Cref{defn-approx}), and a segmentation technique (\Cref{def-segmentation}) for regret analysis.

\subsection{A Measure of Closeness between Two Functions}\label{sec-closeness}

We now introduce our measure of function closeness.

\begin{definition}[Closeness]\label{defn-approx}
	Suppose that $f, g : \Omega \to \RR$ are lower bounded and $\varepsilon, \delta \geq 0$. The functions $f$ and $g$ are said to be \textbf{$( \varepsilon , \delta )$-close} if the following inequalities hold for all $\btheta \in \Omega$:
\begin{align*}
& g ( \btheta ) - \inf_{ \btheta' \in \Omega } g(\btheta') \leq e^{\varepsilon}   \bigg(  f ( \btheta ) - \inf_{ \btheta' \in \Omega } f (\btheta') + \delta
\bigg) , \\[4pt]
& f ( \btheta ) - \inf_{ \btheta' \in \Omega } f(\btheta') \leq e^{\varepsilon}   \bigg(  g ( \btheta ) - \inf_{ \btheta' \in \Omega } g (\btheta') + \delta
\bigg) .
\end{align*}
In this case, we also say that $f$ is $(\varepsilon, \delta)$-close to $g$.
\end{definition}

The closeness measure reflects the conversion between the sub-optimality gaps of two functions. We give a more geometric interpretation through a sandwich-type inclusion of sub-level sets.

\begin{fact}
	For any lower bounded $h: \Omega \to \RR$ and $t \in \RR$, define the sub-level set
	\[
	S (h, t) = \bigg\{ \btheta \in \Omega :~ h (\btheta) \leq \inf_{  \btheta' \in \Omega } h (\btheta') + t \bigg\}.
	\]
	Two lower bounded functions $f,g : \Omega \to \RR$ are $(\varepsilon, \delta)$-close if and only if
	\[
	S \big( g, e^{-\varepsilon} t - \delta \big)
	\subseteq  S ( f, t ) \subseteq S \big( g, e^{\varepsilon} (t + \delta) \big) , \ \  \forall t \in \RR.
	\]
\end{fact}

Intuitively, $\delta$ measures the intrinsic discrepancy between two functions and $\varepsilon$ provides some leeway. The latter allows for a large difference between the sub-optimality gaps $f ( \btheta ) - \inf_{ \btheta' \in \Omega } f(\btheta') $ and $g ( \btheta ) - \inf_{ \btheta' \in \Omega } g (\btheta') $ when $\btheta$ is highly sub-optimal for $f$ or $g$. After all, we are mainly interested in the behaviors of $f$ and $g$ near their minimizers. Similar ideas are also used in the peeling argument in empirical process theory \citep{vdG00}.
Thanks to the scaling factor $e^{\varepsilon}$, our closeness measure gives a more refined characterization than the supremum metric $\| f - g \|_{\infty} = \sup_{  \btheta \in \Omega } |f(\btheta) - g(\btheta)|$. We illustrate this using the elementary example below.

\begin{example}
Let $\Omega = [-1, 1]$ and $a,b \in \Omega$. If $f(\theta) = |\theta - a|$ and $g(\theta) =  2 |\theta - b|$, then $f$ and $g$ are $(\log 2, |a - b| )$-close. In contrast, $\| f - g \|_{\infty} \geq 1$ always, even when $f$ and $g$ have the same minimizer $a=b$. To see this, since $f(-1) = 1 + a$, $g(-1) =2 + 2b$, $f(1) = 1 - a$ and $g(1) = 2 - 2b$, then
\[
\|f - g\|_{\infty}
\ge 
\frac{|f(-1)-g(-1)| + |f(1)-g(1)|}{2}
=
\frac{|1+2b-a|+|1-(2b-a)|}{2}
\ge 
1.
\]
\end{example}


We now provide user-friendly conditions for computing the closeness parameters. The proof is deferred to \Cref{sec-lem-sufficient-proof}.

\begin{lemma}\label{lem-sufficient}
Let $\Omega\subseteq \RR^d$ be closed and convex, with $\diam(\Omega)=M<\infty$. Let $f, g:~ \Omega \to \RR$. Suppose that $g$ is lower bounded.
	\begin{enumerate}
		\item\label{lem-sufficient-sup} If $D_0 = \sup_{ \btheta \in \Omega } | f ( \btheta ) - g ( \btheta ) - c |< \infty$ for some $c \in \RR$,	then $f$ and $g$ are $( 0, 2 D_0  )$-close.
		
		\item\label{lem-sufficient-grad-sup} If $D_1 = \sup_{ \btheta \in \Omega } \| \nabla f ( \btheta ) - \nabla g ( \btheta ) \|_2 < \infty$,	then $f$ and $g$ are $( 0, 2 M D_1   )$-close. 
		
		\item\label{lem-sufficient-grad-square} If the assumption in Part \ref{lem-sufficient-grad-sup} holds and there exists $\rho > 0$ such that $g$ is $\rho$-strongly convex over $\Omega$, then $f$ and $g$ are $\big(\log 2,~\frac{2}{\rho} \min \{ D_1^2 ,  \rho M D_1   \} \big)$-close.
		
		\item\label{lem-sufficient-minimizers} Suppose there exist $0 < \rho \leq L < \infty$ such that 
		$f$ and $g$ are $\rho$-strongly convex and $L$-smooth over $\Omega$. In addition, suppose that $f$ and $g$ attain their minima at some interior points $\btheta^*_f$ and $\btheta^*_g$ of $\Omega$, respectively.
		Then, $f$ and $g$ are $\big(  \log ( \frac{4L }{ \rho} ) , ~ \frac{\rho}{2}  \| \btheta^*_f - \btheta^*_g \|_2^2 \big)$-close.
	\end{enumerate}
\end{lemma}

For Lipschitz losses in \Cref{sec-theory-Lipschitz}, Part \ref{lem-sufficient-sup} of \Cref{lem-sufficient} will be useful for establishing the closeness between the empirical loss $f_{n,k}$ and the population loss $F_{n,k}$ with $D_0 \asymp \sqrt{\frac{d}{Bk}}$, as well as the closeness between two population losses $F_n$ and $F_i$. For strongly convex losses in \Cref{sec-theory-strong-cvx}, Part \ref{lem-sufficient-grad-square} of \Cref{lem-sufficient} applies to the pair $f_{n,k}$ and $F_{n,k}$ with $D_1 \asymp \sqrt{\frac{d}{Bk}}$, and Part \ref{lem-sufficient-minimizers} applies to the pair $F_n$ and $F_i$. We summarize these closeness results in \Cref{table-closeness}.

\begin{table}[h]
\centering
\begin{tabular}{|c|c|c|}
\hline
Function Pair & Strongly Convex Case & Lipschitz Case  \Tstrut\Bstrut \\ \hline
$f_{n,k}$ and $F_{n,k}$ & $\varepsilon\asymp 1,\quad \delta\asymp \frac{d}{Bk}$ & $\varepsilon\asymp 1,\quad \delta \asymp \sqrt{\frac{d}{Bk}}$  \Tstrut\Bstrut \\ \hline
$F_n$ and $F_i$ & $\varepsilon \asymp 1,\quad \delta \asymp \left\|\btheta_n^* - \btheta_i^*\right\|_2^2$ & $\varepsilon \asymp 1,\quad \delta \asymp \left\|F_n - F_i\right\|_{\infty}$ \Tstrut\Bstrut \\ \hline
\end{tabular}
\caption{Results of $(\varepsilon,\delta)$-closeness for \Cref{sec-theory}. Here $\asymp$ may hide constants such as smoothness parameters.} \label{table-closeness}
\end{table}


Our notion of closeness shares some similarities with the equivalence relation, including reflexivity, symmetry, and a weak form of transitivity. See \Cref{lem-approx} below for its nice properties and \Cref{sec-lem-approx-proof} for the proof.

\begin{lemma}\label{lem-approx}
	Let $f, g, h :~ \Omega \to \RR$ be lower bounded. Then,
	\begin{enumerate}
		\item\label{lem-approx-self} $f$ and $f$ are $(0, 0)$-close.
		\item\label{lem-approx-monotonicity} If $f$ and $g$ are $(\varepsilon, \delta)$-close, then $f$ and $g$ are $(\varepsilon', \delta')$-close for any $\varepsilon' \geq \varepsilon$ and $\delta' \geq \delta$.
		\item\label{lem-approx-shift} If $f$ and $g$ are $(\varepsilon, \delta)$-close and $a, b \in \RR$, $f + a$ and $g + b$ are $(\varepsilon , \delta )$-close.
		\item\label{lem-approx-symmetry} If $f$ and $g$ are $(\varepsilon, \delta)$-close, then $g$ and $f$ are $( \varepsilon,  \delta )$-close.
		\item\label{lem-approx-transitivity} If $f$ and $g$ are $(\varepsilon_1, \delta_1)$-close, and $g$ and $h$ are $(\varepsilon_2, \delta_2)$-close, then $f$ and $h$ are $ (  \varepsilon_1 + \varepsilon_2 ,  \delta_1  + \delta_2  )$-close.
		\item\label{lem-approx-general} If $\sup_{ \btheta \in \Omega } f ( \btheta ) - \inf_{ \btheta \in \Omega } f ( \btheta ) < F < \infty$ and $ \sup_{ \btheta \in \Omega } g ( \btheta ) - \inf_{ \btheta \in \Omega } g ( \btheta ) < G < \infty$, then $f$ and $g$ are $( 0,\max\{F,G\} )$-close. 
		\item\label{lem-approx-average} Suppose that $\{ f_i \}_{i=1}^m : \Omega \to \RR$ are lower bounded and $( \varepsilon, \delta )$-close to $g$. If $\{ \lambda_i \}_{i=1}^m \subseteq [ 0 , 1 ]$ and $\sum_{i=1}^{m} \lambda_i = 1$, then $\sum_{i=1}^{m} \lambda_i f_i$ and $g$ are $( \varepsilon , (e^{\varepsilon} + 1) \delta )$-close.
	\end{enumerate}
\end{lemma}

\subsection{Regret Analysis via Segmentation}\label{sec-segmentation}

To study the regret of \Cref{alg-online}, we first investigate its subroutine \Cref{alg-offline} at any given time $n$. We make the following assumption.

\begin{assumption}[Stochastic error]\label{assumption-approximation}
There exist $\varepsilon \geq 0$ and $ \{ \psi (n,k) \}_{n\in\ZZ_+,\,k\in[n-1]}$ such that for all $n\in\ZZ_+$, $\psi ( n , 1 ) \ge \cdots \ge \psi (n , n-1) \ge 0$, and for all $k \in [n-1]$, $f_{n, k}$ and $F_{n,k}$ are $( \varepsilon , \psi ( n , k) )$-close.
\end{assumption}

Assumption \ref{assumption-approximation} states that at time $n$, the stochastic error of pooling data from the most recent $k$ periods is characterized by $\psi(n,k)$. That $\psi(n,k)$ is decreasing in $k$ is consistent with the intuition that pooling more data reduces the stochastic error. In \Cref{table-closeness}, we have seen the closeness between $f_{n, k}$ and $F_{n,k}$ in the settings of \Cref{sec-theory}: $\psi(n,k)\asymp\frac{d}{Bk}$ in the strongly convex case, and $\psi(n,k)\asymp\sqrt{\frac{d}{Bk}}$ in the Lipschitz case, up to logarithmic factors.

We also impose the following conditions on the thresholds $\tau(n,k)$ used in \Cref{alg-online}.

\begin{condition}\label{condition-thresholds}
For all $n\in[N]$, $ \tau ( n , 1 ) \ge \cdots \ge \tau (n , n-1) \ge 0$, and for all $k\in[n-1]$, $\tau ( n , k) \ge 6 e^{5\varepsilon} \psi ( n , k) $. There exists $C \geq 1$ such that for any $n \in [N]$ and $k \in [n-1]$, $\tau (n, k) \leq C\tau (n , ( 2k ) \wedge n )$. Finally, for every $k \in [N-1]$, it holds that $\tau(k+1, k) \le \cdots \le \tau(N,k)$.
\end{condition}

Based on Condition \ref{condition-thresholds}, we can choose the threshold $\tau(n,k)$ as a constant multiple of the stochastic error $\psi(n,k)$. 
Therefore, for the strongly convex losses and the Lipschitz losses in \Cref{sec-theory}, we take $\tau(n,k) \asymp \frac{d}{Bk}$ and $\tau(n,k) \asymp \sqrt{\frac{d}{Bk}}$ up to some logarithmic factors, respectively. Both choices satisfy Condition \ref{condition-thresholds} with $C=2$.

We now present an excess risk bound for \Cref{alg-offline}. We provide a sketch of proof in Appendix \ref{sec-thm-excess-risk-proof-sketch} and a full proof in \Cref{sec-thm-excess-risk-proof}.

\begin{theorem}[Excess risk bound]\label{thm-excess-risk}
Fix $n\in[N]$. Consider \Cref{alg-offline} as a subroutine of \Cref{alg-online}, with $k_{s+1}\le 2k_s$ for each $s\in[m-1]$. Let Assumption \ref{assumption-approximation} and Condition \ref{condition-thresholds} hold. Define
\[
\bar{k} = \max \big\{ k \in [n-1] : \,  F_{n - k}, F_{n - k + 1} \cdots, F_{n-1} \text{ are } 
( \varepsilon,   \psi ( n , k ) )\text{-close to } F_{n-1} \big\}.
\]
Then the output $\btheta_{n}$ of \Cref{alg-offline}  satisfies
\[ 
F_{n-1} ( \btheta_{ n } ) - \inf_{ \btheta \in \Omega } F_{n-1} ( \btheta  )  \leq 2 e^{2 \varepsilon} C \tau( n , \bar{k} \wedge k_m  ).
\]
\end{theorem}

The window $\bar{k}$ is the precise mathematical formulation of the ideal window size $K$ in \Cref{sec-theory-general-overview}. It is the largest $k$ for which the bias between $F_{n-1}$ and each of $F_{n-k},F_{n-k+1},...,F_{n-1}$ is no more than the stochastic error $\psi(n,k)$. It balances the bias and stochastic error, both of which are of the order $\psi(n,\bar{k})$. Consequently, the associated decision $\widehat{\btheta}_{n,\bar{k}}$ has excess risk of the order $\psi(n,\bar{k})$. \Cref{thm-excess-risk} shows that the window $k_{\widehat{s}}$ chosen by \Cref{alg-offline} is a good approximation of $\bar{k}$, in the sense that the excess risk for $\btheta_{n}=\widehat{\btheta}_{n,k_{\widehat{s}}}$ has order $\tau(n,\bar{k}\wedge k_m)$, which is approximately $\psi(n,\bar{k})$. 

We proceed to analyze \Cref{alg-online} by approximating the sequence $\{ F_{n} \}_{n=1}^N$ with approximately stationary pieces. We first define a concept of quasi-stationarity through our notion of function closeness, and then introduce a general definition of segmentation.

\begin{definition}[Quasi-stationarity]\label{defn-quasi-stationary}
	Let $n \in \ZZ_+$, $\varepsilon \geq 0$ and $\delta \geq 0$. A sequence of functions $\{ g_i \}_{i=1}^n$ is said to be \textbf{$(\varepsilon, \delta)$-quasi-stationary} if for all $i,j \in [n]$, $g_i$ and $g_j$ are $(\varepsilon , \delta )$-close.
\end{definition}


\begin{definition}[Segmentation]\label{def-segmentation}
The function sequence $\{F_n\}_{n=1}^N$ is said to consist of $J$ \textbf{quasi-stationary segments} if there exist $\varepsilon \ge 0$, integers $0 = N_0 < N_1 < \cdots < N_J = N-1$ and non-negative numbers $\{\delta_j\}_{j=1}^J$ such that for every $j \in [J]$, 
\begin{itemize}
\item The sequence $\{F_i\}_{i=N_{j-1}+1}^{N_j}$ is $(\varepsilon,  \min_{ N_{j-1} < n \le N_j} \psi ( n, n - N_{j - 1})  )$-quasi-stationary.
\item $F_{N_{j} }$ and $F_{ N_j + 1 }$ are $(\varepsilon , \delta_j)$-close.
\end{itemize}
We call $\{N_j\}_{j=1}^J$ the \textbf{knots} and $\{\delta_j\}_{j=1}^J$ the \textbf{jumps}.
\end{definition}

In \Cref{def-segmentation}, we characterize the non-stationarity of the environment by the number of quasi-segments $J$ as well as the scales of the jumps $\{\delta_j\}_{j=1}^J$ between consecutive segments. It is a generalization of \Cref{def-segmentation-strong-cvx} and \Cref{def-segmentation-Lip} in \Cref{sec-theory}.

We are now ready to present the regret bound for \Cref{alg-online}. We provide a sketch of proof for \Cref{thm-regret} in Appendix \ref{sec-thm-regret-proof-sketch}, and a full proof in \Cref{sec-thm-regret-proof}. 

\begin{theorem}[Regret bound]\label{thm-regret}
Let Assumption \ref{assumption-approximation} and Condition \ref{condition-thresholds} hold. Suppose $\{F_n\}_{n=1}^N$ consists of $J$ quasi-stationary segments with knots $\{N_j\}_{j=1}^J$ and jumps $\{\delta_j\}_{j=1}^J$. Define the quantity $U = \max_{n \in [N]} \big[  \sup_{  \btheta \in \Omega } F_n ( \btheta ) - \inf_{  \btheta \in \Omega } F_n (\btheta) \big]$ and let $\reg ( n ) =   \sum_{i = 1 }^{ n } \min \{ \tau( N,  i ) , U \}$. Then the output $\{\btheta_n\}_{n=1}^N$ of \Cref{alg-online} satisfies
\[
\sum_{n=1}^{N} \left[ F_n ( \btheta_n ) - \inf_{ \btheta_n' \in \Omega } F_n ( \btheta_n' ) \right] \
\le
\left[ F_1 ( \btheta_1 ) - \inf_{ \btheta \in \Omega } F_1 ( \btheta ) \right]
+
3 e^{3 \varepsilon} C^2 \sum_{j = 1}^J \reg( N_j - N_{j - 1} )
+ e^{\varepsilon}  \sum_{j=1}^{J} \delta_{j}.
\]
\end{theorem}

\Cref{thm-regret} contains \Cref{thm-online-strongly-convex} and \Cref{thm-online-Lip} as special cases. Its regret bound consists of three terms. The first term results from our initial guess $\btheta_1$. In the second term, each summand is the regret incurred in the interior of a quasi-stationary segment. The third term is the cost of approximating $F_{N_j + 1}$ by $F_{N_j}$ at the boundary between quasi-stationary segments.

%% file: main_theory_lower.tex
\section{Minimax Lower Bounds and Adaptivity}\label{sec-theory-lower}

In this section, we present minimax lower bounds that match the regret bounds in \Cref{sec-theory} up to logarithmic factors. Since SAWS (\Cref{alg-online}) is agnostic to the amount of distribution shift, our results show its adaptivity to the unknown non-stationarity.

\subsection{Strongly Convex Population Losses}\label{sec-theory-lower-strong-cvx}

To prove the sharpness of \Cref{thm-online-strongly-convex} and \Cref{cor-segmentation-strongly-cvx}, we consider simple classes of online Gaussian mean estimation problems described in Example \ref{eg-Gaussian-mean}. Fix a time horizon $N\in\ZZ_+$.

\begin{definition}[Problem classes]\label{defn-class-strongly-cvx}
	Let $\Omega = B (\bm{0}, 1)$. Define $\cZ$, $\ell$ and $c$ as in \Cref{eg-Gaussian-mean}. For $J\in[N-1]$, define the problem class
	\begin{align*}
	\mathscr{P} ( J )  = \bigg\{ ( \cP_1 , \cdots, \cP_N ) :~ &\cP_n = N( \btheta_n^*, \bI ) \text{ and }  \btheta_n^* \in B(\bm{0},1/2)  ,~\forall n \in [N],\\
	& \text{ there exist $0=N_0<\cdots<N_{J}= N-1$ such that} \\
	& \max_{N_{j - 1} < i, k \leq N_j } \| \btheta_i^* - \btheta_k^* \|_2 
	\leq 
	\sqrt{  \frac{ 8c^2 d}{B ( N_j - N_{j - 1} ) } },
	~\forall j \in [J]
	 \bigg\} .
	\end{align*}
	In addition, for any $V \geq 0 $, define
	\[
	\mathscr{Q} (V) = \bigg\{ ( \cP_1 , \cdots, \cP_N ) :~ \cP_n = N( \btheta_n^*, \bI ),~ \btheta_n^* \in B(\bm{0}, 1/2) ,~ \sum_{ n=1 }^{N - 1} \| \btheta^*_{n+1} - \btheta^*_n \|_2 \leq V \bigg\} .
	\]
\end{definition}

For every problem instance in $\mathscr{P} ( J ) $ or $\mathscr{Q} (V)$, Assumptions \ref{assumption-bounded-domain}, \ref{assumption-strongly-convex} and \ref{assumption-concentration-strong-cvx} hold with $M = 2$, $\sigma_0 = \rho=L=\lambda=1$ and $\sigma=c$. The set $\mathscr{P} ( J )$ consists of minimizer sequences $\{\btheta_n^*\}_{n=1}^N$ with at most $J$ quasi-stationary segments, and $\mathscr{Q} (V)$ consists of minimizer sequences $\{\btheta_n^*\}_{n=1}^N$ with path variation at most $V$.

\Cref{thm-lower-strongly-cvx} below shows that for any algorithm, there exists a problem instance in the class such that the expected regret is at least comparable to the upper bound in \Cref{thm-online-strongly-convex}. See \Cref{sec-thm-lower-strongly-cvx-proof} for a stronger version and its proof.

\begin{theorem}[Lower bound]\label{thm-lower-strongly-cvx}
Assume $N\ge 2$ and that $J \in [N-1]$ divides $N-1$. There exists a universal constant $C>0$ such that
\[
\inf_{\cA} \sup_{ ( \cP_1,\cdots, \cP_N ) \in \mathscr{P} ( J )  } \EE \left[	\sum_{n=1}^{N} \bigg(F_{n} ( \btheta_n ) - F_{n} ( \btheta_n^* ) \bigg) \right] 
\geq C \min \left\{ J\left(\frac{d}{B}+1\right) ,N \right\}.
\]
The infimum is taken over all online algorithms $\cA$ for Problem \ref{problem-online}, and $\{\btheta_n\}_{n=1}^N$ is the output of $\cA$. 
\end{theorem}

Comparing the upper bound in \Cref{thm-online-strongly-convex} and the matching lower bound in \Cref{thm-lower-strongly-cvx}, we see that \Cref{alg-online} achieves the minimax optimal regret up to polylogarithmic factors for every $J$, adapting to the unknown non-stationarity. 

From the stronger version of \Cref{thm-lower-strongly-cvx} in \Cref{sec-thm-lower-strongly-cvx-proof}, we can easily derive a lower bound expressed using the path variation. The proof is deferred to \Cref{sec-cor-lower-strongly-cvx-proof}.

\begin{corollary}[PV-based lower bound]\label{cor-lower-strongly-cvx}
	Assume $N \geq \max \{ 2 ,  d / B \} $ and $V \leq N \min \{ B / d , \sqrt{ d / B } \}$. There is a universal constant $C>0$ such that
	\[
	\inf_{\cA} 	\sup_{ ( \cP_1,\cdots, \cP_N ) \in \mathscr{Q} (V) } \EE \left[
	\sum_{n=1}^{N} \bigg(F_{n} ( \btheta_n ) - F_{n} ( \btheta_n^* ) \bigg) \right] 
	\ge
	C \left[ 1 +
	\frac{d }{ B}  + 
	N^{1/3} \bigg( \frac{V d}{B} \bigg)^{2/3} 
	\right]  .
	\]
The infimum is taken over all online algorithms $\cA$ for Problem \ref{problem-online}, and $\{\btheta_n\}_{n=1}^N$ is the output of $\cA$. 
\end{corollary}

When $V \leq N  ( d / B )^2$, we have $V \leq N^{1/3}  ( V d / B  )^{2/3}$, and the regret bound in \Cref{cor-segmentation-strongly-cvx} simplifies to $1 +
\min  \{  d / B , N \} + N^{1/3}  ( V d / B  )^{2/3}$. Therefore, \Cref{cor-lower-strongly-cvx} shows that \Cref{alg-online} adapts to the unknown path variation when $0 \leq V \leq N \min \{ B / d , (  d / B )^2 \}$.

\subsection{Lipschitz Population Losses}\label{sec-theory-lower-Lip}

Finally, we present minimax lower bounds that match the regret bounds in \Cref{thm-online-Lip} and \Cref{cor-segmentation-Lip} up to logarithmic factors. We consider a class of stochastic linear optimization problems in \Cref{eg-linear-opt}.

\begin{definition}[Stochastic linear optimization]\label{defn-linear-opt}
Define $B_{\infty} ( \bx, r ) = \{ \by \in \RR^d :~ \| \by - \bx \|_{\infty} \leq r \}$ for any $\bx \in \RR^d$ and $r \geq 0$. For any $\bmu \in B_{\infty} ( \bm{0}, 1/2 )$, denote by $\cP ( \bmu  )$ the distribution of $\bz =  \sqrt{d} \bx \circ \by$, where $\bx$ and $\by$ are independent, the entries $\{ x_j \}_{j=1}^d$ of $\bx$ are independent, $\PP ( x_j = \pm 1 ) = \frac{1 }{2} \pm \mu_j$, $\by$ is uniformly distributed over $\{ \bm{e}_j \}_{j=1}^d$, and $\circ$ denotes the entry-wise product. 
Let $\cZ = \RR^d$, $\Omega = B_{\infty} ( \bm{0} , 1 / \sqrt{d} )$, $\ell ( \btheta, \bz ) =  \bz^{\top} \btheta$ and $F_{\bmu} (\cdot) = \EE_{\bz \sim \cP (\bmu) } \ell ( \cdot, \bz )$. 
\end{definition}

When $\by = \bm{e}_j$, $\ell ( \btheta, \bz ) =  \sqrt{d} \theta_j x_j$. We have $|\ell ( \btheta, \bz )| \leq 1$ for all $\btheta \in \Omega$, and $\EE(\bz\bz^\top)=\bI_d$. Hence, $\cP(\bmu)$ satisfies the conditions in \Cref{eg-linear-opt} with $\sigma_0 = 1$. Note that $\EE \bz = \bmu / \sqrt{d}$, $F_{\bmu} (\btheta) = \bmu^{\top} \btheta / \sqrt{d} $ and $\| F_{\bmu} - F_{\bnu} \|_{\infty} = \| \bmu - \bnu \|_1 / d$. We now construct two classes of learning problems similar to those in \Cref{defn-class-strongly-cvx}.

\begin{definition}[Problem classes]
For $J\in[N-1]$, define the problem class
\begin{align*}
	\mathscr{P} ( J )  = \bigg\{ ( \cP_1 , \cdots, \cP_N ) :~ &\cP_n = \cP(\bmu_n^*) \text{ and }  \bmu_n^* \in B_{\infty} (\bm{0}, 1/2)  ,~\forall n \in [N],\\
	& \text{ there exist $0=N_0<\cdots<N_{J}= N-1$ such that} \\
	& \frac{1}{d} \sum_{n = N_{j-1} + 1 }^{ N_j - 1 }
\| \bmu^*_{n+1} - \bmu^*_n \|_1
\leq 
 \sqrt{  \frac{ d}{B ( N_j - N_{j - 1} ) } },
	~\forall j \in [J]
	 \bigg\} .
	\end{align*}
In addition, for any $V \geq 0 $, define
\[
\mathscr{Q} (V) = \bigg\{ ( \cP_1 , \cdots, \cP_N ) :~ \cP_n = \cP (\bmu_n^*),~ \bmu_n^* \in B_{\infty} (\bm{0}, 1/2)  ,~ \frac{1}{d}  \sum_{ n=1 }^{N - 1} \| \bmu^*_{n+1} - \bmu^*_n \|_1 \leq V  \bigg\} .
\]
\end{definition}

For every problem instance in $\mathscr{P} ( J ) $ or $\mathscr{Q} (V)$, Assumption \ref{assumption-concentration-Lip} holds with $\sigma=4$ and $\lambda=2$. The set $\mathscr{P} ( J )$ corresponds to function sequences $\{F_n\}_{n=1}^N$ with at most $J$ quasi-stationary segments, and $\mathscr{Q} (V)$ corresponds to function sequences $\{F_n\}_{n=1}^N$ with path variation at most $V$. We are now ready to present our lower bounds. See \Cref{sec-thm-lower-Lip-proof} for the proof.

\begin{theorem}[Lower bound]\label{thm-lower-Lip}
Assume $N\ge 2$ and that $J\in[N-1]$ divides $N-1$. There exists a universal constant $C>0$ such that
\[
\inf_{\cA} \sup_{ ( \cP_1,\cdots, \cP_N ) \in \mathscr{P} ( J )  } \EE \left[	\sum_{n=1}^{N} \bigg(F_{n} ( \btheta_n ) - \min_{ \btheta_n' \in \Omega } F_{n} ( \btheta_n' ) \bigg) \right]
\ge
C \min \left\{ J + \sqrt{\frac{JNd}{B}},~ N \right\}.
\]
The infimum is taken over all online algorithms $\cA$ for Problem \ref{problem-online}, and $\{\btheta_n\}_{n=1}^N$ is the output of $\cA$. 
\end{theorem}

As the upper bound in \Cref{thm-online-Lip} matches the lower bound in \Cref{thm-lower-Lip}, we see that \Cref{alg-online} achieves the minimax optimal regret up to polylogarithmic factors for every $J$, and thus adapts to the unknown non-stationarity.

Finally, we present a lower bound based on the path variation. The proof is given in \Cref{sec-cor-lower-Lip-proof}.

\begin{corollary}[PV-based lower bound]\label{cor-lower-Lip}
When $N \geq \max \{ 2 , d / B \}$ and $0 \leq V \leq N \min \{ B / d,  \sqrt{d / B}   \} /6$, it holds that
	\[
	\inf_{\cA} \sup_{ ( \cP_1,\cdots, \cP_N ) \in \mathscr{Q} ( V )  } 
	\EE \left[
	\sum_{n=1}^{N} \left(F_{n} ( \btheta_n ) - \min_{ \btheta_n' \in \Omega } F_{n} ( \btheta_n' ) \right) \right]
	\ge C
	\bigg[
	1 + \sqrt{ \frac{ N d }{ B } }  + N^{2/3} \bigg( \frac{V d}{B} \bigg)^{1/3}
	\bigg].
	\]
The infimum is taken over all online algorithms $\cA$ for Problem \ref{problem-online}, and $\{\btheta_n\}_{n=1}^N$ is the output of $\cA$. 
\end{corollary}

When $V \leq N \sqrt{ d / B}$, we have $V \leq N^{2/3} ( V d / B )^{1/3}$, and the regret bound in \Cref{cor-segmentation-Lip} simplifies to $1 + \sqrt{Nd / B} + N^{2/3} ( V d / B )^{1/3}$. Therefore, \Cref{alg-online} adapts to the unknown path variation when $0 \leq V \le N \min \{ \sqrt{d / B}, B / d \} /6$.

%% file: main_experiments.tex
\section{Numerical Experiments}\label{sec-experiments}

In this section, we test the practical performance of our algorithm SAWS (\Cref{alg-online}) on synthetic and real data. To illustrate the adaptivity of our algorithm, we will compare it against fixed-window algorithms $\MA(k)$ that only use a fixed look-back window $k$ in every period $n\in[N]$. The detailed description of $\MA(k)$ is given in \Cref{alg-fixed-window}.

\begin{algorithm}[t]
	{\bf Input:} Window size $k$.\\
	Choose any $\btheta_1\in\Omega$.\\
	{\bf For $n = 2,\cdots, N$:}\\
	\hspace*{.6cm} Let $r = k\wedge(n-1)$, and compute a minimizer $\btheta_n$ of $f_{n,r} = \frac{1}{r} \sum_{i=n-r}^{n-1}f_i$.\\
	{\bf Output:} $\{ \btheta_n \}_{n = 1}^N$.
	\caption{Fixed-Window Moving Average $\MA(k)$}
	\label{alg-fixed-window}
\end{algorithm}

\subsection{Synthetic Data}

In the synthetic data experiment, we take one problem instance from the strongly convex case (\Cref{sec-theory-strong-cvx}), and one from the Lipschitz case (\Cref{sec-theory-Lipschitz}).  For both instances, we consider time horizons $N\in \cN = \{ 250,500,1000,2000,4000,8000 \}$. We will compare against benchmarks $\MA(k)$ with $k\in \left\{ \lceil N^p \rceil : p = \frac{1}{3},\frac{1}{2},\frac{2}{3},1 \right\}$.

\paragraph{Strongly convex instance.} We consider online linear regression (\Cref{eg-linear-regression}) under non-stationarity, with $d=10$, $M=12$, $\sigma_0=1$, $B=1$ and $N\in\cN$. In each period $n\in[N]$, a sample $(\bx_n,y_n)$ is generated from $y_n = \bx_n^\top\btheta_n^* + \varepsilon_n$, with $\bx_n \sim N(\bm{0},\bI_d)$ and $\varepsilon_n \sim N(0,1)$ independent. The minimizer sequence $\{\btheta_n^*\}_{n=1}^N$ is piecewise constant and has the following pattern. Let $n_1=5\lceil N^{1/3} \rceil$, $n_2=5\lceil N^{1/6} \rceil$ and $n_3=5\lceil N^{1/2} \rceil$. The horizon is divided into segments of equal length $n_1+n_2+2n_3$. Within each segment, in the $n_1$-th, $(n_1+n_3)$-th, $(n_1+n_3+n_2)$-th and $(n_1+n_3+n_2+n_3)$-th periods, $\btheta_n^*$ switches to a point sampled uniformly at random from $B(\bm{0},M/4)\subseteq\RR^d$.

\paragraph{Lipschitz instance.} We consider online stochastic linear optimization (\Cref{eg-linear-opt}) under non-stationarity, with $\Omega = \{\btheta\in\RR_+^d: \|\btheta\|_1 \le 1\}$, $d=10$, $B=1$ and $N\in\cN$. For each $n\in[N]$, $\cP_n = N(\bmu_n,\bI_d)$. The sequence $\{\bmu_n\}_{n=1}^N$ is piecewise constant and has the following pattern. Let $n_1=\lceil N^{1/2} \rceil$, $n_2=\lceil N^{1/6} \rceil$ and $n_3=\lceil N^{1/3} \rceil$. The horizon is divided into segments of equal length $n_1+n_2+2n_3$. Within each segment, in the $n_1$-th, $(n_1+n_3)$-th, $(n_1+n_3+n_2)$-th and $(n_1+n_3+n_2+n_3)$-th periods, $\bmu_n$ switches to a point generated by randomly picking half of the entries to be uniform over $\{-1,1\}^{d/2}$, and the other half uniform over $[-1,1]^{d/2}$.

We choose the thresholds $\{\tau(n,k)\}_{n\in\ZZ_+,k\in[n-1]}$ for SAWS according to \Cref{thm-online-strongly-convex} and \Cref{thm-online-Lip}. For the strongly convex instance, we take $\alpha=0.1$ and $C_{\tau} = 0.3$. For the Lipschitz instance, we take $\alpha=0.1$ and $C_{\tau} = 0.5$. 

\begin{figure}[h]
	\centering
    \begin{subfigure}{0.48\textwidth}
    	\centering
        \includegraphics[scale=0.6]{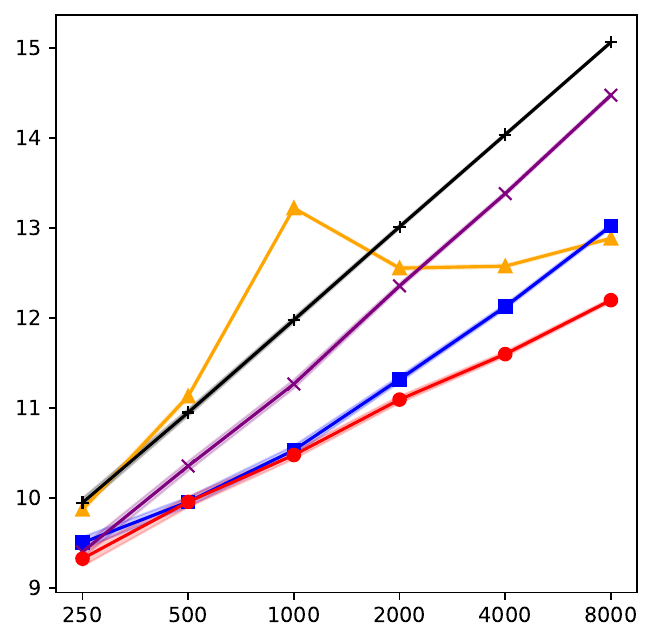}
        \caption{Strongly convex instance.}
	\end{subfigure}
    \begin{subfigure}{0.48\textwidth}
        \centering
        \includegraphics[scale=0.6]{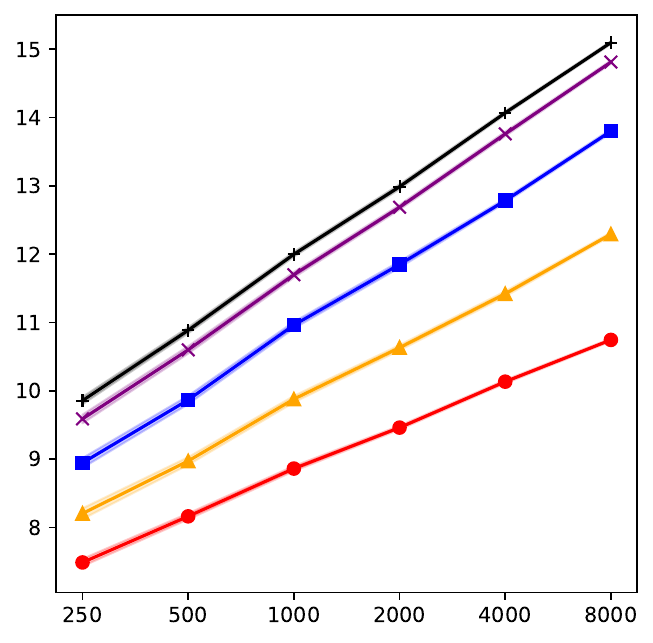}
        \caption{Lipschitz instance.}
	\end{subfigure}
	\caption{Log-log plots of dynamic regrets of SAWS and fixed-window benchmarks on synthetic data. Horizontal axis: time horizon $N\in\cN$. Vertical axis: logarithm of dynamic regret $\log_2\sum_{n=1}^N[F_n(\btheta_n) - \inf_{\btheta'\in\Omega} F_n(\btheta')]$. Red circles: SAWS (\Cref{alg-online}). Orange triangles: $\MA(\lceil N^{1/3} \rceil)$. Blue squares: $\MA(\lceil N^{1/2} \rceil)$. Purple $\times$'s: $\MA(\lceil N^{2/3} \rceil)$. Black $+$'s: $\MA(N)$. \label{fig-synthetic}}
\end{figure}

In \Cref{fig-synthetic}, we present the log-log plots for the dynamic regrets of SAWS and the benchmarks $\MA(k)$. The curves and error bands show the means and $1.96$ times the standard errors over $50$ random seeds, respectively. The latter gives $95$\% confidence intervals for the mean dynamic regrets of the methods, which have small widths.

In both instances, SAWS consistently outperforms the fixed-window benchmarks. The slopes of its curves are generally smaller than those of the benchmarks, indicating smaller orders of dynamic regrets. This demonstrates the adaptivity of SAWS to unknown non-stationarity.

\subsection{Real Data: Electricity Demand Prediction}\label{sec-experiments-electricity}

Our first real data experiment uses a electricity demand dataset maintained by the Australian Bureau of Meteorology and collected by \cite{Koz20}.
We study the daily electricity demand in Victoria, Australia from January 1st, 2016 to October 6th, 2020. In \Cref{fig-electricity-pattern} of \Cref{sec-patterns}, we plot the pattern of the electricity demand over time.

Our task is to use linear regression (\Cref{eg-linear-regression}) to predict the daily electricity demand $y_n$ given features $\bx_n$ on the same day, including minimum and maximum temperatures, rainfall and solar exposure. Along with an additional intercept term, this yields a feature vector of length $d=5$. Each day is treated as a time period, so there are $N=1760$ periods in total. We consider the setting where $M=\diam(\Omega)$ is large, and for simplicity we will set $\Omega = \RR^d$. We set $C_{\tau} = 10$ in SAWS, and will compare it with $\MA(k)$, $k\in\{1,7,14,30,180,365,1826\}$.

In \Cref{fig-loss-electricity}, we plot the per-period losses of SAWS and $\MA(k)$, given by $\frac{1}{N}\sum_{n=1}^N\left[ \frac{1}{2} (y_n - \bx_n^\top\btheta_n)^2 \right]$. Among the fixed-window benchmarks $\MA(k)$ considered, the optimal fixed window is $k^*=30$ days. We see that the performance of SAWS is comparable to that of $\MA(30)$.

In \Cref{fig-windows-electricity}, we visualize the rolling window picked by SAWS. We observe that SAWS adaptively selects rolling windows which roughly align with the non-stationarity pattern in \Cref{fig-electricity-pattern}.

\subsection{Real Data: Hospital Nurse Staffing}\label{sec-experiments-hospital}

Finally, we test our method on an emergency department (ED) visits dataset maintained by the New York City (NYC) government \citep{NYC24}. The dataset contains daily and weekly ED visit counts over time in NYC for various syndromes. In \Cref{fig-ED-pattern} of \Cref{sec-patterns}, we plot the weekly ED visit counts for vomiting from January 7th, 2019 to December 31st, 2023.

We study the problem of nurse staffing for this date range, where the goal is to decide the appropriate number of nurses to schedule each week. Following \cite{KMS23}, we formulate it as a newsvendor problem (\Cref{eg-newsvendor}), take the weekly demand for nurse staffing to be the weekly patient visits divided by $3$, and set the critical ratio as $b/(b+h) = 0.7$. For simplicity, we take $b = 0.7$ and $h=0.3$, and set $\Omega=\RR$. We take $C_{\tau} = 5$ in SAWS, and compare it with $\MA(k)$, $k\in\{1,2,4,26,52,104,208\}$.

\begin{figure}[h]
	\centering
		\begin{subfigure}{0.49\textwidth}
    	\centering
        \includegraphics[scale=0.45]{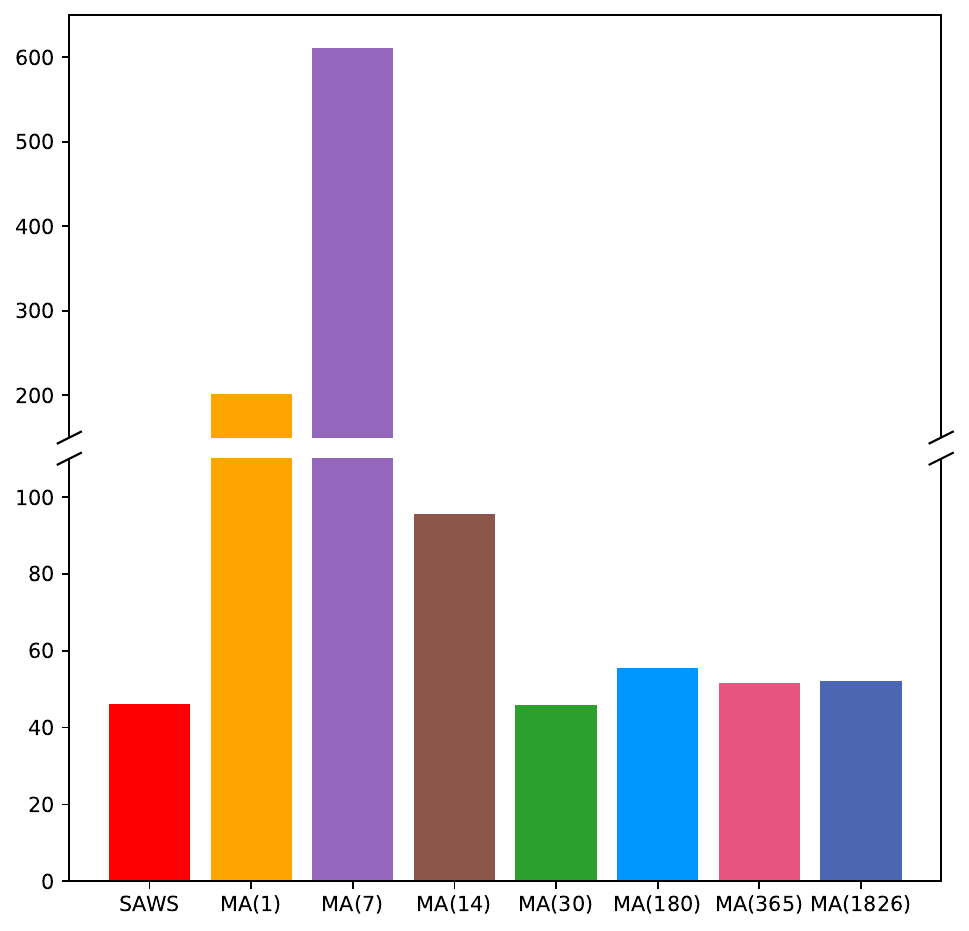}
        \caption{Electricity data. \label{fig-loss-electricity}}
	\end{subfigure}
    \begin{subfigure}{0.49\textwidth}
        \centering
        \includegraphics[scale=0.45]{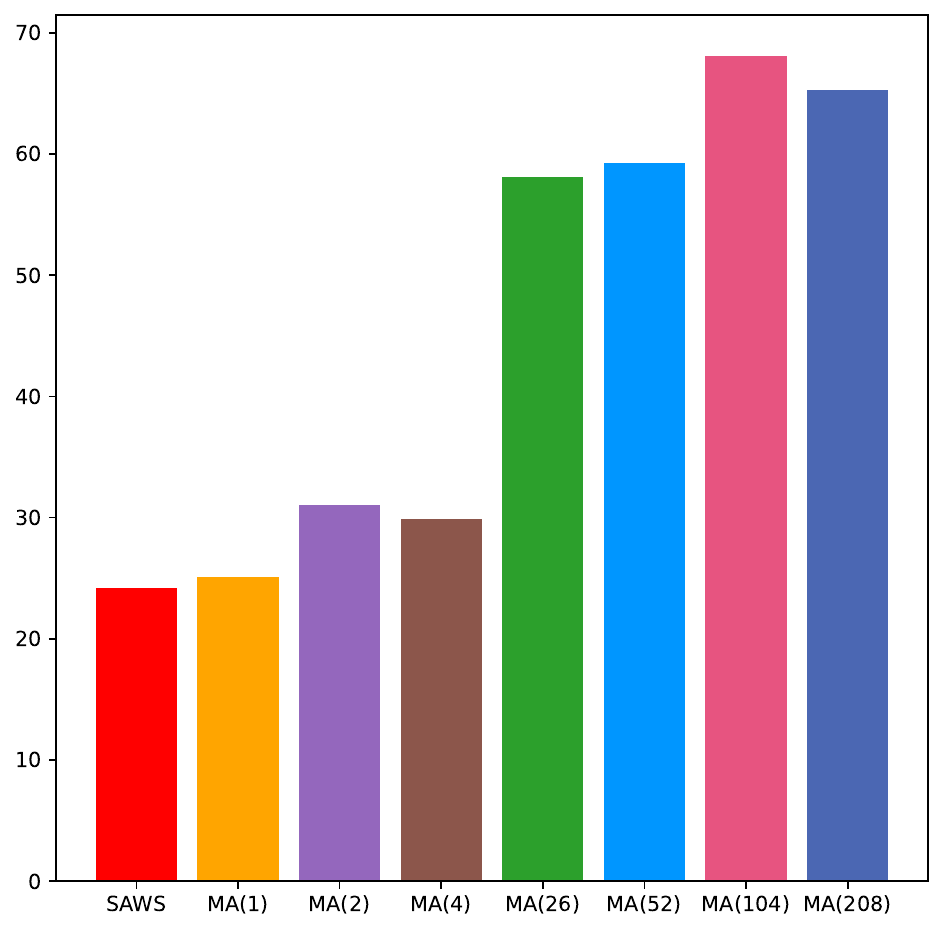}
        \caption{ED visits data. \label{fig-loss-ED}}
	\end{subfigure}
	\caption{Per-period losses of SAWS and fixed-window benchmarks on the electricity data and the ED visits data. Horizontal axis: algorithms. Vertical axis: per-period loss. For the electricity data, the predicted and true demand (unit: megawatt-hour) is scaled by $5\times 10^{-4}$. \label{fig-loss}}
\end{figure}

\begin{figure}[H]
	\centering
		\begin{subfigure}{0.48\textwidth}
    	\centering
        \includegraphics[scale=0.65]{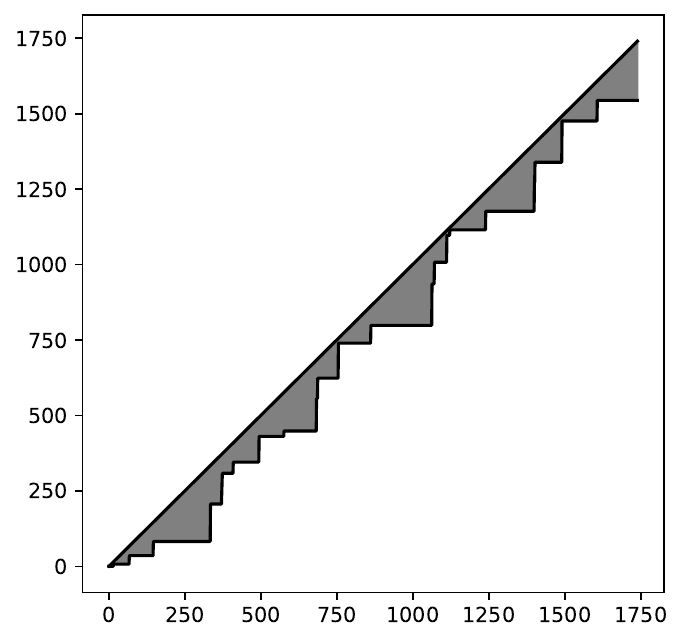}
        \caption{Electricity data. \label{fig-windows-electricity}}
	\end{subfigure}
    \begin{subfigure}{0.48\textwidth}
        \centering
        \includegraphics[scale=0.65]{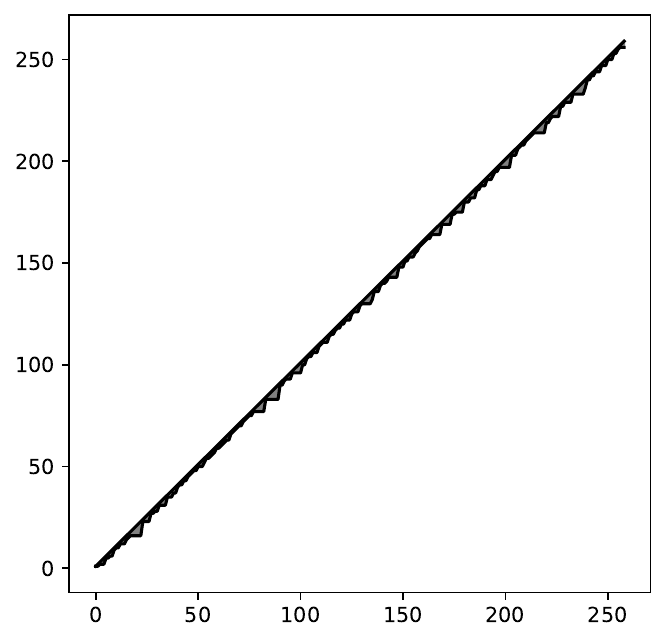}
        \caption{{ED visits data. \label{fig-windows-ED}}}
	\end{subfigure}
	\caption{Rolling windows of SAWS on the electricity data and the ED visits data. Horizontal axis: time period $n$. Vertical axis: endpoints of look-back windows. Lower black curve: left endpoints. Upper black curve: right endpoints ($n-1$). \label{fig-windows}}
\end{figure}

\Cref{fig-loss-ED} plots the per-period losses $\frac{1}{N}\sum_{n=1}^N \left[ h(\theta_n-z_n)_+ + b(z_n-\theta_n)_+ \right]$ of SAWS and $\MA(k)$. In \Cref{fig-windows-ED}, we visualize the rolling windows selected by SAWS. We observe that by adaptively varying the window size, SAWS achieves a lower loss than all fixed-window benchmarks considered.

\subsection{Summary of Experiments}

In our synthetic and real data experiments, the problem instances exhibit different patterns of non-stationarity, which lead to different optimal windows. In practice, as the non-stationarity pattern is generally unknown beforehand, it is not clear \textit{a priori} what the best window should be, or even what candidate windows to choose from. Our experiments show that without any prior knowledge of the non-stationarity, SAWS \emph{adaptively} selects look-back windows for learning, and achieves performance comparable to or even better than the best fixed-window benchmark \emph{in hindsight}.

%% file: main_discussions.tex
\section{Discussions}\label{sec-discussions}

Based on a stability principle, we developed an adaptive approach to learning under unknown non-stationarity. Our algorithm attains optimal dynamic regrets in common problems. As by-products of our analysis, we develop a novel measure of function similarity and a segmentation technique. 

A number of future directions are worth pursuing. First, we do not assume any structure of the underlying non-stationarity. In practice, some prior knowledge or forecast of the dynamics is available. Incorporating them into our method may further boost its performance. 
Second, the threshold sequence in our algorithm relies on knowledge of the function class, smoothness parameters and noise levels. It would be interesting to develop adaptive thresholds for handling these parameters. 
Third, it is also worth investigating whether our approach enjoys good theoretical guarantees with respect to other performance measures, such as the strongly adaptive regret \citep{DGS15}. Finally, an important future direction is to extend our framework to sequential decision-making problems with partial feedback, including bandit problems and reinforcement learning, where the learner only receives feedback on the chosen decisions.
This requires understanding the interplay between the non-stationarity and the exploration-exploitation tradeoff.

%% file: appendix_proof_sketch.tex
\section{Proof Sketches for Main Theorems}

In this section, we provide proof sketches for the main results, namely, \Cref{thm-excess-risk} (excess risk bound in a specific time period), \Cref{thm-regret} (general regret bound), and \Cref{thm-online-strongly-convex} (regret bound in the strongly convex case). The proof sketch for \Cref{thm-online-Lip} (regret bound in the Lipschitz case) parallels that of \Cref{thm-online-strongly-convex} and is thus omitted. For ease of exposition, we will analyze the simpler Algorithms \ref{alg-offline-simple} and \ref{alg-online-simple} instead of their more complicated counterparts Algorithms \ref{alg-offline} and \ref{alg-online}.

The key property we will use is: if $f$ and $g$ are $(\varepsilon,\delta)$-close, then for all $\btheta\in\Omega$ and $R\ge 0$,
\[
f(\btheta) - \inf_{\btheta'\in\Omega} f(\btheta') \lesssim R
\quad\text{implies}\quad
g(\btheta) - \inf_{\btheta'\in\Omega} g(\btheta') \lesssim R + \delta.
\]

\subsection{Proof Sketch for \Cref{thm-excess-risk}}\label{sec-thm-excess-risk-proof-sketch}

The full proof is given in \Cref{sec-thm-excess-risk-proof}, and uses some ideas from \cite{Mat06}. Recall that
\begin{align*}
& \bar{k} = \max \left\{ k \in [n-1] :~  F_{n - k}, F_{n - k + 1} \cdots, F_{n-1} \text{ are } ( \varepsilon,   \psi ( n , k ) )\text{-close to } F_{n-1} \right\}, \\[4pt]
& \widehat{k} = \max \left\{ k\in[n-1] : f_{n,i}(\widehat{\btheta}_{n,k}) - \inf_{\btheta \in \Omega} f_{n,i}(\btheta) \le \tau(n,i),~\forall i\in[k] \right\}.
\end{align*}

We will first prove that $\widehat{k} \ge \bar{k}$. To this end, it suffices to show that for all $i\in[\bar{k}]$, $f_{n,i}(\widehat{\btheta}_{n,\bar{k}}) - \inf_{\btheta \in \Omega} f_{n,i}(\btheta) \le \tau(n,i)$. For all $i\in[\bar{k}]$, since $F_{n-i},...,F_{n-1}$ are $(\varepsilon,\psi(n,\bar{k}))$-close to $F_{n-1}$ and $\psi(n,\bar{k}) \le \psi(n,i)$, then by Part \ref{lem-approx-average} of \Cref{lem-approx}, $F_{n,i}$ is $(\varepsilon,c\psi(n,i))$-close to $F_{n-1}$, with $c=e^{\varepsilon}+1$. Then, for all $i\in[\bar{k}]$,
\begin{align*}
&\quad 
f_{n,\bar{k}}(\widehat{\btheta}_{n,\bar{k}}) - \inf_{\btheta\in\Omega} f_{n,\bar{k}}(\btheta) = 0 \\[4pt]
\quad\Rightarrow &\quad 
F_{n,\bar{k}}(\widehat{\btheta}_{n,\bar{k}}) - \inf_{\btheta\in\Omega} F_{n,\bar{k}}(\btheta) \lesssim \psi(n,\bar{k}) \tag{$f_{n,\bar{k}}$ and $F_{n,\bar{k}}$ are $(\varepsilon,\psi(n,\bar{k}))$-close} \\[4pt]
\quad\Rightarrow &\quad 
F_{n-1}(\widehat{\btheta}_{n,\bar{k}}) - \inf_{\btheta\in\Omega} F_{n-1}(\btheta) \lesssim \psi(n,\bar{k}) \tag{$F_{n,\bar{k}}$ and $F_{n-1}$ are $(\varepsilon,(e^{\varepsilon}+1)\psi(n,\bar{k}))$-close} \\[4pt]
\quad\Rightarrow &\quad 
F_{n,i}(\widehat{\btheta}_{n,\bar{k}}) - \inf_{\btheta\in\Omega} F_{n,i}(\btheta) \lesssim \psi(n,\bar{k}) + \psi(n,i) \lesssim \psi(n,i) \tag{$F_{n,i}$ and $F_{n-1}$ are $(\varepsilon,(e^{\varepsilon}+1)\psi(n,i))$-close} \\[4pt]
\quad\Rightarrow &\quad
f_{n,i}(\widehat{\btheta}_{n,\bar{k}}) - \inf_{\btheta\in\Omega} f_{n,i}(\btheta) \lesssim \psi(n,i). \tag{$f_{n,i}$ and $F_{n,i}$ are $(\varepsilon,\psi(n,i))$-close}
\end{align*}
The condition $\tau(n,k) \ge 6e^{5\varepsilon} \psi(n,k)$ in Condition \ref{condition-thresholds} is used to ensure that the last inequality above implies $f_{n,i}(\widehat{\btheta}_{n,\bar{k}}) - \inf_{\btheta\in\Omega} f_{n,i}(\btheta) \le \tau(n,i)$. This shows that $\widehat{k} \ge \bar{k}$.

Since $\widehat{k} \ge \bar{k}$, then by the definition of $\widehat{k}$,
\begin{align*}
&\quad f_{n,\bar{k}}(\widehat{\btheta}_{n,\widehat{k}}) - \inf_{\btheta\in\Omega} f_{n,\bar{k}}(\btheta) \le \tau(n,\bar{k}) \\[4pt]
\quad\Rightarrow &\quad 
F_{n,\bar{k}}(\widehat{\btheta}_{n,\widehat{k}}) - \inf_{\btheta\in\Omega} F_{n,\bar{k}}(\btheta) \lesssim \tau(n,\bar{k}) + \psi(n,\bar{k}) \lesssim \tau(n,\bar{k}) \tag{$f_{n,\bar{k}}$ and $F_{n,\bar{k}}$ are $(\varepsilon,\psi(n,\bar{k}))$-close}\\[4pt]
\quad\Rightarrow &\quad 
F_{n-1}(\widehat{\btheta}_{n,\widehat{k}}) - \inf_{\btheta\in\Omega} F_{n-1}(\btheta) \lesssim \tau(n,\bar{k}) + \psi(n,\bar{k}) \lesssim \tau(n,\bar{k}). \tag{$F_{n,\bar{k}}$ and $F_{n-1}$ are $(\varepsilon,(e^{\varepsilon}+1)\psi(n,\bar{k}))$-close}
\end{align*}
As $\btheta_n = \widehat{\btheta}_{n,\widehat{k}}$, this finishes the proof.

\subsection{Proof Sketch for \Cref{thm-regret}}\label{sec-thm-regret-proof-sketch}

For clarity, we add a time index to the quantity $\bar{k}$ defined in \Cref{thm-excess-risk}, that is,
\[
\bar{k}_{n-1} = \max \left\{ k \in [n-1] :~  F_{n - k}, F_{n - k + 1} \cdots, F_{n-1} \text{ are } ( \varepsilon,   \psi ( n , k ) )\text{-close to } F_{n-1} \right\}.
\] 
By \Cref{def-segmentation}, if $n\in\{N_{j-1}+1,...,N_j\}$, then $F_{N_{j-1}+1},...,F_{n}$ are $(\varepsilon,\psi(n,n-N_{j-1}))$-close to $F_n$, so $\bar{k}_n \ge n-N_{j-1}$. By \Cref{thm-excess-risk} and Condition \ref{condition-thresholds},
\[
F_n(\btheta_{n+1}) - \inf_{\btheta\in\Omega} F_n(\btheta) \lesssim \tau(n+1,\bar{k}_n) \lesssim \tau(N,\bar{k}_n) \lesssim \tau(N,n-N_{j-1}).
\]
We now convert this into a bound for $F_{n+1}(\btheta_{n+1}) - \inf_{\btheta\in\Omega} F_{n+1}(\btheta)$. There are two cases.
\begin{itemize}
\item If $n\le N_j - 1$, then by \Cref{def-segmentation}, $F_n$ and $F_{n+1}$ are $(\varepsilon,\psi(n+1,n-N_{j-1}+1))$-close, so
\begin{align*}
F_{n+1}(\btheta_{n+1}) - \inf_{\btheta\in\Omega} F_{n+1}(\btheta)
&\lesssim
\tau(N,n-N_{j-1}) + \psi(n+1,n-N_{j-1}+1) \\
&\lesssim
\tau(N,n-N_{j-1}) + \tau(n+1,\bar{k}_n)
\lesssim
\tau(N,n-N_{j-1}).
\end{align*}
\item If $n=N_j$, then by \Cref{def-segmentation}, $F_n = F_{N_j}$ and $F_{n+1} = F_{N_j+1}$ are $(\varepsilon,\delta_j)$-close, so
\[
F_{n+1}(\btheta_{n+1}) - \inf_{\btheta\in\Omega} F_{n+1}(\btheta) \lesssim \tau(N,n-N_{j-1}) + \delta_j.
\] 
\end{itemize}
Moreover, $F_{n+1}(\btheta_{n+1}) - \inf_{\btheta\in\Omega}F_{n+1}(\btheta) \le U$. Therefore,
\begin{align*}
\sum_{n=2}^N \left[ F_n(\btheta_n) - \inf_{\btheta\in\Omega} F_n(\btheta) \right]
&=
\sum_{j=1}^J\sum_{n=N_{j-1}+1}^{N_j}\left[ F_{n+1}(\btheta_{n+1}) - \inf_{\btheta\in\Omega} F_{n+1}(\btheta) \right] \\[4pt]
&\lesssim
\sum_{j=1}^J \left( \sum_{n=N_{j-1}+1}^{N_j} \min \left\{ \tau(N,n-N_{j-1}) , U \right\} + \delta_j \right) \\[4pt]
&=
\sum_{j=1}^J T(N_j-N_{j-1}) + \sum_{j=1}^J \delta_j.
\end{align*}
Adding the term $F_1(\btheta_1) - \inf_{\btheta\in\Omega} F_1(\btheta)$ to both sides finishes the proof.

\subsection{Proof Sketch for \Cref{thm-online-strongly-convex}}\label{sec-thm-online-strongly-convex-proof-sketch}

For notational convenience we will drop the subscript of $J_N$. We will prove the following more refined bound: for every segmentation of $\{\btheta_n^*\}_{n=1}^N$, it holds that
\begin{equation}
\sum_{n=1}^{N} \bigg[ F_{n} ( \btheta_n ) -  F_{n} ( \btheta_n^* ) \bigg]  
\lesssim
1 + \sum_{j=1}^J \min \bigg\{   \frac{ d  }{ B} ,~ N_j - N_{j-1} \bigg\} +  \sum_{j=1}^{J}\|\btheta_{N_j+1}^*-\btheta_{N_j}^*\|_2^2 .
\label{eqn-thm-online-strongly-cvx-refined-sketch}
\end{equation}
Then \eqref{eqn-thm-online-strongly-cvx} follows from
\[
\sum_{j=1}^J \min \bigg\{   \frac{ d  }{ B} ,~ N_j - N_{j-1} \bigg\}
\le 
\min \left\{ \frac{Jd}{B}, N \right\}
\quad\text{and}\quad
\sum_{j=1}^{J}\|\btheta_{N_j+1}^*-\btheta_{N_j}^*\|_2^2 \le JM^2.
\]

To prove \eqref{eqn-thm-online-strongly-cvx-refined-sketch}, we will verify that in the strongly convex case, the segmentation in \Cref{def-segmentation-strong-cvx} translates to \Cref{def-segmentation}, and thus we can apply \Cref{thm-regret}. By a concentration bound for sub-exponential random variables (\Cref{lem-grad-uniform}), with high probability, up to logarithmic factors,
\[
\sup_{\btheta\in\Omega} \| \nabla f_{n,k}(\btheta) - \nabla F_{n,k}(\btheta) \|_2 \lesssim \max\left\{ \sqrt{\frac{d}{Bk}}, \frac{d}{Bk}\right\}.
\]
Since $F_{n,k}$ is strongly convex, then substituting the inequality above into Part \ref{lem-sufficient-grad-square} of \Cref{lem-sufficient} shows that $f_{n,k}$ and $F_{n,k}$ are $(\log 2, \eta)$-close with $\eta \asymp \frac{d}{Bk}$. Thus, we will take
\[
\psi(n,k) \asymp \frac{d}{Bk}.
\]
Moreover, by Part \ref{lem-sufficient-minimizers} of \Cref{lem-sufficient}, $F_{n}$ and $F_i$ are $(\log(4L/\rho), \frac{\rho}{2} \|\btheta_n^* - \btheta_i^*\|_2^2 )$-close. 

Let $\{\btheta_n^*\}_{n=1}^N$ be segmented as in \Cref{def-segmentation-strong-cvx}. Then for every $j\in[J]$,
\[
\max_{N_{j - 1} < i, k \leq N_j } \| \btheta_i^* - \btheta_k^* \|_2^2
\lesssim
\frac{d}{B ( N_j - N_{j - 1} ) } \asymp \min_{N_{j-1}<n\le N_j} \psi(n,n-N_{j-1}),
\]
and thus $F_{N_{j-1}+1},...,F_{N_j}$ are $(\log(4L/\rho),\min_{N_{j-1}<n\le N_j} \psi(n,n-N_{j-1}))$-close. In addition, $F_{N_j}$ and $F_{N_j+1}$ are $(\log(4L/\rho),\delta_j)$-close with $\delta_j = \frac{\rho}{2}\|\btheta_{N_j+1}^* - \btheta_{N_j}^*\|_2^2$. This shows that the segmentation in \Cref{def-segmentation-strong-cvx} is also a segmentation in the sense of \Cref{def-segmentation}. Therefore, \Cref{thm-regret} is applicable, and yields
\begin{align*}
\sum_{n=1}^{N} \left[ F_n ( \btheta_n ) - \inf_{ \btheta_n' \in \Omega } F_n ( \btheta_n' ) \right]
&\lesssim
\left[ F_1 ( \btheta_1 ) - \inf_{ \btheta \in \Omega } F_1 ( \btheta ) \right]
+
\sum_{j = 1}^J T( N_j - N_{j - 1} ) 
+ \sum_{j=1}^{J} \delta_{j} \\[4pt]
&\lesssim
1 + \sum_{j=1}^J\sum_{n=N_{j-1}+1}^{N_j} \min \{ \tau(N,n-N_{j-1}), 1\} +  \sum_{j=1}^{J}\|\btheta_{N_j+1}^* - \btheta_{N_j}^*\|_2^2 \\[4pt]
&\lesssim
1 + \sum_{j=1}^J\sum_{n=N_{j-1}+1}^{N_j} \min \left\{ \frac{d}{B(n-N_{j-1})}, 1 \right\} +  \sum_{j=1}^{J}\|\btheta_{N_j+1}^* - \btheta_{N_j}^*\|_2^2 \\[4pt]
&\lesssim
1 + \sum_{j=1}^J\min\left\{ \frac{d}{B},N_j-N_{j-1} \right\} + \sum_{j=1}^{J}\|\btheta_{N_j+1}^* - \btheta_{N_j}^*\|_2^2.
\end{align*}
This completes the proof.

%% file: appendix_proof_general.tex
\section{Proofs for \Cref{sec-theory-general}}

\subsection{Proof of \Cref{lem-approx}}\label{sec-lem-approx-proof}

The claims in Parts \ref{lem-approx-self}, \ref{lem-approx-monotonicity}, \ref{lem-approx-shift}, \ref{lem-approx-symmetry} and \ref{lem-approx-general} are obviously true. To prove Part \ref{lem-approx-transitivity}, for all $\btheta\in\Omega$,
\begin{align*}
f(\btheta)-\inf_{\btheta'\in\Omega}f(\btheta')
\le 
e^{\varepsilon_1}\left(g(\btheta)-\inf_{\btheta'\in\Omega}g(\btheta')+\delta_1\right)
&\le 
e^{\varepsilon_1}\left[e^{\varepsilon_2}\left(h(\btheta)-\inf_{\btheta'\in\Omega}h(\btheta')+\delta_2\right)+\delta_1\right] \\[4pt]
&\le 
e^{\varepsilon_1+\varepsilon_2}\left(h(\btheta)-\inf_{\btheta'\in\Omega}h(\btheta')+\delta_1+\delta_2\right).
\end{align*}
By Part \ref{lem-approx-symmetry}, $h$ and $g$ are $(\varepsilon_2, \delta_2)$-close, $g$ and $f$ are $(\varepsilon_1, \delta_1)$-close. By the same argument above,
\[
h(\btheta)-\inf_{\btheta'\in\Omega}h(\btheta')
\le 
e^{\varepsilon_1+\varepsilon_2}\left(f(\btheta)-\inf_{\btheta'\in\Omega}f(\btheta')+\delta_1+\delta_2\right).
\]
This shows that $f$ and $h$ are $(\varepsilon_1+\varepsilon_2,\delta_1+\delta_2)$-close.

Finally, we prove Part \ref{lem-approx-average}. Let $f_i^* = \inf_{\btheta'\in\Omega} f_i(\btheta')$ and $g^* = \inf_{\btheta'\in\Omega} g(\btheta')$. By assumption, it holds for all $i\in[m]$ and $\btheta\in\Omega$ that
\begin{align}
& g(\btheta) - g^* \le e^{\varepsilon} \big( f_i(\btheta) - f_i^* + \delta \big), \label{eqn-lem-approx-1} \\[4pt]
& f_i(\btheta) - f_i^* \le e^{\varepsilon} \big( g(\btheta) - g^* + \delta \big). \label{eqn-lem-approx-2}
\end{align}
Let $f = \sum_{i=1}^m \lambda_i f_i$. Multiplying both \eqref{eqn-lem-approx-1} and \eqref{eqn-lem-approx-2} by $\lambda_i$ and summing over $i\in[m]$ yields
\begin{align}
& g(\btheta) - g^* \le e^{\varepsilon} \left( f(\btheta) - \sum_{i=1}^n \lambda_i f_i^* + \delta \right), \label{eqn-lem-approx-3} \\[4pt]
& f(\btheta) - \sum_{i=1}^n \lambda_i f_i^* \le e^{\varepsilon} \big( g(\btheta) - g^* + \delta \big). \label{eqn-lem-approx-4}
\end{align}
We have
\begin{equation}
\sum_{i=1}^n \lambda_i f_i^*
\le \inf_{\btheta'} f(\btheta') 
\le \inf_{\btheta'\in\Omega} \sum_{i=1}^n \lambda_i \left[ f_i^* + e^{\varepsilon} \big( g(\btheta') - g^* + \delta \big) \right]
=
\sum_{i=1}^n \lambda_i f_i^* + e^{\varepsilon} \delta,
\label{eqn-lem-approx-5}
\end{equation}
where the second inequality is due to \eqref{eqn-lem-approx-2}. Substituting \eqref{eqn-lem-approx-5} into \eqref{eqn-lem-approx-3} and \eqref{eqn-lem-approx-4} gives
\begin{align*}
& g(\btheta) - g^* \le e^{\varepsilon} \left( f(\btheta) - \sum_{i=1}^n \lambda_i f_i^* + \delta \right) \le e^{\varepsilon} \left( f(\btheta) - \inf_{\btheta'} f(\btheta') + e^{\varepsilon}\delta + \delta  \right),  \\[4pt]
& f(\btheta) - \inf_{\btheta'} f(\btheta') \le f(\btheta) - \sum_{i=1}^n \lambda_i f_i^* \le e^{\varepsilon} \big( g(\btheta) - g^* + \delta \big). 
\end{align*}
This shows that $f$ and $g$ are $\big(\varepsilon,(e^{\varepsilon}+1)\delta\big)$-close.

\subsection{Proof of \Cref{lem-sufficient}}\label{sec-lem-sufficient-proof}

Part of the proof uses \Cref{lem-approx}.

\noindent{\bf Part \ref{lem-sufficient-sup}.} Thanks to Part \ref{lem-approx-shift} in \Cref{lem-approx}, it suffices to work under the additional assumption $c = 0$.

The function $f$ is clearly lower bounded. 
Define $f^* = \inf_{ \btheta \in \Omega } f ( \btheta )$ and $g^* = \inf_{ \btheta \in \Omega } g ( \btheta )$. Without loss of generality, assume $f^*\ge g^*$. For every $\btheta\in\Omega$,
\[
f^*-g^*
=
[f^*-f(\btheta)] + [f(\btheta)-g(\btheta)] + [g(\btheta)-g^*]
\le 
D_0+[g(\btheta)-g^*].
\]
Taking infimum over all $\btheta\in\Omega$ yields $|f^*-g^*|\le D_0$. Therefore, for all $\btheta\in\Omega$,
\[
|[f(\btheta)-f^*] - [g(\btheta)-g^*]|
\le 
|f(\btheta)-g(\btheta)| + |f^*-g^*|
\le 
2D_0.
\]
This implies that $f$ and $g$ are $(0,2D_0)$-close.

\vspace{1em}

\noindent{\bf Part \ref{lem-sufficient-grad-sup}.} The bounds on the supremum of $ \| \nabla f  - \nabla g  \|_2 $ and the diameter of $\Omega$ imply the existence of a constant $c$ such that
\begin{align}
& \sup_{  \btheta \in \Omega } | f  (\btheta) - g ( \btheta ) - c |  \leq D_1 M .
\label{eqn-lem-sufficient-1}
\end{align}
Then, the desired result follows from Part \ref{lem-sufficient-sup}.

\vspace{1em}

\noindent{\bf Part \ref{lem-sufficient-grad-square}.} By the assumptions, $f$ has at least one minimizer $\btheta_f^* \in \Omega$, $g$ has a unique minimizer $\btheta_g^* \in \Omega$, and
\begin{align}
& g (\btheta ) - g (\btheta_g^*) \geq  \frac{\rho}{2} \| \btheta - \btheta_g^* \|_2^2  
,\qquad \forall \btheta \in \Omega 
 .\label{eqn-lem-sufficient-2} 
\end{align}
Therefore,
\begin{align}
\| \btheta - \btheta_g^* \|_2 
\leq \sqrt{ 2 \rho^{-1} [ g (\btheta ) - g (\btheta_g^*) ] } 
\leq \frac{
	g (\btheta ) - g (\btheta_g^*)
}{2 D_1}   + \frac{D_1}{ \rho } .
\label{eqn-lem-sufficient-3} 
\end{align}

By the definition of $D_1$ and \eqref{eqn-lem-sufficient-3},
\[
g (\btheta ) - g (\btheta_g^*)
\le 
f (\btheta) - f ( \btheta_g^*) + D_1 \| \btheta - \btheta_g^* \|_2 
\le 
f (\btheta) - f ( \btheta_g^*) + \frac{g (\btheta ) - g (\btheta_g^*)
}{2}   + \frac{D_1^2}{ \rho },
\]
which implies
\begin{align}
g (\btheta ) - g (\btheta_g^*) \leq 2 \bigg(	f(\btheta) - f ( \btheta_f^*) + \frac{D_1^2}{ \rho } \bigg)	.\label{eqn-lem-sufficient-4}
\end{align}

Moreover,
\begin{align*}
f(\btheta) - f ( \btheta_f^*)
& \leq g (\btheta) - g ( \btheta_f^*) + D_1 \| \btheta - \btheta_f^* \|_2 \notag\\
& \leq g (\btheta) - g ( \btheta_g^*) + D_1 \| \btheta - \btheta_g^* \|_2 + D_1 \| \btheta_g^* - \btheta_f^* \|_2.
\end{align*}
By \eqref{eqn-lem-sufficient-3} again,
\begin{align}
f(\btheta) - f ( \btheta_f^*)
& \leq 
\frac{3}{2}
[ g (\btheta) - g ( \btheta_g^*) ] + 
\frac{D_1^2}{\rho}
 + D_1 \| \btheta_g^* - \btheta_f^* \|_2.
	\label{eqn-lem-sufficient-5}
\end{align}
To bound $\| \btheta_g^* - \btheta_f^* \|_2$, by \eqref{eqn-lem-sufficient-2},
\[
\frac{\rho}{2} \| \btheta_f^* - \btheta_g^* \|_2^2  \leq
g (\btheta_f^* ) - g (\btheta_g^*) \leq  
f ( \btheta_f^* ) - f (\btheta_g^*) + D_1 \| \btheta_f^* - \btheta_g^* \|_2 \leq 
D_1 \| \btheta_f^* - \btheta_g^* \|_2.
\]
Hence, $\| \btheta_g^* - \btheta_f^* \|_2 \leq 2 D_1 / \rho$. Substituting it into \eqref{eqn-lem-sufficient-5} yields
\begin{align}
f(\btheta) - f ( \btheta_f^*)
& \leq 
\frac{3}{2}
[ g (\btheta) - g ( \btheta_g^*) ] + 
\frac{3 D_1^2}{\rho}
= \frac{3}{2}
\bigg(
 g (\btheta) - g ( \btheta_g^*)  + 
\frac{2 D_1^2}{\rho}
\bigg)
.
\label{eqn-lem-sufficient-6}
\end{align}
According to \eqref{eqn-lem-sufficient-4} and \eqref{eqn-lem-sufficient-6}, $f$ and $g$ are $(\log 2, 2 D_1^2/\rho)$-close. On the other hand, Part \ref{lem-sufficient-grad-sup} implies that $f$ and $g$ are always $( \log 2, 2 M D_1   )$-close. Combining the two results finishes the proof.

\vspace{1em}

\noindent{\bf Part \ref{lem-sufficient-minimizers}.} More generally, we will prove that if $f$ and $g$ are $\rho$-strongly convex and $L$-smooth over $\Omega$, then $f$ and $g$ are $(\log(4L/\rho), \delta)$-close with 
\[
\delta = \frac{\rho}{2} \|  \btheta_{f}^* - \btheta_{g}^*  \|_2^2
+ \frac{\rho}{4L^2}
\| \nabla {f}(\btheta_{f}^*) - \nabla {g}(\btheta_{g}^*) \|_2^2.
\] 
When $\btheta_f^*$ and $\btheta_g^*$ are interior points of $\Omega$, we have $\nabla {f}(\btheta_{f}^*) = \nabla {g}(\btheta_{g}^*) = \bm{0}$ and Part \ref{lem-sufficient-minimizers} follows. 

By strong convexity, both $f$ and $g$ have unique minimizers $\btheta_f^*$ and $\btheta_g^*$, respectively. We have \eqref{eqn-lem-sufficient-2} and
\begin{align}
& f(\btheta ) - f(\btheta_f^*)
\leq \langle \nabla f(\btheta_f^*) , \btheta - \btheta_f^*  \rangle
+ \frac{L}{2} \| \btheta - \btheta_{f}^* \|_2^2 .
\label{eqn-lem-sufficient-10}
\end{align}

We start by working on $\langle \nabla f(\btheta_f^*), \btheta - \btheta_f^*  \rangle$. Note that
\begin{align*}
	\langle \nabla {f}(\btheta_{f}^*) , \btheta - \btheta_{f}^* \rangle
	&  = \langle \nabla {f}(\btheta_{f}^*) - \nabla {g}(\btheta_{g}^*)  , \btheta - \btheta_{f}^* \rangle
	+ \langle \nabla {g}(\btheta_{g}^*) , \btheta - \btheta_{f}^* \rangle \notag\\
	& = \langle  \nabla {f}(\btheta_{f}^*) - \nabla {g}(\btheta_{g}^*) , \btheta - \btheta_{f}^* \rangle
	+ \langle \nabla {g}(\btheta_{g}^*) , \btheta - \btheta_{g}^*  \rangle 
	- \langle \nabla {g}(\btheta_{g}^*) , \btheta_{f}^* - \btheta_{g}^* \rangle .
\end{align*}
We have
\begin{align*}
	&  \langle  \nabla {f}(\btheta_{f}^*) - \nabla {g}(\btheta_{g}^*) , \btheta - \btheta_{f}^* \rangle
	\leq  \frac{
		(L/2)
		\|  \btheta - \btheta_{f}^* \|_2^2 + (2/L) \| \nabla {f}(\btheta_{f}^*) - \nabla {g}(\btheta_{g}^*) \|_2^2 
	}{2} ,\\
	&  \langle  \nabla {g}(\btheta_{g}^*) , \btheta - \btheta_{g}^* \rangle 
	\leq 
	g (\btheta ) - g (\btheta_{g}^*) .
\end{align*}
By the optimality of $\btheta_g^*$, we have $\langle \nabla {g}(\btheta_{g}^*) , \btheta_{f}^* - \btheta_{g}^* \rangle \ge 0$. Combining the estimates above yields
\begin{align*}
\langle \nabla {f}(\btheta_{f}^*) , \btheta - \btheta_{f}^* \rangle
&\leq 
\frac{L}{4}	\|  \btheta - \btheta_{f}^* \|_2^2 + \frac{1}{L} \| \nabla {f}(\btheta_{f}^*) - \nabla {g}(\btheta_{g}^*) \|_2^2 
+ [g (\btheta ) - g (\btheta_{g}^*) ].
\end{align*}
Plugging this into \eqref{eqn-lem-sufficient-10}, we get
\[ f(\btheta ) - f(\btheta_f^*)
	\leq  [g (\btheta ) - g (\btheta_{g}^*)]
	+ \frac{3L}{4} \| \btheta - \btheta_{f}^* \|_2^2 
	+
	\frac{
		\| \nabla {f}(\btheta_{f}^*) - \nabla {g}(\btheta_{g}^*) \|_2^2 
	}{L}	.
\]

It remains to control $ \| \btheta - \btheta_{f}^* \|_2^2 $. By elementary inequalities and \eqref{eqn-lem-sufficient-2},
\begin{align*}
\| \btheta - \btheta_{f}^* \|_2^2 
& \leq  (
\| \btheta - \btheta_g^* \|_2
+ \| \btheta_g^* - \btheta_f^* \|_2
 )^2 
\leq 
2
\| \btheta - \btheta_g^* \|_2^2
+ 
2
\| \btheta_g^* - \btheta_f^* \|_2^2 \\
& \leq \frac{4}{\rho} [g (\btheta ) - g (\btheta_g^*) ] + 2
\| \btheta_g^* - \btheta_f^* \|_2^2 .
\end{align*}
Therefore,
\begin{align*}
 f(\btheta ) - f(\btheta_f^*)
&	\leq \frac{4 L}{\rho} [g (\btheta ) - g (\btheta_{g}^*)]
	+
	\bigg(
\frac{3}{2} L \|  \btheta_{f}^* - \btheta_{g}^*  \|_2^2
	+ L^{-1}
		\| \nabla {f}(\btheta_{f}^*) - \nabla {g}(\btheta_{g}^*) \|_2^2 
	\bigg) \\
& \le \frac{4 L}{\rho} \bigg[
g (\btheta ) - g (\btheta_{g}^*)
+
\bigg(
\frac{\rho}{2} \|  \btheta_{f}^* - \btheta_{g}^*  \|_2^2
+ \frac{\rho}{4L^2}
\| \nabla {f}(\btheta_{f}^*) - \nabla {g}(\btheta_{g}^*) \|_2^2 
\bigg)
\bigg]	.
\end{align*}
By symmetry, the inequality continues to hold after swapping $f$ and $g$. This completes the proof.

\subsection{Proof of \Cref{thm-excess-risk}}\label{sec-thm-excess-risk-proof}

We prove the following extended version of \Cref{thm-excess-risk}. Part \ref{thm-excess-risk-Part-2} will be useful for proving \Cref{thm-regret}.

\begin{theorem}[Excess risk bound]\label{thm-excess-risk-general}
Fix $n\in[N]$. Consider \Cref{alg-offline} as a subroutine of \Cref{alg-online}, with $k_{s+1}\le 2k_s$ for each $s\in[m-1]$. Let Assumption \ref{assumption-approximation} and Condition \ref{condition-thresholds} hold. Define
\begin{align*}
\bar{k} = \max \{ k \in [n-1] :~  F_{n - k}, F_{n - k + 1} \cdots, F_{n-1} \text{ are } ( \varepsilon,   \psi ( n , k ) )\text{-close to } F_{n-1} \}.
\end{align*}
Let $\btheta_{n}$ be the output of \Cref{alg-offline}, and let $\widehat{s}$ be the corresponding window index. 
\begin{enumerate}
\item\label{thm-excess-risk-Part-1} It holds that $\displaystyle F_{n-1} ( \btheta_{ n } ) - \inf_{ \btheta \in \Omega } F_{n-1} ( \btheta  )  \leq 2 e^{2 \varepsilon} C \tau( n , \bar{k} \wedge k_m  )$.

\item\label{thm-excess-risk-Part-2}  If $k_m \leq \bar{k}$, then $ \widehat{s}  = m$. If $k_m > \bar{k}$, then $\widehat{s} = m$ or $k_{ \widehat{s} + 1 } > \bar{k}$. 
\end{enumerate}
\end{theorem}

The proof borrows some ideas from \cite{Mat06}. First, we invoke a useful lemma.

\begin{lemma}\label{lem-excess-risk}
	Consider the setting of \Cref{thm-excess-risk}.
	Define
	\[
	\bar{s}  = \max \{ s \in [m] :~  F_{ n - k_s}, F_{ n - k_s + 1} , \cdots, F_{ n-1 } \text{ are } ( \varepsilon,  \psi ( n , k_s ) )\text{-close to }F_{n-1} \}.
	\]
	Then, we have $\widehat{s} \geq \bar{s}$ and $
	F_{n-1} ( \btheta_{ n } ) - \inf_{ \btheta \in \Omega } F_{n-1} ( \btheta  )  \leq 2 e^{2 \varepsilon} \tau( n  ,  k_{\bar{s}}  )$ .
\end{lemma}

\begin{proof}[\bf Proof of \Cref{lem-excess-risk}] 
By definition,
	\[
	\widehat{s} = \max \left\{ s\in[m] : f_{n,k_i}(\widehat{\btheta}_{n,k_s}) -  f_{n,k_i}(\widehat{\btheta}_{n,k_i}) \le \tau(n,k_i),\ \forall i\in[s] \right\}.
	\]
	To prove that $\widehat{s} \ge \bar{s}$, it suffices to show that $f_{n,k_i}(\widehat{\btheta}_{n,k_{\bar{s}}}) -  f_{n,k_i}(\widehat{\btheta}_{n,k_i}) \le \tau(n,k_i)$ for all $i\in[\bar{s}]$. Take arbitrary $i \in [ \bar{s}]$. By Assumption \ref{assumption-approximation}, $k_i \le k_{\bar{s}}$ implies $\psi(n,k_i) \ge \psi(n,k_{\bar{s}})$.
	By Part \ref{lem-approx-average} of \Cref{lem-approx}, $F_{n, k_i}$ and $F_{n-1}$ are $ ( \varepsilon, (e^{\varepsilon} + 1) \psi( n, k_i ) )$-close. Assumption \ref{assumption-approximation} states that $f_{n, k_i}$ and $F_{n,k_i}$ are $(\varepsilon , \psi( n, k_i) )$-close. By Parts \ref{lem-approx-monotonicity} and \ref{lem-approx-transitivity} in \Cref{lem-approx}, $f_{n, k_i}$ and $F_{n-1}$ are $( 2 \varepsilon ,  3 e^{\varepsilon}  \psi ( n, k_i) )$-close. In particular, $f_{n, k_{\bar{s}}}$ and $F_{n-1}$ are $( 2 \varepsilon , 3 e^{\varepsilon}   \psi ( n,  k_{\bar{s}} ) )$-close. By Part \ref{lem-approx-transitivity} in \Cref{lem-approx}, $f_{n, k_{\bar{s}}}$ and $f_{n, k_i}$ are $( 4 \varepsilon , 6 e^{\varepsilon}  \psi ( n, k_i) )$-close. Therefore,
	\begin{align}
	f_{n, k_i} ( \widehat\btheta_{n, k_{\bar{s}}} ) -  f_{n, k_i} ( \widehat\btheta_{n, k_i} ) 
	& = f_{n, k_i} ( \widehat\btheta_{n, k_{\bar{s}}} ) - \inf_{ \btheta \in \Omega }  f_{n, k_i} (\btheta) \notag \\[4pt]
	&\le e^{4\varepsilon}
	\bigg( 
	f_{n,  k_{\bar{s}}} ( \widehat\btheta_{n,  k_{\bar{s}}} ) - \inf_{ \btheta \in \Omega }  f_{n,  k_{\bar{s}}} (\btheta) 
+  6  e^{\varepsilon}   \psi ( n, k_i)
	\bigg) \notag\\[4pt]
	& = 6e^{5\varepsilon} \psi(n,k_i) \le \tau(n,k_i).
	\label{eqn-thm-excess-risk-proof-1}
	\end{align}
	Since \eqref{eqn-thm-excess-risk-proof-1} holds for all $i \in [\bar{s}]$, then by the definition of $\widehat{s} $, we have $\widehat{s}  \ge \bar{s}$. This implies
	\begin{equation}
	f_{n,  k_{\bar{s}}} ( \widehat\btheta_{n , k_{\widehat{s}} } ) - \inf_{\btheta\in\Omega} f_{n,k_{\bar{s}}}(\btheta)
	\le 
	\tau(n,k_{\bar{s}}).
	\label{eqn-thm-excess-risk-proof-2}
	\end{equation}
	
Recall that $f_{n, k_{\bar{s}}}$ and $F_{n-1}$ are $( 2 \varepsilon , 3 e^{\varepsilon}   \psi  ( n,   k_{\bar{s}}) )$-close. By Condition \ref{condition-thresholds}, $3 e^{\varepsilon}   \psi  ( n,   k_{\bar{s}}) \le \tau ( n , k_{\bar{s}} )$.
Then, by \eqref{eqn-thm-excess-risk-proof-2},
\begin{align*}
F_{n-1}(\widehat{\btheta}_{n,k_{\widehat{s}}})-\inf_{\btheta\in\Omega}F_{n-1}(\btheta)
&\le 
e^{2\varepsilon}\left(f_{n,k_{\bar{s}}}(\widehat{\btheta}_{n,k_{\widehat{s}}})-\inf_{\btheta\in\Omega}f_{n,k_{\bar{s}}}(\btheta)+3 e^{\varepsilon}   \psi  ( n,   k_{\bar{s}}) \right) \\[4pt]
&\le 
e^{2\varepsilon}\big( \tau(n,k_{\bar{s}}) + 3 e^{\varepsilon}   \psi ( n, k_{\bar{s}}) \big) \\[4pt]
&=
2e^{2\varepsilon}\tau(n,k_{\bar{s}}).
\end{align*}
This finishes the proof.
\end{proof}

If $\bar{k} \geq k_m$, then the definitions of $\bar{k}$ and $\bar{s}$ imply that $\bar{s} = m$. 
By \Cref{lem-excess-risk},
\begin{align*}
	F_{n-1} ( \btheta_{n} ) - \inf_{ \btheta \in \Omega } F_{n-1} (\btheta) 
	\leq 
	2 e^{2 \varepsilon} \tau( n , k_{ m } )  
	\leq 	2 e^{2 \varepsilon} C \tau( n , \bar{k} \wedge  k_{ m }  )  
	.
\end{align*}
Since $m \geq \widehat{s} \geq \bar{s}$, we have $\widehat{s} = m$ and $k_{\widehat{s}} = k_m$.

Finally, consider the case $\bar{k} < k_m $. If $\widehat{s} = m$, nothing needs to be done. Suppose that $\widehat{s} < m$. Then, $\bar{s} + 1 \leq \widehat{s} + 1 \leq m$. The definitions of $\bar{k}$ and $\bar{s}$ imply that $\bar{k} < k_{ \bar{s} + 1 } \leq k_m$. Then, $k_{ \widehat{s} + 1 } \geq k_{ \bar{s} + 1 } > \bar{k}$. Meanwhile, Condition \ref{condition-thresholds} gives
\[
\tau( n , k_{ \bar{s} } ) \leq C \tau( n, k_{ \bar{s} + 1 } ) \leq C \tau ( n , \bar{k} ) .
\]
By this and \Cref{lem-excess-risk},
\begin{align*}
	F_{n-1} ( \btheta_{n} ) - \inf_{ \btheta \in \Omega } F_{n-1} (\btheta) 
	\leq 	2 e^{2 \varepsilon} C \tau( n , \bar{k}   )  
	=  	2 e^{2 \varepsilon} C \tau( n , \bar{k} \wedge  k_{ m }  )  
	.
\end{align*}
This proves \Cref{thm-excess-risk-general}.

\subsection{Proof of \Cref{thm-regret}}\label{sec-thm-regret-proof}

We start with a useful lemma.

\begin{lemma}\label{lem-window}
Choose any $j \in [J]$ and $n \in \{ N_{j - 1} + 1  , \cdots,  N_j  \}$. For the $n$-th iteration of \Cref{alg-online},
\begin{enumerate}
\item\label{lem-window-kbar} the quantity $\bar{k}$ defined in \Cref{thm-excess-risk} satisfies $\bar{k} \geq n - N_{j-1}$;
\item\label{lem-window-selected} $k_m \geq K_n \geq ( n - N_{j-1} + 1 ) / 2$.
\end{enumerate}
\end{lemma}

\begin{proof}[\bf Proof of \Cref{lem-window}]

\noindent{\bf Part \ref{lem-window-kbar}.} \Cref{def-segmentation} and Part \ref{lem-approx-monotonicity} of \Cref{lem-approx} imply that $F_{N_{j-1} + 1} , \cdots , F_{n}$ are $(\varepsilon , \psi ( n, n - N_{j-1} ))$-close to $F_{n}$. Therefore, the quantity $\bar{k}$ defined in \Cref{thm-excess-risk} satisfies $\bar{k} \geq n - N_{j-1}$.

\vspace{1em}
\noindent{\bf Part \ref{lem-window-selected}.} 
By definition, $K_n = k_{\widehat{s}} \le k_m$. It suffices to prove $ K_n \geq ( n - N_{j-1} + 1 ) / 2$. Let $n_j = N_j - N_{j-1}$. When $n_j = 1$, nothing needs to be proved. Suppose that $n_j \geq 2$. The base case $n = N_{j-1} + 1$ is trivial. 

Suppose that $1 \leq r < n_j$ and $K_n \geq (n - N_{j-1}+1)/2$ holds for $n = N_{j-1} + r$. That is, $K_{N_{j-1} + r} \geq (r+1)/2$.
Consider the case $n = N_{j-1} + r + 1$. We need to show that $K_n \geq (r+2)/2$.

In the $n$-th iteration of \Cref{alg-online}, the maximum candidate window is $k_{m} = K_{N_{j - 1} + r} + 1$. Hence,
\begin{align}
k_m \geq \frac{r+1}{2} + 1 = \frac{r + 3}{2}.
\label{eqn-lem-window-selected-1}
\end{align}
If $ k_{m} \leq \bar{k} $, then Part \ref{thm-excess-risk-Part-2} of \Cref{thm-excess-risk-general} implies that $\widehat{s} = m$. By \eqref{eqn-lem-window-selected-1}, $K_{n} = k_m \geq (r+3)/2$. 
If $k_m > \bar{k}$, then Part \ref{thm-excess-risk-Part-2} of \Cref{thm-excess-risk-general} implies that $\widehat{s} = m$ or $k_{\widehat{s} + 1} > \bar{k}$. 
\begin{itemize}
\item In the first case, \eqref{eqn-lem-window-selected-1} leads to $K_{n} = k_m \geq (r+3)/2$. 
\item In the second case, we use Part \ref{lem-window-kbar} to get $\bar{k} \geq n - N_{j-1} = r + 1$. Then, $k_{\widehat{s} + 1} \geq r + 2$. By the construction of $\{ k_s \}_{s=1}^m$,
\[
K_n = k_{\widehat{s}} \geq \frac{ k_{\widehat{s} + 1}  }{2} \geq \frac{r + 2}{2}.
\]
\end{itemize}
Hence, $K_n \geq (r+2)/2$. The proof is finished by induction. 
\end{proof}

We come back to the proof of \Cref{thm-regret}. Choose any $j \in [J]$ and $n \in \{ N_{j - 1} + 1  , \cdots,  N_j  \}$.
By \Cref{thm-excess-risk},
\[
F_n ( \btheta_{ n+1  } ) - \inf_{ \btheta \in \Omega } F_n ( \btheta  )  \leq 2 e^{2 \varepsilon} C \tau( n + 1 , \bar{k} \wedge k_m  ) .
\]
\Cref{lem-window} implies that $ n - N_{j-1} \leq 2 ( \bar{k} \wedge k_m )$. By Condition \ref{condition-thresholds},
\[
\tau( n + 1 , \bar{k} \wedge k_m  ) \leq \tau( N , \bar{k} \wedge k_m  ) \leq C \tau\left( N , 2 ( \bar{k} \wedge k_m ) \wedge (N-1)  \right) \leq 
C \tau( N , n - N_{j-1}  ).
\]
Consequently,
\begin{equation}
F_n ( \btheta_{ n + 1 } ) - \inf_{ \btheta \in \Omega } F_n ( \btheta  )  \le 2 e^{2 \varepsilon} C^2 \tau( N , n - N_{j-1} ) .
\label{eqn-thm-regret-1}
\end{equation}

To complete the proof, we now use \eqref{eqn-thm-regret-1} to derive a bound for $F_{n+1} ( \btheta_{ n + 1 } ) - \inf_{ \btheta \in \Omega } F_{n+1} ( \btheta  )$. There are two cases.
\begin{enumerate}
\item If $n\le N_j-1$, then by \Cref{def-segmentation}, $F_n$ and $F_{n+1}$ are $\left(\varepsilon,\psi(n+1,n-N_{j-1}+1) \right)$-close. Then by Condition \ref{condition-thresholds},
\begin{align*}
\psi(n+1,n-N_{j-1}+1)
\le 
\tau(n+1,n-N_{j-1}+1)
\le 
\tau(N,n-N_{j-1}+1)
\le
\tau(N,n-N_{j-1}),
\end{align*}
so $F_n$ and $F_{n+1}$ are $\left(\varepsilon,\tau(N,n-N_{j-1}) \right)$-close. This implies
\begin{align*}
F_{n+1}(\btheta_{n+1}) - \inf_{\btheta\in\Omega} F_{n+1}(\btheta)
&\le 
e^{\varepsilon} \left( F_n(\btheta_{n+1}) - \inf_{\btheta\in\Omega} F_n(\btheta) + \tau(N,n-N_{j-1}) \right) \\[4pt]
&\le 
e^{\varepsilon} \left( 2 e^{2 \varepsilon} C^2 \tau( N , n - N_{j-1} ) + \tau(N,n-N_{j-1}) \right) \\[4pt]
&\le 
3e^{3\varepsilon} C^2 \tau(N,n-N_{j-1}).
\end{align*}
By the definition of $U$ and the fact that $C\ge 1$,
\begin{align}
F_{n+1}(\btheta_{n+1}) - \inf_{\btheta\in\Omega} F_{n+1}(\btheta)
&\le 
\min\left\{3e^{3\varepsilon} C^2 \tau(N,n-N_{j-1}) , U\right\} \notag \\[4pt]
&\le 
3e^{3\varepsilon} C^2 \min\left\{\tau(N,n-N_{j-1}) , U\right\}.
\label{eqn-thm-regret-case-1}
\end{align}
\item If $n=N_j$, then by \Cref{def-segmentation}, $F_n=F_{N_j}$ and $F_{n+1}=F_{N_j+1}$ are $(\varepsilon,\delta_j)$-close, so
\begin{align}
F_{n+1}(\btheta_{n+1}) - \inf_{\btheta\in\Omega} F_{n+1}(\btheta)
&\le 
e^{\varepsilon} \left( F_n(\btheta_{n+1}) - \inf_{\btheta\in\Omega} F_n(\btheta) + \delta_j \right) \notag\\[4pt]
&\le 
e^{\varepsilon} \min\left\{ 2e^{2\varepsilon}C^2\tau(N,n-N_{j-1}), U \right\} + e^{\varepsilon}\delta_j \notag\\[4pt]
&\le 
2e^{3\varepsilon} C^2 \min\left\{\tau(N,n-N_{j-1}) , U\right\} + e^{\varepsilon}\delta_j.
\label{eqn-thm-regret-case-2}
\end{align}
\end{enumerate}
Combining \eqref{eqn-thm-regret-case-1} and \eqref{eqn-thm-regret-case-2}, we obtain that
\begin{align*}
\sum_{n=2}^N \left[ F_n(\btheta_n) - \inf_{\btheta\in\Omega} F_n(\btheta) \right]
&=
\sum_{j=1}^{J} \sum_{n=N_{j-1}+1}^{N_j} \left[ F_{n+1}(\btheta_{n+1}) - \inf_{\btheta\in\Omega} F_{n+1}(\btheta) \right] \\[4pt]
&\le 
\sum_{j=1}^{J} \left[ 3e^{3\varepsilon} C^2 \sum_{n=N_{j-1}+1}^{N_j} \min\left\{\tau(N,n-N_{j-1}) , U\right\} + e^{\varepsilon}\delta_j \right] \\[4pt]
&= 
3e^{3\varepsilon} C^2\sum_{j=1}^{J} \reg(N_j-N_{j-1}) + e^{\varepsilon} \sum_{j=1}^{J} \delta_j.
\end{align*}
The proof is completed by adding $F_1(\btheta_1) - \inf_{\btheta\in\Omega} F_1(\btheta)$ to both sides of the inequality.

%% file: appendix_proof.tex
\section{Proofs for \Cref{sec-theory}}

\subsection{Verifications of Examples \ref{eg-Gaussian-mean}, \ref{eg-linear-regression}, \ref{eg-logistic-regression} and \ref{eg-robust-regression}}\label{sec-eg-strongly-convex}

{\bf \Cref{eg-Gaussian-mean} (Gaussian mean estimation).}
Since $F_n(\btheta)=\frac{1}{2}\|\btheta-\btheta_n^*\|_2^2+\sigma_0^2d/2$, $\nabla\ell(\btheta,\bz)=\btheta-\bz$, $\nabla F_n(\btheta)=\btheta-\btheta_n^*$ and $\nabla^2\ell(\btheta,\bz) = \nabla^2F_n(\btheta) =\bI_d$. Assumptions \ref{assumption-strongly-convex} and \ref{assumption-concentration-strong-cvx} clearly hold. To see why $c \geq 1/2$, note that for $\bz\sim\cP_n = N(\btheta_n^*,\sigma_0^2\bI_d)$, we have $\bu^\top(\bz-\btheta_n^*) \sim N(0,\sigma_0^2)$ for all $\bu\in\SSS^{d-1}$, so
\[
\| \nabla\ell(\btheta,\bz) - \nabla F_n(\btheta) \|_{\psi_1} = \| \bz - \btheta_n^* \|_{\psi_1} \geq \frac{1}{2} \sqrt{ \EE_{z \sim N(0, \sigma_0^2)} [z^2] }  = \frac{1}{2}\sigma_0. 
\]

\vspace{1em}
\noindent{\bf \Cref{eg-linear-regression} (Linear regression).}
Define $\bSigma_n = \EE(\bx_n\bx_n^\top)$. From $\ell(\btheta,\bz_n)=\frac{1}{2}[\bx_n^\top(\btheta-\btheta_n^*)-\varepsilon_n]^2$ and $\EE(\varepsilon_n|\bx_n)=0$ we obtain that $F_n(\btheta)=\frac{1}{2}(\btheta-\btheta_n^*)^\top
\bSigma_n
(\btheta-\btheta_n^*)+\EE\varepsilon_n^2/2$, $\nabla\ell(\btheta,\bz_n)=[\bx_n^\top(\btheta-\btheta_n^*)-\varepsilon_n]\bx_n$, $\nabla^2\ell(\btheta,\bz_n)=\bx_n\bx_n^\top$, $\nabla F(\btheta)= \bSigma_n (\btheta-\btheta_n^*)$, and $\nabla^2F(\btheta)=\bSigma_n$. Then,
\begin{align*}
& \sup_{\btheta\in\Omega}\|\nabla\ell(\btheta,\bz_n)-\nabla F_n(\btheta)\|_{\psi_1}
\lesssim
\sup_{\btheta\in\Omega}\|\bx_n^\top(\btheta-\btheta_n^*)-\varepsilon_n\|_{\psi_2}\|\bx_n\|_{\psi_2}
\lesssim (M+1)\sigma_0^2,
\\
& \EE
\bigg[ 
\sup_{ \btheta \in \Omega } \| \nabla^2 \ell ( \btheta, \bz_n ) \|_2
\bigg]
=
\EE\|\bx_n\bx_n^\top\|_2
=
\EE \left[ \|\bx_n\|_2^2 \right]
\lesssim \sigma_0^2d .
\end{align*}
Assumption \ref{assumption-concentration-strong-cvx} holds with $\sigma \asymp (M+1)\sigma_0^2$ and $\lambda \asymp \sigma_0$. For all $\btheta\in\Omega$ and $\bv\in\SSS^{d-1}$, $\bv^\top\nabla^2F_n(\btheta)\bv = \EE(\bv^\top\bx_n)^2 \lesssim \sigma_0^2$, so Assumption \ref{assumption-strongly-convex} holds with $\rho \asymp \gamma\sigma_0^2$ and $L \asymp \sigma_0^2$.

\vspace{1em}
\noindent{\bf \Cref{eg-logistic-regression} (Logistic regression).}
The logistic loss is given by $\ell(\btheta,\bz)=b(\bx^\top\btheta)-y\bx^\top\btheta$, where $b(t)=\log(1+e^t)$. Then $b'(t)=1/(1+e^{-t})\in(0,1)$, $b''(t)=1/(2+e^t+e^{-t})\in(0,1/4]$, $\nabla\ell(\btheta,\bz)=\bx[b'(\bx^\top\btheta)-y]$, $\nabla^2\ell(\btheta,\bz)=b''(\bx^\top\btheta)\bx\bx^\top$. Since $\|\bx_n\|_{\psi_1}\le\sigma_0$, then
\begin{align*}
& \|\nabla\ell(\btheta,\bz_n)-\nabla F_n(\btheta)\|_{\psi_1}
\lesssim
\|\nabla\ell(\btheta,\bz_n)\|_{\psi_1}
\lesssim
\|\bx_n\|_{\psi_1}
\leq 
\sigma_0, \\
&
\EE
\bigg[ 
\sup_{ \btheta \in \Omega } \| \nabla^2 \ell ( \btheta, \bz_n ) \|_2
\bigg]
\lesssim \EE\|\bx_n\bx_n^\top\|_2
= \EE \left[\|\bx_n\|_2^2 \right]
\lesssim
\sigma_0^2d.
\end{align*}
Assumption \ref{assumption-concentration-strong-cvx} holds with $\sigma\asymp\sigma_0$ and $\lambda\asymp\sigma_0$.

Next we find upper and lower bounds on the eigenvalues of $\nabla^2F_n$ over $\Omega$. For all $\btheta\in\Omega$ and $\bv\in\SSS^{d-1}$, $\bv^\top\nabla^2F_n(\btheta)\bv
=
\EE\left[b''(\bx_n^\top\btheta)\cdot(\bv^\top\bx_n)^2\right]
\lesssim
\sigma_0^2$, which implies $L \asymp \sigma_0^2$. We now lower bound the eigenvalues of $\nabla^2F_n$ to get $\rho$. 
Fix $\btheta\in\Omega$, take $E>0$, and define an event $\cE=\{|\bx_n^\top\btheta|\le E M \}$. If $\btheta=0$, then $\PP(\cE^c)=0$. If $\btheta\neq0$, then $\|\btheta\|_2\le M/2$ and $\|\bx_n\|_{\psi_1}\le\sigma_0$ imply that
\[
\PP(\cE^c)
\le 
\PP\left(\left|\bx_n^\top\frac{\btheta}{\|\btheta\|_2}\right|>E\right)
\le 
2\exp(-2cE/\sigma_0)
\] 
for some universal constant $c>0$. For all $\bv\in\SSS^{d-1}$,
\begin{align*}
\bv^\top\nabla^2F_n(\btheta)\bv
=
\EE\left[b''(\bx_n^\top\btheta)\cdot(\bv^\top\bx_n)^2\right]
&\ge 
\EE\left[b''(\bx_n^\top\btheta)\cdot(\bv^\top\bx_n)^2 1_{\cE}\right] \\
&\ge 
\frac{1}{2+e^{EM}+e^{-EM}}\EE\left[(\bv^\top\bx_n)^2-(\bv^\top\bx_n)^2\mathbf{1}_{\cE^c}\right] \\
&\ge 
\frac{1}{2+e^{EM}+e^{-EM}}\left[\EE(\bv^\top\bx_n)^2-\sqrt{\EE(\bv^\top\bx_n)^4\cdot\PP(\cE^c)}\right] \\
&\ge 
\frac{1}{2+e^{EM}+e^{-EM}}\left[\gamma\sigma_0^2-2^{9/2}\sigma_0^2\cdot\exp(-cE/\sigma_0)\right].
\end{align*}
Taking $E= \max\{ -(\sigma_0/c)\log\left( 2^{7/2}\gamma \right),1\}$ yields $\nabla^2F_n\succeq c_1\gamma\sigma_0^2\bI_d$ over $\Omega$, where $c_1=\left[2(2+e^{EM}+e^{-EM})\right]^{-1}$. We can take $\rho = c_1\gamma\sigma_0^2$.

Finally, basic theories of generalized linear models imply that the true parameter $\btheta_n^*$ is the minimizer of $F_n$. This completes the verification of Assumption \ref{assumption-strongly-convex}.

\vspace{1em}
\noindent{\bf \Cref{eg-robust-regression} (Robust linear regression).}
Take $u = 8M\sigma_0\sqrt{\log(c_0/\gamma)}$ with $c_0 > 16$. The function $h_u$ is convex and $u$-Lipschitz continuous on $\RR$. Its derivative
\[
h_u'(t) =
\begin{cases}
u,&\quad\text{if } t > u \\
t,&\quad\text{if } |t| \le u \\
-u,&\quad\text{if } t < u
\end{cases}
\]
is $1$-Lipschitz continuous and has range $[-u,u]$. Then $\nabla \ell(\btheta,\bz) = -h_u'(y-\bx^\top\btheta) \bx$, so
\[
\sup_{\btheta\in\Omega} \|\nabla\ell(\btheta,\bz_n) - \nabla F_n(\btheta) \|_{\psi_1}
\lesssim
\sup_{\btheta\in\Omega} \|\nabla\ell(\btheta,\bz_n)\|_{\psi_1}
\lesssim
\sup_{\btheta\in\Omega} \| h_u'(y_n-\bx_n^\top\btheta) \|_{\psi_2} \| \bx_n \|_{\psi_2}
\lesssim 
M\sigma_0^2,
\]
\begin{align*}
\EE
\left[ 
\sup_{ \substack{ \btheta,\btheta' \in \Omega \\ \btheta \neq \btheta' } }
\frac{\| \nabla \ell( \btheta , \bz_n) - \nabla \ell (\btheta',\bz_n) \|_2}{\| \btheta - \btheta' \|_2}
\right]
&=
\EE
\left[ 
\sup_{ \substack{ \btheta,\btheta' \in \Omega \\ \btheta \neq \btheta' } }
\frac{| h_u'(y_n-\bx_n^\top\btheta) - h_u'(y_n-\bx_n^\top\btheta') |\cdot\|\bx_n\|_2}{\| \btheta - \btheta' \|_2}
\right] \\[4pt]
&\le 
\EE
\left[ 
\sup_{ \substack{ \btheta,\btheta' \in \Omega \\ \btheta \neq \btheta' } }
\frac{| \bx_n^\top(\btheta - \btheta') |\cdot\|\bx_n\|_2}{\| \btheta - \btheta' \|_2}
\right]
\le 
\EE\left[ \|\bx_n\|_2^2 \right]
\lesssim
\sigma_0^2d.
\end{align*}
Thus, Assumption \ref{assumption-concentration-strong-cvx} holds with $\sigma\asymp M\sigma_0^2$ and $\lambda\asymp\sigma_0$.

We proceed to verify Assumption \ref{assumption-strongly-convex}. Clearly $F_n$ is convex. For all $\btheta,\btheta'\in \RR^d$,
\begin{align*}
\left\langle \nabla F_n(\btheta) - \nabla F_n(\btheta'), ~ \btheta - \btheta' \right\rangle
&=
\EE\left[ - \left( h_u'(y_n-\bx_n^\top\btheta) - h_u'(y_n-\bx_n^\top\btheta') \right) \cdot \bx_n^\top (\btheta - \btheta')  \right] \\[4pt]
&\le 
\EE\left[ \left| h_u'(y_n-\bx_n^\top\btheta) - h_u'(y_n-\bx_n^\top\btheta') \right| \cdot \left| \bx_n^\top (\btheta - \btheta') \right|  \right] \\[4pt]
&\le 
\EE\left[ \left( \bx_n^\top (\btheta - \btheta') \right)^2 \right]
\le 
2\sigma_0^2 \| \btheta - \btheta' \|_2^2,
\end{align*}
which implies that $F_n$ is $2\sigma_0^2$-smooth over $\RR^d$.

Take random variables $\varepsilon\sim\cQ_n^*$ and $\varepsilon'\sim\cQ_n$ such that $\varepsilon,\varepsilon',\bx_n$ are independent. Define $G,H:\RR^d\to\RR$ by
\[
G(\btheta) = 
(1-p)\EE \left[ h_u\left( \bx_n^\top(\bbeta_n^* - \btheta) + \varepsilon \right) \right]
\quad\text{and}\quad
H(\btheta) = p \EE \left[ h_u\left( \bx_n^\top(\bbeta_n^* - \btheta) + \varepsilon' \right) \right].
\] 
Then $F_n = G + H$. Fix $\btheta,\btheta'\in \Omega$. Define an event $\cE = \left\{ |\bx_n^\top(\bbeta_n^* - \btheta) + \varepsilon| \le u \right\} \cap \left\{ |\bx_n^\top(\bbeta_n^* - \btheta') + \varepsilon| \le u \right\}$. Since $\|\bx_n^\top(\bbeta_n^*-\btheta) + \varepsilon \|_{\psi_2} \le 2M\sigma_0$, then by Proposition 2.5.2 in \cite{Ver18}, $\PP(\cE^c) \le 4\exp\left( -u^2/(8M\sigma_0)^2 \right)$. This implies
\begin{align*}
& \frac{1}{1-p} \langle \nabla G(\btheta) - \nabla G(\btheta'), ~ \btheta - \btheta' \rangle \\[4pt]
&=
\EE\left[ \left( h_u'(y_n - \bx_n^\top\btheta) - h_u'(y_n-\bx_n^\top\btheta') \right) \cdot \bx_n^\top (\btheta - \btheta') \right] \\[4pt]
&=
\EE\left[ \left( \bx_n^\top (\btheta - \btheta')\right)^2 \bm{1}_{\cE} \right]
+
\EE\left[ \left( h_u'(y_n - \bx_n^\top\btheta) - h_u'(y_n-\bx_n^\top\btheta') \right) \cdot \bx_n^\top(\btheta - \btheta') \cdot \bm{1}_{\cE^c} \right] \\[4pt]
&\ge 
\EE\left[ \left( \bx_n^\top(\btheta - \btheta') \right)^2 \bm{1}_{\cE} \right]
-
\EE\left[ \left| h_u'(y_n - \bx_n^\top\btheta) - h_u'(y_n-\bx_n^\top\btheta') \right| \cdot \left| \bx_n^\top(\btheta - \btheta') \right| \cdot \bm{1}_{\cE^c} \right] \\[4pt]
&\ge 
\EE\left[ \left( \bx_n^\top(\btheta - \btheta') \right)^2 \bm{1}_{\cE} \right]
-
\EE\left[ \left( \bx_n^\top(\btheta - \btheta') \right)^2 \bm{1}_{\cE^c} \right] \\[4pt]
&=
\EE\left[ \left( \bx_n^\top(\btheta - \btheta') \right)^2 \right]
-
2\EE\left[ \left( \bx_n^\top(\btheta - \btheta') \right)^2 \bm{1}_{\cE^c} \right] \\[4pt]
&\ge
\EE\left[ \left( \bx_n^\top(\btheta - \btheta') \right)^2 \right] - 2\sqrt{\EE\left[ \left( \bx_n^\top (\btheta - \btheta')  \right)^4 \right] \PP(\cE^c) } \\[4pt]
&\ge 
\sigma_0^2\|\btheta - \btheta'\|_2^2
\left[
\gamma
-
16 \exp\left( - \frac{u^2}{(8M\sigma_0)^2} \right) \right] \\[4pt]
&=
\left(1 - \frac{16}{c_0} \right) \gamma \sigma_0^2 \|\btheta - \btheta'\|_2^2,
\end{align*}
which shows that $G$ is $\rho$-strongly convex over $\Omega$, with $\rho = (1-p) (1-16/c_0) \gamma\sigma_0^2 > 0$. Since $\varepsilon$ has a symmetric distribution, then $G$ has a unique minimizer $\bbeta^*_n$ over $\RR$. Since $H$ is convex, then $F_n$ is $\rho$-strongly convex over $\Omega$.

It remains to show that $F_n$ has a minimizer in the interior of $\Omega$. For all $\btheta,\btheta'\in\RR^d$,
\[
|H(\btheta) - H(\btheta')|
\le 
pu \cdot \EE \left| \bx_n^\top(\btheta - \btheta') \right|
\le 
pu\sigma_0 \| \btheta - \btheta' \|_2.
\] 
Thus, $H$ is $\lambda_H$-Lipschitz continuous over $\RR^d$, with $\lambda_H = pu\sigma_0$. Recall that $G$ is $\rho$-strongly convex over $B(\bbeta_n^*,M/4)$. For $(p^{-1}-1)\gamma$ sufficiently large, $\lambda_H < \rho \cdot (M/4)$, so by Lemma F.2 in \cite{DWa23}, $F_n = G+H$ has a unique minimizer $\btheta_n^*$ over $\RR^d$, and $\|\btheta_n^* - \bbeta_n^*\|_2 < M/4$, which implies that $\btheta_n^*$ is an interior point of $\Omega$.

\subsection{Proof of \Cref{lem-PV-strongly-cvx}}\label{sec-lem-PV-strongly-cvx-proof}

More precisely, we will prove that
\[
J \leq 1 + 
\bigg( \frac{ \rho  }{ M \sigma \max\{\sigma/(\rho M),1\}} \bigg)^{1/3}
\bigg(
\frac{BN}{d}
\bigg)^{1/3}
V^{2/3}  .
\] 
We prove by construction. Define
\[
V(j, k) = \sum_{i = j}^{k - 1} \| \btheta^*_{i + 1} - \btheta^*_i \|_2 , \qquad \forall j \leq k 
\]
and $V = V(1, N)$. Let $N_0 = 0$. For $j\in\ZZ_+$, define
\[
N_j = \max \left\{ 
n \geq N_{j - 1} + 1 :~ V ( N_{j - 1} + 1, n ) 
\leq 
\sqrt{ \frac{2 M \sigma }{\rho} \max\left\{\frac{\sigma}{\rho M},1\right\} \cdot \frac{d}{B ( n - N_{j - 1} ) } }
\right\}.
\]
Let $J' = \max \{ j:~ N_j \leq N - 1 \}$. Then for every $j \in [J']$,
\[
\max_{N_{j - 1} < i, k \leq N_j } \| \btheta_i^* - \btheta_k^* \|_2 \leq 
V( N_{j - 1} + 1, N_j )
\leq 
\sqrt{ \frac{2 M \sigma }{\rho} \max\left\{\frac{\sigma}{\rho M},1\right\} \cdot \frac{d}{B ( N_j - N_{j - 1} ) } }.
\]
This shows that $\{\btheta_n^*\}_{n=1}^N$ consists of $J'$ quasi-stationary segments, hence $J \le J'$ by the minimality of $J$.

It remains to prove an upper bound on $J'$. By definition, for all $j \in [J '- 1]$ we have
\begin{align*}
V ( N_{j - 1} + 1 , N_j + 1 ) 
& > 
\sqrt{ \frac{2 M \sigma }{\rho} \max\left\{\frac{\sigma}{\rho M},1\right\} \cdot \frac{d}{B ( N_j - N_{j - 1} + 1 ) } } \\[4pt]
& \ge 
\sqrt{ \frac{M \sigma }{\rho} \max\left\{\frac{\sigma}{\rho M},1\right\} \cdot \frac{d}{B ( N_j - N_{j - 1} ) } }.
\end{align*}
Define $n_j = N_j - N_{j-1}$. From
\[
\sum_{j = 1}^{J' - 1} V ( N_{j - 1} + 1 , N_j + 1 ) = V ( 1 , N_{J' - 1} + 1 ) \leq V ( 1, N -1 ) \le V,
\]
we obtain that
\begin{align*}
\sum_{j=1}^{J'-1} n_j^{-1/2} \leq V \sqrt{ \frac{ \rho B }{ d M \sigma \max\{\sigma/ (\rho M),1\}} } .
\end{align*}
By H\"{o}lder's inequality,
\begin{align*}
J' - 1 
& = 
\sum_{j=1}^{J'-1} n_j^{1/3} n_j^{-1/3}
\leq \bigg( \sum_{j=1}^{J'-1} ( n_j^{1/3} )^{3} \bigg)^{1/3} \bigg( \sum_{j=1}^{J'-1} ( n_j^{-1/3} )^{3/2} \bigg)^{2/3} \\[4pt]
& = 
\bigg( \sum_{j=1}^{J'-1} n_j \bigg)^{1/3} \bigg( \sum_{j=1}^{J'-1} n_j^{-1/2} \bigg)^{2/3} 
\leq 
N^{1/3} \cdot V^{2/3} \bigg( \frac{ \rho B }{ d M \sigma \max\{\sigma/ (\rho M),1\}} \bigg)^{1/3} \\[4pt]
& =  
\bigg( \frac{ \rho  }{ M \sigma \max\{\sigma/ (\rho M),1\}} \bigg)^{1/3} \cdot \bigg( \frac{ B N }{d} \bigg)^{1/3} V^{2/3}.
\end{align*}
The claimed upper bound follows from $J\le J'$.

We note that similar techniques that relate segmentation and path variation through H\"{o}lder's inequality are also used in \cite{BWa19} for one-dimensional mean estimation, and in \cite{CLL19} and \cite{SKp22} for non-stationary bandits.

\subsection{Proof of \Cref{thm-online-strongly-convex}}\label{sec-thm-online-strongly-convex-proof}

For notational convenience, we drop the subscript of $J_N$. We will prove the following more refined bound:
\begin{equation}
\sum_{n=1}^{N} \bigg[ F_{n} ( \btheta_n ) -  F_{n} ( \btheta_n^* ) \bigg]  
\lesssim
1 + \sum_{j=1}^J \min \bigg\{   \frac{ d  }{ B} ,~ N_j - N_{j-1} \bigg\} +  \sum_{j=1}^{J}\|\btheta_{N_j+1}^*-\btheta_{N_j}^*\|_2^2 .
\label{eqn-thm-online-strongly-cvx-refined}
\end{equation}
Then the regret bound in \Cref{thm-online-strongly-convex} follows from
\[
\sum_{j=1}^J \min \bigg\{   \frac{ d  }{ B} ,~ N_j - N_{j-1} \bigg\}
\le 
\min \left\{ \frac{Jd}{B}, N \right\}
\quad\text{and}\quad
\sum_{j=1}^{J}\|\btheta_{N_j+1}^*-\btheta_{N_j}^*\|_2^2 \le JM^2.
\]

Define a condition number $\kappa=L/\rho$. We invoke the following lemma, proved in \Cref{sec-lem-strongly-convex-proof}.

\begin{lemma}\label{lem-offline-strongly-convex}
Let Assumptions \ref{assumption-bounded-domain}, \ref{assumption-strongly-convex} and \ref{assumption-concentration-strong-cvx} hold. Choose any $\alpha \in (0, 1]$. There exists a universal constant $C_{\psi} \ge 1$ such that if
\[
\psi (n, k ) = C_{\psi}  M \sigma \max\left\{ \frac{\sigma}{\rho M}, 1 \right\}
\cdot
\frac{d }{ B k } 
\log  ( 1 + \alpha^{-1} + B n + M \lambda^2 \sigma^{-1}
) ,
\]
then with probability at least $1 - \alpha$, it holds that for all $n \in \ZZ_+$ and $k \in [n-1]$, $f_{n, k}$ and $F_{n, k}$ are $ ( \log 2 ,  \psi ( n , k )  )$-close in the sense of \Cref{defn-approx}.
\end{lemma}	

Suppose that the event in \Cref{lem-offline-strongly-convex} happens, which has probability at least $1-\alpha$. We will apply the general result in \Cref{thm-regret}. To this end, we need to verify Assumption \ref{assumption-approximation} and Condition \ref{condition-thresholds}, and show that the segmentation in \Cref{def-segmentation-strong-cvx} is also a segmentation in the sense of \Cref{def-segmentation}.
\begin{itemize}
\item (Assumption \ref{assumption-approximation}) Clearly $\psi(n,k)$ satisfies Assumption \ref{assumption-approximation}.

\item (Condition \ref{condition-thresholds}) If $C_{\tau}' \geq  C_{\psi} $ and
\[
\tau ( n , k ) \geq 6\cdot 2^5 C_{\tau}' M \sigma  \max \bigg\{  
\frac{\sigma}{ \rho M } 
, 1 \bigg\} 
\cdot
\frac{d }{ B k } 
\log  ( 1 + \alpha^{-1} + B n + M \lambda^2 \sigma^{-1}
),
~\forall n\in\ZZ_+,~k\in[n-1],
\]
then Assumption \ref{assumption-approximation} holds. Since $\rho,L,M,r, \lambda, \sigma,A$ are constants, then there exists a constant $\bar{C}_{\tau}$ such that when $C_{\tau} > \bar{C}_{\tau}$ and
\[
\tau ( n , k ) = 
C_{\tau}
\frac{d }{ B k } 
\log  ( \alpha^{-1} +  B + n 
),
\quad\forall n\in\ZZ_+,~k\in[n-1],
\]
Condition \ref{condition-thresholds} holds. 

\item (\Cref{def-segmentation}) Let $\{\btheta_n^*\}_{n=1}^N$ be segmented according to \Cref{def-segmentation-strong-cvx}. Fix $j\in[J]$. By Part \ref{lem-sufficient-minimizers} of \Cref{lem-sufficient}, for all $i,k\in\{N_{j-1}+1,...,N_j\}$, $F_i$ and $F_k$ are $(\log(4\kappa),(\rho/2)\|\btheta_i^*-\btheta_k^*\|_2^2)$-close. By \Cref{def-segmentation-strong-cvx},
\[
\frac{\rho}{2}\|\btheta_i^*-\btheta_k^*\|_2^2
\le 
M\sigma\max\left\{\frac{\sigma}{\rho M},1\right\}\frac{d}{B(N_j-N_{j-1})}
\le 
\min_{N_{j-1}<n\le N_j}\psi(n,n-N_{j-1}),
\]
so that show that the segmentation is also a segmentation in the sense of \Cref{def-segmentation} with $\varepsilon=\log(4\kappa)$ and $\delta_j = \|\btheta_{N_j+1}^* - \btheta_{N_j}^* \|_2^2$. 
\end{itemize}

We will now apply \Cref{thm-regret}. On the one hand, $ \sup_{ \btheta \in \Omega } F_{n} ( \btheta ) - \inf_{ \btheta \in \Omega } F_{n} ( \btheta ) \leq L M^2 \lesssim 1$, $\forall n \in [N]$. On the other hand, according to Part \ref{lem-sufficient-minimizers} in \Cref{lem-sufficient}, for each $n \in [N-1]$, $F_n$ and $F_{n+1}$ are $(\log(4\kappa),(\rho/2)\|\btheta_{n+1}^*-\btheta_{n}^*\|_2^2)$-close. Then, \Cref{thm-regret} implies
\begin{align*}
\sum_{n=1}^{N}\bigg[F_{n}(\btheta_{n})-F_{n}(\btheta_n^*)\bigg]
& \lesssim
1 + \sum_{j = 1}^J \reg( N_j - N_{j - 1} ) +   \sum_{j=1}^{J} \| \btheta^*_{N_j + 1} - \btheta^*_{N_j} \|_2^2 ,
\end{align*}
where $\lesssim$ only hides a constant factor and $\reg(n) = \sum_{ i=1 }^n \min \{ \tau (N, i) , 1 \}$. To finish the proof, note that
\[
\reg(n)
\le 
\min \left\{ \sum_{ i=1 }^n \tau (N, i) ,~ n \right\}
\lesssim
\min \left\{ \sum_{ i=1 }^n \frac{d}{Bi} ,~ n \right\} \lesssim \min \left\{ \frac{d}{B}, n \right\},
\]
where $\lesssim$ hides polylogarithmic factors of $B$, $N$ and $\alpha^{-1}$. Therefore,
\[
\sum_{n=1}^{N}\bigg[F_{n}(\btheta_{n})-\inf_{\btheta\in\Omega}F_{n}(\btheta_n^*)\bigg]
\lesssim
1 + \sum_{j=1}^J \min \left\{ \frac{d}{B}, ~ N_j - N_{j-1} \right\} + \sum_{j=1}^J \| \btheta^*_{N_j + 1} - \btheta^*_{N_j} \|_2^2,
\]
where $\lesssim$ hides polylogarithmic factors of $B$, $N$ and $\alpha^{-1}$. This proves \eqref{eqn-thm-online-strongly-cvx-refined}.

\subsection{Proof of \Cref{lem-offline-strongly-convex}}\label{sec-lem-strongly-convex-proof}
 
We will use the following concentration inequality, proved in \Cref{sec-lem-grad-uniform-proof}.

\begin{lemma}[Uniform concentration of gradient]\label{lem-grad-uniform}
Let Assumptions \ref{assumption-bounded-domain} and \ref{assumption-concentration-strong-cvx} hold. Define
\[
U ( n, k , \delta ) = \frac{d }{ B k } 
\log  ( 1 + \delta^{-1} + B k + M \lambda^2 \sigma^{-1}
 ) .
\]
There exists a universal constant $C \ge 1$ such that
\begin{align*}
& \PP \bigg( \sup_{  \btheta \in \Omega } \| \nabla f_{n, k} (\btheta) - \nabla F_{n, k} (\btheta) \|_2 
\geq 
C \sigma  \max \{
U ( n, k , \delta ) , \sqrt{
	U ( n, k , \delta ) 
}
\}
\bigg) \leq  \delta  , \qquad \forall \delta \in (0, 1].
\end{align*}
\end{lemma}

Choose any $\alpha \in(0,1]$ and let
\[
U ( n, k ) = \frac{d }{ B k } 
\log \bigg(
1 + \frac{2 n^3}{\alpha } + B k + M \lambda^2 \sigma^{-1}
\bigg) .
\]
Define an event
\[
\cA = \bigg\{
\sup_{  \btheta \in \Omega } \| \nabla f_{n, k} (\btheta) - \nabla F_{n, k} (\btheta) \|_2 
<  
C \sigma \max \{
U ( n, k  ) , \sqrt{
	U ( n, k ) 
}
\}
,~~\forall n \in \ZZ_+, ~ k \in [n-1]
\bigg\}
\]
By \Cref{lem-grad-uniform} and union bounds,
\[
\PP ( \cA ) \geq
1-\sum_{n=1}^{\infty}\sum_{k=1}^n \frac{\alpha}{2 n^3}
=
1-\frac{\alpha}{2}\sum_{n=1}^{\infty}\sum_{k=1}^n\frac{1}{n^3}
=
1-\frac{\pi^2}{12}\alpha
>
1-\alpha.
\]
Here we used Euler's celebrated identity $\sum_{n=1}^{\infty} n^{-2} = \pi^2 / 6$.

From now on we assume that $\cA$ happens. Take $n\in\ZZ_+$ and $k \in [n-1]$. Define $U = U (n, k)$ and $D = C \sigma \max \{
U , \sqrt{ U }	\}$. 
Part \ref{lem-sufficient-grad-square} of \Cref{lem-sufficient} shows that $f_{n, k}$ and $F_{n,k}$ are $( \log 2 ,   2\min\{D^2/\rho, DM\} )$-close. By direct calculation, $2DM = 2MC\sigma \max \{ U, \sqrt{U} \}$ and $\frac{2D^2}{\rho} = \frac{2C^2\sigma^2}{\rho} \max \{ U^2, U\}$, so
\begin{align*}
2 \min \bigg\{ \frac{D^2}{\rho}, 	DM \bigg\}
&\le 
\frac{2C\sigma}{\rho} \max \bigg\{ C\sigma , \rho M \bigg\}
\min \bigg\{
\max \{
U , \sqrt{U} \} , ~ \max \{ U^2, U \}
\bigg\} \\[4pt]
& \le 2C^2 M \sigma \max\left\{ \frac{\sigma}{\rho M}, 1 \right\} U.
\end{align*}
Therefore, $f_{n, k}$ and $F_{n, k}$ are
\[
\bigg(
\log 2 ,~ 2C^2 M \sigma \max\left\{ \frac{\sigma}{\rho M}, 1 \right\} U(n,k)
\bigg)
\]
-close. The facts $k \in [n-1]$ and $B \geq 1$ force
\[
U(n, k) \lesssim \frac{d }{ B k } 
\log  ( 1 + \alpha^{-1} + B n + M \lambda^2 \sigma^{-1}
) .
\]

On top of the above, we can find a universal constant $C_{\psi} \ge 1$ such that when
\[
\psi (n, k ) =  C_{\psi} M \sigma \max\left\{ \frac{\sigma}{\rho M}, 1 \right\}
\frac{d }{ B k } 
\log  ( 1 + \alpha^{-1}+ B n + M \lambda^2 \sigma^{-1}
 ) ,
\]
we have
\[
\PP \bigg(
\text{for all } n \in \ZZ_+ \text{ and } k \in [n-1],~
f_{n, k}
\text{ and }
F_{n, k}
\text{ are }
(\log 2 , \psi (n, k) )
\text{-close} 
\bigg) \geq  1 - \alpha .
\]

\subsection{Proof of \Cref{lem-grad-uniform}}\label{sec-lem-grad-uniform-proof}

Choose any $\varepsilon > 0$. Since $\diam (\Omega) = M$ (Assumption \ref{assumption-bounded-domain}), a standard volume argument (Lemma 5.2 in \cite{Ver10}) shows that $\Omega$ has an $\varepsilon$-net $\cN$ with $| \cN | \leq ( 1 + M / \varepsilon )^d $. Then, for every $\btheta \in \Omega$, there exists $\pi(\btheta) \in \cN$ such that $\|\btheta - \pi(\btheta) \|_2 \le \varepsilon$. Since
\begin{align*}
& \| \nabla f_{n, k} (\btheta) - \nabla F_{n, k} (\btheta) \|_2 \\
& \leq \| \nabla f_{n, k} (\btheta) - \nabla f_{n, k} (\pi(\btheta)) \|_2 
+ \| \nabla f_{n, k} (\pi(\btheta)) - \nabla F_{n, k} (\pi(\btheta)) \|_2 
+ \| \nabla F_{n, k} (\pi(\btheta)) - \nabla F_{n, k} (\btheta) \|_2,
\end{align*}
then
\begin{align}
\sup_{  \btheta \in \Omega } \| \nabla f_{n, k} (\btheta) - \nabla F_{n, k} (\btheta) \|_2
& \le 
\max_{  \btheta \in \cN } \| \nabla f_{n, k} (\btheta) - \nabla F_{n, k} (\btheta) \|_2 \notag \\[4pt]
&\qquad + \sup_{\btheta\in\Omega} \| \nabla f_{n, k} (\btheta) - \nabla f_{n, k} (\pi(\btheta)) \|_2 \notag \\[4pt]
& \qquad + \sup_{\btheta\in\Omega}\| \nabla F_{n, k} (\btheta) - \nabla F_{n, k} (\pi(\btheta)) \|_2. \label{eqn-lem-grad-uniform-proof-0}
\end{align}

By Assumption \ref{assumption-concentration-strong-cvx}, \Cref{lem-subg-norm} and union bounds, there exists a universal constant $C>0$ such that for all $s\ge 0$,
\begin{equation}
\PP \bigg[ \max_{  \btheta \in \cN } \| \nabla f_{n, k} (\btheta) - \nabla F_{n, k} (\btheta) \|_2
\geq \sigma s
\bigg]
\le  
\exp \Big(  d [ \log 5  + \log ( 1 + M / \varepsilon ) ] - C B k \min \{ s^2, s \}  \Big).
\label{eqn-lem-grad-uniform-proof-1}
\end{equation}
By Assumption \ref{assumption-concentration-strong-cvx},
\begin{align}
\sup_{\btheta\in\Omega}\| \nabla F_{n, k} (\btheta) - \nabla F_{n, k} (\pi(\btheta)) \|_2
&\le 
\EE\left[ \sup_{\btheta\in\Omega} \| \nabla f_{n, k} (\btheta) - \nabla f_{n, k} (\pi(\btheta)) \|_2 \right] \notag \\[4pt]
&\le 
\EE\left[ \varepsilon \cdot \sup_{\btheta,\btheta'\in\Omega} \frac{\| \nabla f_{n,k}(\btheta) - \nabla f_{n,k} (\btheta') \|_2 }{\| \btheta - \btheta' \|_2 } \right]
\le 
\varepsilon \lambda^2 d. \label{eqn-lem-grad-uniform-proof-2}
\end{align}
By Markov's inequality, for all $\delta\in(0,1]$,
\begin{equation}
\PP \bigg(
\sup_{\btheta\in\Omega} \| \nabla f_{n, k} (\btheta) - \nabla f_{n, k} (\pi(\btheta)) \|_2 \ge \frac{2\varepsilon\lambda^2 d}{\delta}
\bigg)  \leq  \frac{\delta}{2}.
\label{eqn-lem-grad-uniform-proof-3}
\end{equation}
Substituting \eqref{eqn-lem-grad-uniform-proof-1}, \eqref{eqn-lem-grad-uniform-proof-2} and \eqref{eqn-lem-grad-uniform-proof-3} into \eqref{eqn-lem-grad-uniform-proof-0}, we obtain that for all $s\ge 0$, $\varepsilon>0$ and $\delta\in(0,1]$,
\[
\PP \bigg( \sup_{  \btheta \in \Omega } \| \nabla f_{n, k} (\btheta) - \nabla F_{n, k} (\btheta) \|_2 
\geq \sigma s + \frac{ 3 \varepsilon \lambda^2 d  }{ \delta } \bigg) 
\le 
\exp \Big(  d  \log ( 5 + 5 M / \varepsilon ) - C B k \min \{ s^2, s \}  \Big) +
\frac{\delta}{2}.
\]

Define $t = \min\{s^2,s\}$, then $s = \max\{\sqrt{t},t\}$. Take $\varepsilon = \frac{ \delta \sigma }{ 3 \lambda^2 B k }$, then
\begin{align*}
& \PP \bigg[ \sup_{  \btheta \in \Omega } \| \nabla f_{n, k} (\btheta) - \nabla F_{n, k} (\btheta) \|_2 
\geq \sigma \bigg( \max\{\sqrt{t},t\} + \frac{ d }{ B k } \bigg) \bigg] \\[4pt]
&\le
\exp \bigg[   d   \log 
\bigg(
5 + \frac{ 15 M \lambda^2 B k }{ \delta \sigma }
\bigg)
- C B k  t  \bigg] +
\frac{\delta}{2} \\[4pt]
&\le 
\exp\left[ d \big( \log (B k) + \log (1 / \delta) + \log  (5 + 15 M \lambda^2 / \sigma ) \big) - CBkt \right] + \frac{\delta}{2}.
\end{align*}
There exists a universal constant $C'>0$ such that when
\[
t \geq \frac{C' d }{ B k } 
\log  ( 1 + \delta^{-1} + B k + M \lambda^2 / \sigma ) ,
\]
it holds that
\[
CBkt \ge d \big( \log (B k) + \log (1 / \delta) + \log  (5 + 15 M \lambda^2 / \sigma ) \big) + \log(2/\delta),
\]
and hence
\begin{align*}
& \PP \bigg[ \sup_{  \btheta \in \Omega } \| \nabla f_{n, k} (\btheta) - \nabla F_{n, k} (\btheta) \|_2 
\geq \sigma \bigg( \max \{ \sqrt{ t } , t \} + \frac{ d }{ B k } \bigg) \bigg]
\leq  \delta .
\end{align*}

Define
\[
U ( n, k , \delta ) = \frac{d }{ B k } 
\log  (  1 + \delta^{-1} + B k + M \lambda^2 \sigma^{-1}
 ) .
\]
Based on the above derivations, we can find a sufficiently large constant $C''>0$ such that for all $\delta\in(0,1]$,
\begin{align*}
& \PP \bigg( \sup_{  \btheta \in \Omega } \| \nabla f_{n, k} (\btheta) - \nabla F_{n, k} (\btheta) \|_2 
\ge 
C'' \sigma \max \{
U ( n, k , \delta ) , \sqrt{
U ( n, k , \delta ) 
}
\}
\bigg) \leq  \delta.
\end{align*}

\subsection{Proof of \Cref{cor-segmentation-strongly-cvx}}\label{sec-cor-segmentation-strongly-cvx-proof}

For notational convenience we drop the subscripts of $J_N$ and $V_N$. According to \Cref{lem-PV-strongly-cvx}, $J -1 \lesssim   ( BN / d  )^{1/3} V^{2/3}$. By the more refined bound \eqref{eqn-thm-online-strongly-cvx-refined} for \Cref{thm-online-strongly-convex}, with probability at least $1-\alpha$, it holds for all $N\in\ZZ_+$ that
\begin{align*}
	& \sum_{n=1}^{N} \bigg[ F_{n} ( \btheta_n ) - F_{n} ( \btheta_n^* ) \bigg] 
	\lesssim  1 +
	\sum_{j=1}^J
	\min \bigg\{   \frac{ d }{ B} , ~N_j - N_{j-1} \bigg\} 
	+ \sum_{n=1}^{N-1}\|\btheta_{n+1}^*-\btheta_n^*\|_2^2 ,
\end{align*}
where $\lesssim$ only hides a polylogarithmic factor of $B$, $N$ and $\alpha^{-1}$. 
Note that
\begin{align*}
	& \sum_{j=1}^J
	\min \bigg\{  \frac{ d }{ B} , ~N_j - N_{j-1} \bigg\} 
	\le
	\min \bigg\{   \frac{ d  }{ B} , N \bigg\}
	+ (J - 1)  \frac{ d  }{ B} , \\
	& \sum_{n=1}^{N-1}\|\btheta_{n+1}^*-\btheta_n^*\|_2^2
	\le 
	\max_{n\in[N-1]}\|\btheta_{n+1}^*-\btheta_n^*\|_2\cdot\sum_{n=1}^{N-1}\|\btheta_{n+1}^*-\btheta_n^*\|_2
	\lesssim V.
\end{align*}
The proof is completed by combining the above estimates.

\subsection{Functional Variation-Based Regret Bound}\label{sec-FV-bound-strong-cvx}

In this subsection, we prove that the more refined version of \Cref{thm-online-strongly-convex} in \Cref{sec-thm-online-strongly-convex-proof} implies the functional variation-based regret bound $\widetilde{\cO}(\sqrt{ \FVB N})$ in \cite{BGZ15}, where $\FVB = \sum_{n=1}^{N-1} \sup_{\btheta\in\Omega^*} |F_{n+1}(\btheta) - F_n(\btheta)|$ is the functional variation (FV) and $\Omega^*$ is the convex hull of the minimizers $\{\btheta_n^*\}_{n=1}^N$. We will use the following lemma.

\begin{lemma}\label{lem-inf-norm-to-2-norm}
Let $\Omega\subseteq\RR^d$ be closed and convex. Let $f,g:\Omega\to\RR$ attain their minima at some points $\btheta_f^*$ and $\btheta_g^*$ in $\Omega$, respectively. Suppose that $g$ is $\rho$-strongly convex over $\Omega$, and $\btheta_g^*$ is an interior point of $\Omega$. Then,
\[
\frac{\rho}{4} \| \btheta_f^* - \btheta_g^* \|_2^2 \le \max_{\btheta\in\{\btheta_f^*,\btheta_g^*\}} |f(\btheta) - g(\btheta)|.
\]
\end{lemma}

\begin{proof}[\bf Proof of \Cref{lem-inf-norm-to-2-norm}]
Since $g$ is $\rho$-strongly convex over $\Omega$ and $\btheta_g^*$ is an interior point of $\Omega$, then
\begin{align*}
\|\btheta_f^* - \btheta_g^* \|_2^2
\le 
\frac{2}{\rho}\left[ g(\btheta_f^*) - g(\btheta_g^*) \right]
&=
\frac{2}{\rho}\left[ \left( g(\btheta_f^*) - f(\btheta_f^*) \right) + \left( f(\btheta_f^*) - f(\btheta_g^*) \right) + \left( f(\btheta_g^*) - g(\btheta_g^*) \right) \right] \\[4pt]
&\le 
\frac{4}{\rho} \cdot \max_{\btheta\in\{\btheta_f^*,\btheta_g^*\}} |f(\btheta) - g(\btheta)|.
\end{align*}
This completes the proof. 
\end{proof}

We now show that the number of quasi-stationary segments $J_N$ is bounded by $1 + \cO(\sqrt{N \FVB B/d})$.

\begin{lemma}[From functional variation to segmentation]\label{lem-FV-strong-cvx} \
Suppose $\{\btheta_n^*\}_{n=1}^N$ consists of $J$ quasi-stationary segments, and define $\FVB = \sum_{n=1}^{N-1} \sup_{\btheta\in\Omega^*} |F_{n+1}(\btheta) - F_n(\btheta)|$ where $\Omega^*$ is the convex hull of $\{\btheta_n^*\}_{n=1}^N$. Then
\[
J \le 1 + 2\sqrt{\frac{1}{M\sigma\max\{\sigma/(\rho M),1\}}} \cdot \sqrt{\frac{N \FVB B}{d}}.
\]
\end{lemma}

\begin{proof}[\bf Proof of \Cref{lem-FV-strong-cvx}] 
We prove by construction. Define
\[
V(j, k) = \sum_{i = j}^{k - 1} \sup_{\btheta\in\Omega^*} | F_{i + 1} (\btheta) - F_i (\btheta) | , \qquad \forall j \leq k,
\]
and $V = V (1,N)$. Let $N_0 = 0$. For $j\in\ZZ_+$, define
\[
N_j = \max \left\{ 
n \geq N_{j - 1} + 1 :~ V ( N_{j - 1} + 1, n ) 
\leq 
\frac{M\sigma}{2} \max\left\{\frac{\sigma}{\rho M},1\right\} \cdot \frac{d}{B ( n - N_{j - 1} ) } 
\right\}.
\]
Let $J' = \max \{ j:~ N_j \leq N - 1 \}$. Then for every $j \in [J']$, by \Cref{lem-inf-norm-to-2-norm},
\begin{align*}
\max_{N_{j - 1} < i, k \leq N_j } \| \btheta_i^* - \btheta_k^* \|_2
\le 
\sqrt{\frac{4}{\rho}\sup_{\btheta\in\Omega^*} |F_i(\btheta) - F_k(\btheta)|}
&\le 
\sqrt{\frac{4}{\rho} \cdot V ( N_{j - 1} + 1, N_j )} \\[4pt]
&\le
\sqrt{ \frac{2 M \sigma }{\rho} \max\left\{\frac{\sigma}{\rho M},1\right\} \cdot \frac{d}{B ( N_j - N_{j - 1} ) } }.
\end{align*}
This shows that $\{\btheta_n^*\}_{n=1}^N$ consists of $J'$ quasi-stationary segments. Thus, $J \le J'$ by the minimality of $J$.

It remains to prove an upper bound on $J'$. By definition, for all $j \in [J '- 1]$ we have
\begin{align*}
V ( N_{j - 1} + 1 , N_j + 1 ) 
& > 
 \frac{ M \sigma }{2} \max\left\{\frac{\sigma}{\rho M},1\right\} \cdot \frac{d}{B ( N_j - N_{j - 1} + 1 ) }  \\[4pt]
& \ge 
\frac{M \sigma }{4} \max\left\{\frac{\sigma}{\rho M},1\right\} \cdot \frac{d}{B ( N_j - N_{j - 1} ) }.
\end{align*}
Define $n_j = N_j - N_{j-1}$, then
\[
\FVB \ge \sum_{j=1}^{J'-1} V (N_{j-1}+1,N_j+1)
\ge 
\frac{ M \sigma }{4} \max\left\{\frac{\sigma}{\rho M},1\right\}\cdot\frac{d}{B} \sum_{j=1}^{J'-1} n_j^{-1}.
\]
By Cauchy-Schwarz inequality,
\[
J' - 1 
= 
\sum_{j=1}^{J'-1} n_j^{1/2} n_j^{-1/2}
\leq \left( \sum_{j=1}^{J'-1}  n_j  \right)^{1/2} \left( \sum_{j=1}^{J'-1} n_j^{-1} \right)^{1/2}
\le N^{1/2}\cdot C\sqrt{\frac{ \FVB B}{d}}
=
C\sqrt{\frac{N \FVB B}{d}},
\]
where $C = 2  \left( M\sigma\max\{\frac{\sigma}{\rho M},1\} \right)^{-1/2}$. The claimed upper bound follows from $J\le J'$.
\end{proof}

We are ready to derive a functional variation-based regret bound for \Cref{alg-online}.

\begin{corollary}[FV-based regret bound]\label{cor-FV-strong-convex}
Consider the setting of \Cref{thm-online-strongly-convex}. Define $\FVB = \sum_{n = 1 }^{N - 1} \sup_{\btheta\in\Omega^*} | F_{n + 1} (\btheta) - F_n (\btheta) | $, where $\Omega^*$ is the convex hull of $\{\btheta_n^*\}_{n=1}^N$. With probability at least $1-\alpha$, the output of \Cref{alg-online} satisfies
\[
\sum_{n=1}^{N} \bigg[ F_{n} ( \btheta_n ) - F_{n} ( \btheta_n^* ) \bigg]  
\lesssim
1 + \frac{d}{B} +
\sqrt{\frac{N \FVB d}{B}} + \FVB ,\quad\forall N\in\ZZ_+.
\]
Here $\lesssim $ only hides a polylogarithmic factor of $B$, $N$ and $\alpha^{-1}$.
\end{corollary}

\begin{proof}[\bf Proof of \Cref{cor-FV-strong-convex}]
For notational convenience, we drop the subscript $N$ of $J_N$. By the more refined bound \eqref{eqn-thm-online-strongly-cvx-refined} for \Cref{thm-online-strongly-convex}, with probability at least $1-\alpha$, it holds for all $N\in\ZZ_+$ that
\[
\sum_{n=1}^{N} \bigg[ F_{n} ( \btheta_n ) - F_{n} ( \btheta_n^* ) \bigg] 
	\lesssim  1 +
	\sum_{j=1}^J
	\min \bigg\{   \frac{ d }{ B} , ~N_j - N_{j-1} \bigg\} 
	+ \sum_{n=1}^{N-1}\|\btheta_{n+1}^*-\btheta_n^*\|_2^2 ,
\]
where $\lesssim$ only hides a polylogarithmic factor of $B$, $N$ and $\alpha^{-1}$. 
By \Cref{lem-FV-strong-cvx}, $J -1 \lesssim  \sqrt{N \FVB B/d}$, so
\[
\sum_{j=1}^J
\min \bigg\{  \frac{ d }{ B} , ~N_j - N_{j-1} \bigg\}
\le 
\min \bigg\{   \frac{ d  }{ B} , N \bigg\} + (J - 1)  \frac{ d  }{ B}
\lesssim
\frac{d}{B} + \sqrt{\frac{N \FVB d}{B}}.
\]
By \Cref{lem-inf-norm-to-2-norm},
\[
\sum_{n=1}^{N-1}\|\btheta_{n+1}^*-\btheta_n^*\|_2^2 \lesssim \sum_{n=1}^N \sup_{\btheta\in\Omega^*} |F_{n+1}(\btheta) - F_n(\btheta)| = \FVB.
\]
The proof is completed by combining the above estimates.
\end{proof}


\subsection{Verifications of Examples \ref{eg-linear-opt}, \ref{eg-quantile-regression}, \ref{eg-newsvendor} and \ref{eg-SVM}}\label{sec-eg-Lipschitz}

{\bf \Cref{eg-linear-opt} (Stochastic linear optimization).} Note that $\ell (\btheta, \bz_n) = \btheta^{\top} \bz_n$ and $F_n (\btheta) = \btheta^{\top} ( \EE \bz_n )$. For any $\btheta_1, \btheta_2 \in \Omega$,
\begin{align*}
& \|\ell(\btheta_1,\bz_n)- \ell(\btheta_2,\bz_n) - [F_n(\btheta_1)-F_n(\btheta_2)]  \|_{\psi_2} \\
& \leq  \| \ell(\btheta_1,\bz_n) \|_{\psi_2} + \| \ell(\btheta_2,\bz_n)  \|_{\psi_2} + \EE | \ell(\btheta_1,\bz_n) | + \EE | \ell(\btheta_2,\bz_n) | \leq 4 \sigma_0 .
\end{align*}
By Jensen's inequality,
\begin{align*}
& \|\EE\bz_n\|_2^2 = \| (\EE\bz_n) (\EE\bz_n)^{\top} \|_2
\leq \| \EE (\bz_n \bz_n^{\top}) \|_2 \leq \sigma_0^2, \\
&( \EE\|\bz_n\|_2 )^2 \leq
\EE \|\bz_n\|_2^2 = \Tr [ \EE (\bz_n \bz_n^{\top}) ] \leq \sigma_0^2 d.
\end{align*}
This implies
\begin{align*}
& \sup_{\substack{\btheta_1,\btheta_2\in\Omega \\ \btheta_1\neq\btheta_2}} \frac{\big|F_n(\btheta_1)- F_n(\btheta_2)  \big|}{\|\btheta_1-\btheta_2\|_2}
= 
\|\EE\bz_n\|_2 \le \sigma_0, \\[4pt]
& \EE
\left[ 
\sup_{\substack{\btheta_1,\btheta_2\in\Omega \\ \btheta_1\neq\btheta_2}} \frac{\big|\ell(\btheta_1,\bz_n)- \ell(\btheta_2,\bz_n)  \big|}{\|\btheta_1-\btheta_2\|_2}
\right] 
= 
\EE \| \bz_n \|_2 \leq \sigma_0 \sqrt{d}.
\end{align*}
Thus, Assumption \ref{assumption-concentration-Lip} holds with $\sigma=4\sigma_0$ and $\lambda=\sigma_0$.

\vspace{1em}
\noindent{\bf \Cref{eg-quantile-regression} (Quantile regression).}
Note that $\rho_{\nu}$ is 1-Lipschitz. Hence, 
\[
|\ell(\btheta_1,\bz_n)- \ell(\btheta_2,\bz_n)| 
= | \rho_{\nu}(y_n-\bx_n^\top\btheta_1) - \rho_{\nu}(y_n-\bx_n^\top\btheta_2) |
\leq |\bx_n^\top(\btheta_1-\btheta_2)|.
\]
We have
\begin{align*}
& | F_n (\btheta_1)- F_n (\btheta_2)| 
\leq \EE|\ell(\btheta_1,\bz_n)- \ell(\btheta_2,\bz_n)| 
\leq \sqrt{ \EE |\bx_n^\top(\btheta_1-\btheta_2)|^2 }
\leq \sigma_0 \| \btheta_1-\btheta_2 \|_2 .
\end{align*}
As a result,
\begin{align*}
& \EE
\left[ 
\sup_{\substack{\btheta_1,\btheta_2\in\Omega \\ \btheta_1\neq\btheta_2}} \frac{\big|\ell(\btheta_1,\bz_n)- \ell(\btheta_2,\bz_n)  \big|}{\|\btheta_1-\btheta_2\|_2}
\right] 
\le 
\EE \| \bx_n \|_2
\leq \sqrt{ \EE \| \bx_n \|_2^2 }
 \lesssim \sigma_0 \sqrt{d} , \\
&\sup_{\substack{\btheta_1,\btheta_2\in\Omega \\ \btheta_1\neq\btheta_2}} \frac{ |F_n(\btheta_1)- F_n(\btheta_2) |}{\|\btheta_1-\btheta_2\|_2} 
\le 
\sigma_0 .
\end{align*}
In addition,
\begin{align*}
& \sup_{\btheta_1,\btheta_2\in\Omega}\big\|\ell(\btheta_1,\bz_n)- \ell(\btheta_2,\bz_n) - [F_n(\btheta_1)-F_n(\btheta_2)] \big\|_{\psi_2} \\
&\leq  \sup_{\btheta_1,\btheta_2\in\Omega}\big\|\ell(\btheta_1,\bz_n)- \ell(\btheta_2,\bz_n) \big\|_{\psi_2} 
+ \sup_{\btheta_1,\btheta_2\in\Omega} |F_n(\btheta_1)-F_n(\btheta_2)|
\\
&  \le 
 \sup_{\btheta_1,\btheta_2\in\Omega}\| \bx_n^{\top}  (\btheta_1 - \btheta_2 ) \|_{\psi_2} 
 + \sigma_0 \sup_{\btheta_1,\btheta_2\in\Omega}  \| \btheta_1-\btheta_2 \|_2 
 \le 
2M\sigma_0.
\end{align*}
Therefore, Assumption \ref{assumption-concentration-Lip} holds with $\sigma\asymp M\sigma_0$ and $\lambda\asymp \sigma_0$.

\vspace{1em}
\noindent{\bf \Cref{eg-newsvendor} (Newsvendor problem).} The verification is similar to that of \Cref{eg-quantile-regression} and thus omitted.

\vspace{1em}
\noindent{\bf \Cref{eg-SVM} (Support vector machine).} The verification is similar to that of \Cref{eg-quantile-regression} and thus omitted.

\subsection{Proof of \Cref{lem-PV-Lip}}\label{sec-lem-PV-Lip-proof}

We will prove that
\begin{equation}
J \leq 1 + 
\frac{2}{\sigma^{2/3}}\left(\frac{BN}{d}\right)^{1/3}V^{2/3}
\label{eqn-lem-PV-Lip}
\end{equation}
by constructing a segmentation. Define $N_0=0$, $V(j,k)=\sum_{i=j}^{k-1}\|F_{i+1}-F_i\|_{\infty}$ for $j\le k$, and
\[
N_j=\max\left \{  N_{j-1}+1 \leq n \leq N - 1:V(N_{j-1}+1,n)\le \frac{\sigma}{2}\sqrt{\frac{d}{B(n-N_{j-1})}} \right\} ,
\qquad j \geq 1.
\]
Let $J'=\max\{j:N_j\le N\}$. Then for all $j\in[J]$,
\[
\max_{N_{j-1}<i,k\le N_j}\|F_i-F_k\|_{\infty}
\le 
V(N_{j-1}+1,N_j)
\le 
\frac{\sigma}{2}\sqrt{\frac{d}{B(N_j-N_{j-1})}},
\]
Thus, $\{F_n\}_{n=1}^N$ consists of $J'$ quasi-stationary segment, so $J \le J'$ by the minimality of $J$. 

We now show that $J'$ is upper bounded by the right hand side of \eqref{eqn-lem-PV-Lip}. If $J' = 1$, then the bound trivially holds. 
Suppose that $J' \geq 2$. For every $j\in[J'-1]$, by the definition of $N_j$,
\[
V(N_{j-1}+1,N_j+1)
>
\frac{\sigma}{2}\sqrt{\frac{d}{B(N_j-N_{j-1}+1)}}
\ge 
\frac{\sigma}{2\sqrt{2}}\sqrt{\frac{d}{B(N_j-N_{j-1})}}.
\]
Let $n_j=N_j-N_{j-1}$, then
\begin{align*}
V
\ge 
\sum_{j=1}^{J'-1}V(N_{j-1}+1,N_j+1)
>
\frac{\sigma}{2\sqrt{2}}\sqrt{\frac{d}{B}}
\cdot \sum_{j=1}^{J'-1}n_j^{-1/2}.
\end{align*}
By H\"{o}lder's inequality,
\begin{align*}
J' - 1 
&= 
\sum_{j=1}^{J'-1} n_j^{1/3} n_j^{-1/3}
\le \left[ \sum_{j=1}^{J'-1} \left( n_j^{1/3} \right)^{3} \right]^{1/3} \left[ \sum_{j=1}^{J'-1} \left( n_j^{-1/3} \right)^{3/2} \right]^{2/3} \\[6pt]
& = 
\left( \sum_{j=1}^{J'-1} n_j \right)^{1/3} \left( \sum_{j=1}^{J'-1} n_j^{-1/2} \right)^{2/3} 
\leq 
N^{1/3}\left(\frac{2\sqrt{2}}{\sigma}\sqrt{\frac{B}{d}}V\right)^{2/3} 
 =  
\frac{2}{\sigma^{2/3}}\left(\frac{BN}{d}\right)^{1/3}V^{2/3}.
\end{align*}
The proof is finished by combining the cases $J' = 1$ and $J' \geq 2$.

\subsection{Proof of \Cref{thm-online-Lip}}\label{sec-thm-online-Lip-proof}

For notational convenience, we drop the subscript of $J_N$. We will prove the following more refined bound:
\begin{equation}
\sum_{n=1}^N \bigg[ F_{n} ( \btheta_n ) - \inf_{ \btheta_n' \in \Omega } F_{n} ( \btheta_n' ) \bigg]
\lesssim
1 + \sum_{j=1}^J
\min\bigg\{ 
\sqrt{\frac{d (N_j - N_{j-1}) }{B}},~ N_j-N_{j-1} \bigg\} + \sum_{j=1}^J \|F_{N_j + 1} - F_{N_j}\|_{\infty} .
\label{eqn-thm-online-Lip-refined}
\end{equation}
Then the regret bound in \Cref{thm-online-Lip} follows from
\[
\sum_{j=1}^J\sqrt{N_j-N_{j-1}} \le \sqrt{J\left(\sum_{j=1}^J(N_j - N_{j-1})\right)} \le \sqrt{JN}
\quad\text{and}\quad
\sum_{j=1}^J \|F_{N_j + 1} - F_{N_j}\|_{\infty} \le J\lambda M.
\]

To prove \eqref{eqn-thm-online-Lip-refined}, we invoke the following lemma, which is proved in \Cref{sec-lem-Lip-proof}.

\begin{lemma}\label{lem-offline-Lip}
Let Assumptions \ref{assumption-bounded-domain} and \ref{assumption-concentration-Lip} hold. Let $\alpha \in (0, 1]$. Then there exists a universal constant $C_{\psi} \geq 1$ such that if
\[
\psi (n, k ) =  C_{\psi} \sigma\sqrt{\frac{d\log(1+\alpha^{-1}+Bn+\lambda\sigma^{-1})}{Bk}},
\]
then with probability at least $1 - \alpha$, it holds that for all $n \in \ZZ_+$ and $k \in [n-1]$, $f_{n, k}$ and $F_{n, k}$ are $ ( 0 ,  \psi ( n , k )  )$-close in the sense of \Cref{defn-approx}.
\end{lemma}

Suppose that the event in \Cref{lem-offline-Lip} holds, which happens with probability at least $1-\alpha$. We will apply the general result in \Cref{thm-regret}. To begin with, we need to verify Assumptions \ref{assumption-approximation} and Condition \ref{condition-thresholds}, and show that the segmentation in \Cref{def-segmentation-Lip} is also a segmentation in the sense of \Cref{def-segmentation}.
\begin{itemize}
	\item (Assumption \ref{assumption-approximation}) Clearly $\psi(n,k)$ satisfies Assumption \ref{assumption-approximation}.
	
	\item (Condition \ref{condition-thresholds}) If we take $C_{\tau}' \geq C_{\psi} $ and set
	\[
	\tau(n,k)=6C_{\tau}'\sigma\sqrt{\frac{d\log(1+\alpha^{-1}+Bn+\lambda\sigma^{-1})}{Bk}},\quad\forall n\in\ZZ_+,~k\in[n-1],
	\]
	then Assumption \ref{assumption-approximation} holds. Since $\rho,L,M,r, \lambda, \sigma$ are constants, there exists a constant $\bar{C}_{\tau}$ such that when $C_{\tau} \ge \bar{C}_{\tau}$ and
	\[
	\tau ( n , k ) = 
	C_{\tau}
	\sqrt{
	\frac{d }{ B k } 
	\log  ( \alpha^{-1} + B + n 
	)} ,
	\quad\forall n\in\ZZ_+,~k\in[n-1],
	\]
	Condition \ref{condition-thresholds} holds. 
	
	\item (\Cref{def-segmentation}) Let $\{F_n\}_{n=1}^N$ be segmented according to \Cref{def-segmentation-Lip}. Fix $j\in[J]$. By Part \ref{lem-sufficient-sup} of \Cref{lem-sufficient}, for all $i,k\in\{N_{j-1}+1,...,N_j\}$, $F_i$ and $F_k$ are $(0,2\|F_i-F_k\|_{\infty})$-close. By Assumption \ref{def-segmentation-Lip}, 
	\[
	2\|F_i-F_k\|_{\infty}
	\le 
	\sigma\sqrt{\frac{d}{B(N_j-N_{j-1})}}
	\le 
	\min_{N_{j-1}<n\le N_j}\psi(n,n-N_{j-1}),
	\]
	so the segmentation is also a segmentation in the sense of \Cref{def-segmentation} with $\varepsilon=0$ and $\delta_j = \|F_{j+1} - F_j\|_{\infty}$. 
\end{itemize}

We will now apply \Cref{thm-regret}. On the one hand, Assumptions \ref{assumption-bounded-domain} and \ref{assumption-concentration-Lip} force $ \sup_{ \btheta \in \Omega } F_{n} ( \btheta ) - \inf_{ \btheta \in \Omega } F_{n} ( \btheta ) \leq \lambda M \lesssim 1$, $\forall n \in [N]$. On the other hand, according to Part \ref{lem-sufficient-minimizers} in \Cref{lem-sufficient}, for each $n \in [N-1]$, $F_n$ and $F_{n+1}$ are $( 0, \| F_{n+1} - F_{n} \|_{\infty} )$-close. Then, \Cref{thm-regret} implies
\begin{align*}
\sum_{n=1}^{N} \bigg[ F_{n}(\btheta_{n})- \inf_{\btheta\in\Omega} F_{n}(\btheta) \bigg]
& \lesssim
1 + \sum_{j = 1}^J \reg( N_j - N_{j - 1} ) +   \sum_{j=1}^{J} \| F_{N_j + 1} - F_{N_j} \|_{\infty} ,
\end{align*}
where $\lesssim$ only hides a constant factor and $\reg(n) = \sum_{ i=1 }^n \min \{ \tau (N, i) , 1 \}$. Finally, note that
\[
\reg(n)
\le 
\min \left\{ \sum_{ i=1 }^n \tau (N, i) ,~ n \right\}
\lesssim
\min \left\{ \sum_{ i=1 }^n \sqrt{\frac{d}{Bi}} ,~ n \right\} \lesssim \min \left\{ \sqrt{\frac{dn}{B}}, n \right\},
\]
where $\lesssim$ hides polylogarithmic factors of $B$, $N$ and $\alpha^{-1}$. Therefore,
\[
\sum_{n=1}^{N}\bigg[F_{n}(\btheta_{n})-\inf_{\btheta\in\Omega}F_{n}(\btheta_n^*)\bigg]
\lesssim
1 + \sum_{j=1}^J \min \left\{ \sqrt{\frac{d (N_j - N_{j-1})}{B}}, ~ N_j - N_{j-1} \right\} + \sum_{j=1}^J \| F_{N_j + 1} - F_{N_j} \|_{\infty},
\]
where $\lesssim$ hides polylogarithmic factors of $B$, $N$ and $\alpha^{-1}$. This proves \eqref{eqn-thm-online-Lip-refined}.

\subsection{Proof of \Cref{lem-offline-Lip}}\label{sec-lem-Lip-proof}

We use the following lemma, proved in \Cref{sec-lem-fcn-uniform-proof}.

\begin{lemma}[Uniform concentration of function]\label{lem-fcn-uniform}
Let Assumptions \ref{assumption-bounded-domain} and \ref{assumption-concentration-Lip} hold. There exists a universal constant $C>0$ such that with
\[
W(n,k,\delta)
=
C\sigma\sqrt{\frac{d\log(1+\delta^{-1}+Bk+\lambda\sigma^{-1})}{Bk}},
\]
it holds that for all $\btheta_0\in\Omega$,
\[
\PP\bigg( \sup_{\btheta\in\Omega}\big| f_{n,k}(\btheta)-F_{n,k}(\btheta) - [f_{n,k}(\btheta_0)-F_{n,k}(\btheta_0)] \big| 
\ge 
W(n,k,\delta)
\bigg)
\le 1-\delta,
\quad\forall\delta\in(0,1].
\]
\end{lemma}

Take arbitrary $\alpha \in(0,1]$. By \Cref{lem-fcn-uniform} and union bounds, the event
\[
\cB = \bigg\{
\sup_{\btheta\in\Omega}\big| f_{n,k}(\btheta)-F_{n,k}(\btheta) - [f_{n,k}(\btheta_0)-F_{n,k}(\btheta_0)] \big| 
<
W\left(n,k,\frac{2n^3}{\alpha}\right),
~~\forall n\in \ZZ_+,~k\in[n-1]
\bigg\}
\]
has probability
\[
\PP(\cB)
\ge 
1-\sum_{n=1}^{\infty}\sum_{k=1}^n \frac{\alpha}{2 n^3}
=
1-\frac{\alpha}{2}\sum_{n=1}^{\infty}\sum_{k=1}^n\frac{1}{n^3}
=
1-\frac{\pi^2}{12}\alpha
>
1-\alpha.
\]
There exists a universal constant $C'\ge 1$ such that
\[
W\left(n,k,\frac{2n^3}{\alpha}\right)
\le 
C'\sigma\sqrt{\frac{d\log(1+\alpha^{-1}+Bn+\lambda\sigma^{-1})}{Bk}}.
\]
By Part \ref{lem-sufficient-sup} of \Cref{lem-sufficient} and Part \ref{lem-approx-monotonicity} of \Cref{lem-approx}, when the event $\mathcal{B}$ happens, we have that for all $n\in\ZZ_+$ and $k\in[n-1]$, $f_{n,k}$ and $F_{n,k}$ are $(0,\psi(n,k))$-close, where
\[
\psi(n,k)
=
2C'\sigma\sqrt{\frac{d\log(1+\alpha^{-1}+Bn+\lambda\sigma^{-1})}{Bk}}.
\]

\subsection{Proof of \Cref{lem-fcn-uniform}}\label{sec-lem-fcn-uniform-proof}

Take arbitrary $\varepsilon > 0$. Since $\diam (\Omega) = M$ (Assumption \ref{assumption-bounded-domain}), a standard volume argument (Lemma 5.2 in \cite{Ver10}) shows that $\Omega$ has an $\varepsilon$-net $\cN$ with $| \cN | \leq ( 1 + M / \varepsilon )^d $. Fix arbitrary $\btheta_0\in\Omega$. For any $\btheta \in \Omega$, there exists $\btheta' \in \cN$ such that $\|\btheta-\btheta'\|_2\le\varepsilon$, so
\begin{align*}
& \big| f_{n,k}(\btheta)-F_{n,k}(\btheta) - [f_{n,k}(\btheta_0)-F_{n,k}(\btheta_0)] \big| \\
& \le 
\big| f_{n,k}(\btheta)-F_{n,k}(\btheta) - [f_{n,k}(\btheta')-F_{n,k}(\btheta')] \big|
+
\big| f_{n,k}(\btheta')-F_{n,k}(\btheta') - [f_{n,k}(\btheta_0)-F_{n,k}(\btheta_0)] \big| \\
& \le 
\varepsilon \frac{\big| f_{n,k}(\btheta)-F_{n,k}(\btheta) - [f_{n,k}(\btheta')-F_{n,k}(\btheta')] \big|}{\|\btheta-\btheta'\|_2}
+
\big| f_{n,k}(\btheta')-f_{n,k}(\btheta_0) - [F_{n,k}(\btheta')-F_{n,k}(\btheta_0)] \big|.
\end{align*}
Then,
\begin{align}
&\sup_{  \btheta \in \Omega } \big| f_{n,k}(\btheta)-F_{n,k}(\btheta) - [f_{n,k}(\btheta_0)-F_{n,k}(\btheta_0)] \big| \notag  \\[4pt]
&\qquad\le
\varepsilon\sup_{\substack{\btheta_1,\btheta_2\in\Omega \\ \btheta_1\neq\btheta_2}}\frac{\big| f_{n,k}(\btheta_1)-f_{n,k}(\btheta_2) - [F_{n,k}(\btheta_1)-F_{n,k}(\btheta_2)] \big|}{\|\btheta_1-\btheta_2\|_2} \notag \\[4pt]
&\qquad\qquad + 
\max_{\btheta\in\cN}\big| f_{n,k}(\btheta)-f_{n,k}(\btheta_0) - [F_{n,k}(\btheta)-F_{n,k}(\btheta_0)] \big|.
\label{eqn-lem-fcn-uniform-proof-0}
\end{align}

By Markov's inequality and Assumption \ref{assumption-concentration-Lip}, for all $t>0$,
\begin{equation}
\PP\left( \varepsilon\sup_{\substack{\btheta_1,\btheta_2\in\Omega \\ \btheta_1\neq\btheta_2}}\frac{\big| f_{n,k}(\btheta_1)-f_{n,k}(\btheta_2) - [F_{n,k}(\btheta_1)-F_{n,k}(\btheta_2)] \big|}{\|\btheta_1-\btheta_2\|_2} 
\ge t \right)
\le 
\frac{2\varepsilon\lambda\sqrt{d}}{t}.
\label{eqn-lem-fcn-uniform-proof-1}
\end{equation}
By Assumption \ref{assumption-concentration-Lip}, for $\bz\sim\cP_n$, for all $\btheta\in\Omega$, $\big\| \ell(\btheta,\bz)-\ell(\btheta_0,\bz) - [F_n(\btheta)-F_n(\btheta_0)] \big\|_{\psi_2}\le\sigma$. By union bound and a Hoeffding-type inequality (Proposition 5.10 in \cite{Ver10}), there exists a universal constant $c>0$ such that for all $t\ge 0$,
\begin{align}
\PP\left(\max_{\btheta\in\cN}\big| f_{n,k}(\btheta)-F_{n,k}(\btheta) - [f_{n,k}(\btheta_0)-F_{n,k}(\btheta_0)] \big| > t\right)
& \le 
|\cN|\cdot e \cdot \exp\left(-\frac{cBkt^2}{\sigma^2}\right) \notag \\[4pt]
& =
\exp\left[1 + d\log\left(1+\frac{M}{\varepsilon}\right) - \frac{cBkt^2}{\sigma^2}\right]. \label{eqn-lem-fcn-uniform-proof-2}
\end{align}
Substituting \eqref{eqn-lem-fcn-uniform-proof-1} and \eqref{eqn-lem-fcn-uniform-proof-2} into \eqref{eqn-lem-fcn-uniform-proof-0} shows that for all $t > 0$,
\begin{multline*}
\PP\left( \sup_{  \btheta \in \Omega } \big| f_{n,k}(\btheta)-F_{n,k}(\btheta) - [f_{n,k}(\btheta_0)-F_{n,k}(\btheta_0)] \big|  \ge 2t \right) \\
\le 
\frac{2\varepsilon\lambda\sqrt{d}}{t} + \exp\left[1 + d\log\left(1+\frac{M}{\varepsilon}\right) - \frac{cBkt^2}{\sigma^2}\right].
\end{multline*}
Fix $\delta\in(0,1]$. Then
\begin{align}
\frac{2\varepsilon\lambda\sqrt{d}}{t} \le \frac{\delta}{2}
&\quad\Leftrightarrow\quad
t \ge \frac{4\varepsilon\lambda\sqrt{d}}{\delta},
\label{eqn-lem-fcn-uniform-proof-3} \\
\exp\left[1 + d\log\left(1+\frac{M}{\varepsilon}\right) - \frac{cBkt^2}{\sigma^2}\right] \le \frac{\delta}{2}
&\quad\Leftrightarrow\quad
t \ge \frac{\sigma}{\sqrt{cBk}}\sqrt{
1 + \log\frac{2}{\delta} + d\log\left(1+\frac{M}{\varepsilon}\right)}.
\label{eqn-lem-fcn-uniform-proof-4}
\end{align}
Let $\varepsilon = \frac{\delta \sigma}{4\lambda\sqrt{Bk}}$. For \eqref{eqn-lem-fcn-uniform-proof-3} and \eqref{eqn-lem-fcn-uniform-proof-4} to hold, we require
\[
t \ge \frac{\sigma}{\sqrt{cBk}}\sqrt{\max\left\{
\sqrt{c}d,
~1+\log\frac{2}{\delta}+d\log\left(1+\frac{4\lambda}{\delta\sigma}\sqrt{Bk}\right)
\right\}}.
\]
Since
\[
\log\left(1+\frac{4\lambda}{\delta\sigma}\sqrt{Bk}\right)
\le 
6\log(\delta^{-1}+Bk+\lambda\sigma^{-1}),
\]
then there exists a universal constant $C>0$ such that with
\[
W(n,k,\delta)=C\sigma\sqrt{\frac{d\log(1+\delta^{-1}+Bk+\lambda\sigma^{-1})}{Bk}},
\]
we have
\[
\PP\bigg( \sup_{\btheta\in\Omega}\big| f_{n,k}(\btheta)-F_{n,k}(\btheta) - [f_{n,k}(\btheta_0)-F_{n,k}(\btheta_0)] \big| 
\ge 
W(n,k,\delta)
\bigg)
\le 1-\delta.
\]

\subsection{Proof of \Cref{cor-segmentation-Lip}}\label{sec-cor-segmentation-Lip-proof}

For notational convenience we drop the subscripts of $J_N$ and $V_N$. By \Cref{lem-PV-Lip}, $J\lesssim 1 + (BN/d)^{1/3}V^{2/3}$. By the more refind bound \eqref{eqn-thm-online-Lip-refined} for \Cref{thm-online-Lip},
\begin{align*}
\sum_{n=1}^N \bigg[ F_{n} ( \btheta_n ) - \inf_{ \btheta \in \Omega } F_{n} ( \btheta ) \bigg]
& \lesssim
1 +
\sum_{j=1}^J
\min\bigg\{
\sqrt{\frac{d ( N_j - N_{j-1} ) }{B}} ,~ N_j - N_{j-1}
\bigg\}
+
\sum_{n=1}^J \|F_{N_j + 1} - F_{N_j}\|_{\infty} \\[4pt]
& \lesssim
1 + 
\sqrt{\frac{d  }{B}}
\sum_{j=1}^J
\sqrt{ N_j - N_{j-1} }
+
\sum_{n=1}^J \|F_{N_j + 1} - F_{N_j}\|_{\infty} \\[4pt]
& \lesssim 
1 + \sqrt{ \frac{ J N d}{ B } }  + V \\[4pt]
&\lesssim
1 + \sqrt{ \frac{ N d}{ B } } +  N^{2/3} \bigg( \frac{V d}{B} \bigg)^{1/3} + V.
\end{align*}
Here $\lesssim$ only hides a polylogarithmic factor of $B$, $N$ and $\alpha^{-1}$.

%% file: appendix_proof_lower.tex
\section{Proofs for \Cref{sec-theory-lower}}

\subsection{Proof of \Cref{thm-lower-strongly-cvx}}\label{sec-thm-lower-strongly-cvx-proof}

We present a stronger version of the lower bound, which will be used later.

\begin{theorem}\label{thm-lower-strongly-cvx-0}
	Let $\Omega = B (\bm{0}, 1)$. Choose any integer $N \geq 2$, $J \in [N - 1]$, $\bN \in \ZZ_+^J$ satisfying $N_1 <  \cdots < N_J = N - 1$ and $\br \in [0, 1]^J$. Define $N_0 = 0$ and $r_0 = 1$. For $\gamma \in [0, 1]$, define
	\begin{align*}
		\mathscr{P} ( \bN, \br, \gamma ) & = \bigg\{ ( \cP_1 , \cdots, \cP_N ) :~ \cP_n = N( \btheta_n^*, \bI_d ) \text{ and }  \btheta_n^* \in B(\bm{0},1/2)  ,~\forall n \in [N],\\
		& \qquad 
		\max_{N_{j - 1} < i, k \leq N_j } \| \btheta_i^* - \btheta_k^* \|_2 
		\leq 
		\sqrt{  \frac{ 8 \gamma c^2 d}{B ( N_j - N_{j - 1} ) } }  
		,~
		\|  \btheta^*_{N_{j} + 1 } -  \btheta^*_{N_{j} } \|_2 \leq r_j ,~\forall j \in [J] \bigg\} .
	\end{align*}
	There is a universal constant $C>0$ such that
	\begin{align*}
		& \inf_{\cA}	\sup_{ ( \cP_1,\cdots, \cP_N ) \in \mathscr{P} ( \bN, \br , \gamma )  } \EE \bigg[
		\sum_{n=1}^{N} \bigg(F_{n} ( \btheta_n ) -  F_{n} ( \btheta_n^* ) \bigg) \bigg] 
		\\ &
		\geq C  \bigg[ 1 +  \sum_{ j=1 }^J  \bigg( r_j^2 +
		\min \bigg\{ \frac{ \gamma d }{ B  } ,~  N_{j} - N_{j-1}  - 1  \bigg\}  +
		\min \bigg\{ \frac{d}{ B  } ,~ r_{j-1}^2 ( N_{j} - N_{j-1}  - 1  ) \bigg\} 
		\bigg) \bigg] .
	\end{align*}
\end{theorem}

We now show that \Cref{thm-lower-strongly-cvx} is a special case of \Cref{thm-lower-strongly-cvx-0}. Fix $J\in[N-1]$ and take $N_j = j(N-1)/J$ and $r_j = 1$ for $j\in[J]$. Let $\gamma = 1$. Then $\mathscr{P} ( \bN, \br, \gamma ) \subseteq \mathscr{P}(J)$, so
\begin{align*}
& \inf_{\cA}	\sup_{ ( \cP_1,\cdots, \cP_N ) \in \mathscr{P} ( J )  } \EE \bigg[
		\sum_{n=1}^{N} \bigg(F_{n} ( \btheta_n ) -  F_{n} ( \btheta_n^* ) \bigg) \bigg] \\[4pt]
&\ge 
\inf_{\cA}	\sup_{ ( \cP_1,\cdots, \cP_N ) \in \mathscr{P} ( \bN, \br , \gamma )  } \EE \bigg[
		\sum_{n=1}^{N} \bigg(F_{n} ( \btheta_n ) -  F_{n} ( \btheta_n^* ) \bigg) \bigg] \\[4pt]
&\ge 
C  \bigg[ 1 +  \sum_{ j=1 }^J  \bigg( 1 + \min \bigg\{ \frac{ d }{ B  } ,~ \frac{N-1}{J}  - 1  \bigg\} \bigg) \bigg] \\[4pt]
&\ge 
C \min\left\{ J \left( \frac{d}{B} + 1\right),~  N-1 \right\}
\ge 
\frac{C}{2} \min\left\{ J \left( \frac{d}{B} + 1\right),~  N \right\}.
\end{align*}

Below we prove \Cref{thm-lower-strongly-cvx-0}. For simplicity, we write $ \mathfrak{M}  (\gamma)$ as a shorthand for the worst-case risk over $\mathscr{P} ( \bN, \br , \gamma ) $. We will prove
\begin{align}
	& \mathfrak{M}  (\gamma)
	\gtrsim  1 +
	\sum_{ j=1 }^J  \min \bigg\{ \frac{ \gamma d }{ B  } ,~  N_{j} - N_{j-1}   - 1    \bigg\} 
	+  \sum_{j=1}^{J} r_j^2 , \qquad \forall \gamma \in [0, 1], 
	\label{eqn-thm-lower-strongly-cvx-1}
	\\
	& \mathfrak{M}  (0) \gtrsim  
	\sum_{ j=1 }^J  \min \bigg\{ \frac{d}{ B  } ,~ r_{j-1}^2  ( N_{j} - N_{j-1}  - 1   ) \bigg\} 
	.
	\label{eqn-thm-lower-strongly-cvx-2}
\end{align}
\Cref{thm-lower-strongly-cvx-0} immediately follows from the above estimates.

Choose any algorithm $\cA$ and denote by $\{ \btheta_n \}_{n=1}^N$ the output. For any fixed instance $( \cP_1 , \cdots, \cP_N ) \in \mathscr{P} ( \bN, \br , \gamma )$, we have $F_{n} ( \btheta_n ) - F_{n} ( \btheta_n^* ) = \EE_{\cA} \| \btheta_n - \btheta^*_n \|_2^2$. 
Here we write $\EE_{\cA}$ to emphasize the randomness over algorithm $\cA$'s output. 
For any probability distribution $\cQ$ over $\mathscr{P} ( \bN, \br , \gamma )$,
\begin{align}
	& \sup_{ ( \cP_1,\cdots, \cP_N ) \in \mathscr{P} ( \bN, \br, \gamma )  } 
	\EE_{\cA} \bigg[
	\sum_{n=1}^{N} \bigg(F_{n} ( \btheta_n ) -  F_{n} ( \btheta_n^* ) \bigg) \bigg] 
	\notag\\
	& = 
	\frac{1}{2} 
	\sup_{ ( \cP_1,\cdots, \cP_N ) \in \mathscr{P} ( \bN, \br , \gamma )  } 
	\bigg\{ 
	\sum_{n=1}^{N} \EE_{\cA} \| \btheta_n - \btheta^*_n \|_2^2
	\bigg\}
	\notag \\&
	\geq \frac{1}{2} \EE_{  ( \cP_1,\cdots, \cP_N ) \sim \cQ  } \bigg[
	\sum_{n=1}^{N} \EE_{\cA} \bigg( \| \btheta_n - \btheta^*_n \|_2^2
	\bigg| \cP_1,\cdots,\cP_{n-1}
	\bigg)
	\bigg] \notag \\
	& = \frac{1}{2} \sum_{n=1}^{N} 
	\underbrace{
		\EE_{  ( \cP_1,\cdots, \cP_N ) \sim \cQ  } \bigg[
		\EE_{\cA} \bigg( \| \btheta_n - \btheta^*_n \|_2^2
		\bigg| \cP_1,\cdots,\cP_{n-1}
		\bigg)
		\bigg]
	}_{ R(n) } \notag \\
	&= \frac{1}{2}  \sum_{ j=0 }^{J - 1} \sum_{ n= N_{j} + 2 }^{ N_{j+1} } R( n )
	+ \frac{1}{2}  \sum_{ j=0 }^J R ( N_j + 1 ) .
	\label{eqn-thm-lower-strongly-cvx-0}
\end{align}

\subsubsection{Proof of \eqref{eqn-thm-lower-strongly-cvx-1}}

For any $\bN \in \ZZ_+^J$ satisfying $N_1 <  \cdots < N_J = N - 1$ and $\br \in [0, 1]^J$, we generate $\{ \btheta^*_n \}_{n=1}^N$ by a Markov process:
\begin{itemize}
	\item Let $\btheta^*_0 = \bm{0}$, $r_0 = 1$, $N_0 = 0$ and $N_{J+1} = N$.
	\item For $j = 0, 1, \cdots, J$,
	\begin{itemize}
		\item Draw $\btheta^*_{N_j + 1}$ uniformly at random from $B \left( \btheta^*_{N_j} - \frac{r_j}{4\| \btheta^*_{N_j} \|_2} \btheta^*_{N_j}, \frac{r_j}{4} \right)$, with the convention $\bm{0} / 0 = \bm{0}$;
		\item If $N_{j+1} - N_{j} \geq 2$, let $r_j' = \min  \left\{
		\sqrt{  \frac{ 8 \gamma c^2 d}{B ( N_{j+1} - N_{j } ) } }  ,~ 1
		\right\}$ and draw $\{ \btheta^*_{n}  \}_{n = N_j + 2 }^{N_{j+1}}$ uniformly at random from $B \left( \btheta^*_{N_j + 1} - \frac{ r_j' }{4\| \btheta^*_{N_j + 1 } \|_2} \btheta^*_{N_j + 1 }, \frac{ r_j' }{ 4 } \right)$.
	\end{itemize}
\end{itemize}

The fact $r_j \in [0, 1]$ and \Cref{lem-ball} ensure that
\[
\btheta^*_{N_j + 1}  \in
B \bigg( \btheta^*_{N_j} - \frac{r_j}{4\| \btheta^*_{N_j} \|_2} \btheta^*_{N_j}, r_j /4 \bigg) \subseteq 
B (\bm{0}, 1/2) \cap  B ( \btheta^*_{N_j} , r_j /2 ) .
\]
Based on that, we use $r_j' \in [0, 1]$ and \Cref{lem-ball} to get
\[
\btheta^*_{n} \in
B \bigg( \btheta^*_{N_j + 1 } - \frac{ r_j' }{4\| \btheta^*_{N_j + 1 } \|_2} \btheta^*_{N_j + 1 }, r_j' /4 \bigg) \subseteq 
B (\bm{0}, 1/2) \cap  B ( \btheta^*_{N_j + 1} , r_j' /2 ) , \qquad N_j + 2 \leq n \leq N_{j+1} .
\]
Hence, the problem instance $( \cP_1, \cdots, \cP_N )$ induced by $\{ \btheta^*_n \}_{n=1}^N$ belongs to the class $\mathscr{P} ( \bN, \br, \gamma )$. The aforementioned procedure defines a probability distribution $\cQ $ over $\mathscr{P} ( \bN, \br, \gamma )$.

Choose any $j \in \{ 0, 1, \cdots, J \}$. Given $\{ \btheta^*_{n} \}_{n=0}^{N_j}$, $\btheta^*_{N_j + 1}$ is uniformly distributed in a ball with radius $r_j/4$. There exists a universal constant $c_1$ such that $R ( N_j + 1 ) \geq c_1 \cdot r_j^2$ holds regardless of the algorithm. Since $r_0 = 1$, we have
\begin{align}
\sum_{ j=0 }^J R ( N_j + 1 ) \gtrsim 1 + \sum_{ j=1 }^J r_j^2.
\label{eqn-proof-lower-boundary}
\end{align}

If $N_{j+1} \geq N_{j} + 2$. For any $n \in \{  N_{j} + 2 , N_j + 3 , \cdots, N_{j+1} \}$, the conditional uncertainty in $\btheta^*_{ n }$ given $\{ \btheta^*_i \}_{i=1}^{n-1}$ implies that $R(n) \geq c_1 \cdot r_j'^2$. Hence,
\begin{align}
\sum_{ n= N_{j} +2 }^{ N_{j+1}  } R( n ) \gtrsim ( N_{j+1} - N_j - 1 ) r_j'^2
\gtrsim \min  \bigg\{
\frac{ \gamma d}{B }  ,~  N_{j+1} - N_{j }  - 1 
\bigg\} 
.
\label{eqn-proof-lower-interior}
\end{align}
This bound trivially holds when $N_{j+1} = N_j + 1$. Combining the estimates  \eqref{eqn-thm-lower-strongly-cvx-0}, \eqref{eqn-proof-lower-boundary} and \eqref{eqn-proof-lower-interior} yields \eqref{eqn-thm-lower-strongly-cvx-1}.

\subsubsection{Proof of \eqref{eqn-thm-lower-strongly-cvx-2}}

For any $\bN \in \ZZ_+^J$ satisfying $N_1 <  \cdots < N_J = N - 1$ and $\br \in [0, 1]^J$, we generate $\{ \btheta^*_n \}_{n=1}^N$ by a Markov process:
\begin{itemize}
	\item Let $\btheta^*_0 = \bm{0}$, $r_0 = 1$, $N_0 = 0$ and $N_{J+1} = N$.
	\item For $j = 0, 1, \cdots, J$:
	\begin{itemize}
		\item Draw $\btheta^*_{N_j + 1}$ uniformly from $B ( \btheta^*_{N_j} - \frac{r_j}{4\| \btheta^*_{N_j} \|_2} \btheta^*_{N_j}, \frac{r_j}{4} )$, with the convention that $\bm{0} / 0 = \bm{0}$;
		\item If $N_{j+1} - N_{j} \geq 2$, set $\btheta^*_{N_j + 1} = \btheta^*_{N_j + 2} = \cdots = \btheta^*_{N_{j+1} }$.
	\end{itemize}
\end{itemize}
The fact $\br \in [0, 1]^J$ and \Cref{lem-ball} ensure that
\[
B \bigg( \btheta^*_{N_j} - \frac{r_j}{4\| \btheta^*_{N_j} \|_2} \btheta^*_{N_j}, r_j /4 \bigg) \subseteq 
B (\bm{0}, 1/2) \cap  B ( \btheta^*_{N_j} , r_j /2 ) .
\]
Hence, the problem instance $( \cP_1, \cdots, \cP_N )$ induced by $\{ \btheta^*_n \}_{n=1}^N$ belongs to the class $\mathscr{P} ( \bN, \br )$. 

Now, we choose any $j \in \{ 0, 1, \cdots , J - 1 \}$ with $N_{j+1} \geq N_j + 2$, and study the error $\sum_{ n= N_{j} + 2 }^{ N_{j+1}  } R( n )$. 
By construction, we have $\btheta^*_{ N_{j} + 1 } = 
\btheta^*_{ N_{j} + 2 } = \cdots = \btheta^*_{ N_{j+1}  } $. Hence, $\{ \cD_i \}_{ i = N_j + 1 }^{ N_{j+1} }$ are i.i.d., each of which consists of $B$ samples from $\btheta^*_{ N_{j} + 1 }$. For each $n \in \{  N_{j} + 2 , N_j + 3 , \cdots, N_{j+1} \}$, the algorithm $\cA$ examines $\{ \cD_i \}_{i=1}^{n-1}$ and predicts $\btheta^*_{ N_{j} + 1 }$.

Imagine an oracle algorithm $\cB$ that wants to predict $\btheta^*_{ N_{j} + 1 }$ based on data $\{ \cD_i \}_{i=1}^{N_{j+1} - 1}$ and true values of $\{ \btheta^*_{i} \}_{i=1}^{N_j}$. Denote by $\widehat\btheta^{\cB} $ the output. Thanks to our Markovian construction, the past data $\{ \cD_i \}_{i = 1 }^{N_j}$ are independent of $\btheta^*_{ N_{j} + 1 }$ given $\{ \btheta^*_{i} \}_{i=1}^{N_j}$. Then, $\bar\btheta^{\cB} = \EE ( \widehat\btheta^{\cB} |  \{ \btheta^*_{i} \}_{i=1}^{N_j} ) $ is an estimator of $\btheta^*_{ N_{j} + 1 }$ that only depends on $\{ \cD_i \}_{i=N_{j} + 1}^{N_{j+1} - 1}$ and $\{ \btheta^*_{i} \}_{i=1}^{N_j} $. In addition, the Rao-Blackwell theorem implies that
\begin{align}
	\EE \| \bar\btheta^{\cB} - \btheta^*_{ N_{j} + 1 } \|_2^2
	\leq 
	\EE \| \widehat\btheta^{\cB} - \btheta^*_{ N_{j} + 1 } \|_2^2 .
	\label{eqn-Rao-Blackwell}
\end{align}
Note that $\{ \cD_i \}_{i=N_{j} + 1}^{N_{j+1} - 1}$ consists of $B (N_{j+1} - N_j - 1)$ i.i.d.~samples from $N( \btheta^*_{ N_{j} + 1 } , \bI_d )$. Also, conditioned on $\{ \btheta^*_{n} \}_{n=0}^{N_j}$, $\btheta^*_{N_j + 1}$ is uniformly distributed in a ball with radius $r_j/4$. Using standard tools \citep{Tsy09}, we can prove a lower bound
\begin{align}
	\EE \| \bar\btheta^{\cB} - \btheta^*_{ N_{j} + 1 } \|_2^2
	\geq c_2 \min \bigg\{ r_j^2, \frac{d}{ B (N_{j+1} - N_j - 1) } \bigg\} 
	\label{eqn-thm-lower-strongly-cvx}
\end{align}
with $c_2$ being is a universal constant. This lower bound holds for every oracle algorithm $\cB$ due to \eqref{eqn-Rao-Blackwell}.

At each time $n \in \{  N_{j} + 2 , N_j + 3 , \cdots, N_{j+1} \}$, $\cA$ wants to achieve the same goal as an oracle algorithm $\cB$ with less information. Consequently, the lower bound \eqref{eqn-thm-lower-strongly-cvx} holds for $\cA$. We have
\begin{align}
	& \sum_{ n= N_{j} +2 }^{ N_{j+1}  } R( n ) \gtrsim \min \bigg\{ (N_{j+1} - N_j - 1  ) r_j^2,~ \frac{d}{ B  } \bigg\} , \qquad 0 \leq J \leq J - 1 , \notag \\
	& \sum_{ j=0 }^{J - 1}
	\sum_{ n= N_{j} +2 }^{ N_{j+1}  } R( n ) 
	\gtrsim \sum_{ j=1 }^J  \min \bigg\{ (N_{j} - N_{j-1} - 1   ) r_{j-1}^2,~ \frac{d}{ B  } \bigg\} .
	\label{eqn-proof-lower-interior-2}
\end{align}
The last inequality trivially holds when $N_{j+1} - N_j = 1$.  Combining \eqref{eqn-thm-lower-strongly-cvx-0} and \eqref{eqn-proof-lower-interior-2} yields \eqref{eqn-thm-lower-strongly-cvx-2}.

\subsection{Proof of \Cref{cor-lower-strongly-cvx}}\label{sec-cor-lower-strongly-cvx-proof}

It suffices to prove $\cL \asymp \cU$, where
\begin{align*}
	& \cL = \sup_{ ( \cP_1,\cdots, \cP_N ) \in \mathscr{Q} (V) } \EE \bigg[
	\sum_{n=1}^{N} \bigg(F_{n} ( \btheta_n ) - F_{n} ( \btheta_n^* ) \bigg) \bigg] , \\
	& \cU = 1 +
	\frac{d }{ B}  + 
	N^{1/3} \bigg( \frac{V d}{B} \bigg)^{2/3} .
\end{align*}

Let $\mathscr{P} ( \bN, \br , \gamma ) $ be defined as in \Cref{thm-lower-strongly-cvx-0}. Whenever $\br \in [0, 1]^J$ and $\sum_{ j=1 }^J r_j \leq V$ hold, we have $\mathscr{P} ( \bN, \br , 0 ) \subseteq \mathscr{Q} (V) $ and thus \Cref{thm-lower-strongly-cvx-0} forces
\begin{align}
	&\cL \geq 
	\sup_{ ( \cP_1,\cdots, \cP_N ) \in \mathscr{P} ( \bN, \br , 0 )  } \EE \bigg[
	\sum_{n=1}^{N} \bigg(F_{n} ( \btheta_n ) -  F_{n} ( \btheta_n^* ) \bigg) \bigg] 
	\notag \\ &
	\gtrsim  1 + \sum_{ j=0 }^J  r_j^2 + \sum_{ j=1 }^J
	\min \bigg\{ \frac{d}{ B  } ,~ r_{j-1}^2 ( N_{j} - N_{j-1}  - 1  ) \bigg\} 
	\notag \\ &
	\geq 1 +  \sum_{ j=1 }^J 
	\min \bigg\{ \frac{d}{ B  } ,~ r_{j-1}^2 ( N_{j} - N_{j-1}  ) \bigg\} 
	,
	\label{thm-lower-strongly-cvx-PV}
\end{align}
where $r_0 = 1$. We will choose appropriate $J, \bN$ and $\br$ to make this lower bound have the desired form.
First of all, let $J = \min   \{ 
N - 1,~ \lfloor ( BNV^2 / d )^{1/3} \rfloor  + 1 
\}$. The assumption $V \leq  N \sqrt{ d / B }$ yields $ ( BNV^2 / d )^{1/3} \leq  N$ and
\begin{align}
	J  \asymp  ( BNV^2 / d )^{1/3} .
	\label{thm-lower-strongly-cvx-PV-1}
\end{align}


If $V < \sqrt{ \frac{d}{ B N } } $, then $J = 1$. Take $N_1 = N - 1$ and $r_1 = 0$. From \eqref{thm-lower-strongly-cvx-PV} and $N \geq d / B$ we get
\[
\cL \gtrsim 
1 + \min  \{ d/B ,~  N  \} = 1 + \frac{d}{B}.
\]
Since
\[
N^{1/3} \bigg( \frac{V d}{B} \bigg)^{2/3}  \leq N^{1/3} \bigg( \frac{d}{B} \bigg)^{2/3} \bigg(
\sqrt{ \frac{d}{ B N } }
\bigg)^{2/3}   = \frac{d}{B} ,
\]
we have $\cU \asymp 1 + d / B$ and $\cL \asymp \cU$.

Now, suppose that $V \geq \sqrt{ \frac{d}{ B N } }$, which implies $2 \leq J \leq N - 1$. Define $Q = \lfloor \frac{ N - 1 }{ J - 1 }  \rfloor$. Let $N_j = j Q$ for $j \in [J - 1]$ and $N_J = N - 1$. By \eqref{thm-lower-strongly-cvx-PV} and $r_0 = 1$,
\begin{align}
	\cL
	& \gtrsim 1   + 
	\min \bigg\{ \frac{d}{ B  } ,~ Q \bigg\} 
	+
	\sum_{ j=2 }^{J - 1}
	\min \bigg\{ \frac{d}{ B  } ,~ r_{j-1}^2 Q \bigg\} 
	.
	\label{thm-lower-strongly-cvx-PV-2}
\end{align}
We now choose the $r_j$'s. By \eqref{thm-lower-strongly-cvx-PV-1}, there exists a constant $C_1$ such that $J \geq C_1 ( BNV^2 / d )^{1/3} $. Using the assumption $V \leq N  B / d $, we get
\[
\frac{V}{ J } \leq \frac{ V }{C_1   ( BNV^2 / d )^{1/3} } = C_1^{-1} \bigg( \frac{ V d }{ B N } \bigg)^{1/3} \leq C_1^{-1} .
\]
Define $r_j = \frac{C_1 V}{J}$ for $j \in [J]$. We have $\br \in [0, 1]^J$. Then, \eqref{thm-lower-strongly-cvx-PV-2} leads to
\begin{align*}
	\cL
	& \gtrsim 1 + 
	\min \bigg\{ \frac{d}{ B  } ,~ Q \bigg\} 
	+
	(J - 2)
	\min \bigg\{ \frac{d}{ B  } ,~ \bigg( \frac{V}{J} \bigg)^2 Q \bigg\} .
\end{align*}
If $J = 2$, then $Q \asymp N$ and $\cL \gtrsim 1 +  \min  \{ \frac{d}{ B  } ,~ N  \} = 1 + \frac{d}{B}$. Similar to the $J = 1$ case, we can derive that $\cL \asymp \cU$. If $J \geq 3$, then $\cL \gtrsim 1 +  J
\min  \{ \frac{d}{ B  } ,~  ( \frac{V}{J}  )^2 Q  \}
$. By \eqref{thm-lower-strongly-cvx-PV-1},
\begin{align*}
	&\bigg( \frac{V}{J } \bigg)^2 Q 
	=   \frac{V^2}{J^3 }\cdot JQ \asymp  \frac{V^2 N }{J^3 }  
	\gtrsim \frac{ V^2 N }{
		BNV^2 / d  
	} 
	= \frac{d}{B} , \\
	& \cL
	\gtrsim 1 +   \frac{ J d}{ B  } 
	\asymp 1 +   \bigg( \frac{BNV^2}{d} \bigg)^{1/3} 
	\frac{ d}{ B  }   = 1 + N^{1/3} \bigg( \frac{ V d  }{B } \bigg)^{ 2/3 } .
\end{align*}
The above bound shows that $ \frac{ J d}{ B  } \asymp N^{1/3}  (   V d  / B  )^{ 2/3 }$. Hence, $d / B \lesssim N^{1/3}  (   V d  / B  )^{ 2/3 }$ and $\cU \asymp \cL$.

\subsection{Proof of \Cref{thm-lower-Lip}}\label{sec-thm-lower-Lip-proof}

We prove the following stronger version of \Cref{thm-lower-Lip}.

\begin{theorem}\label{thm-lower-Lip-0}
Choose any integer $N \geq 2$, $J \in [N - 1]$ and $\bN \in \ZZ_+^J$ satisfying $N_1 <  \cdots < N_J = N - 1$ and $\br \in [0, 1]^J$. Define $N_0 = 0$, $\bmu^*_0 = \bm{0}$ and
\begin{align*}
\mathscr{P} ( \bN, \br ) & = \bigg\{ ( \cP_1 , \cdots, \cP_N ) :~ \cP_n = \cP (\bmu_n^*) \text{ and }  \bmu_n^* \in B_{\infty} (\bm{0}, 1/2)  ,~\forall n \in [N],\\
& 
\frac{1}{d} \sum_{n = N_{j-1} + 1 }^{ N_j - 1 }
\| \bmu^*_{n+1} - \bmu^*_n \|_1
\leq 
 \sqrt{  \frac{ d}{B ( N_j - N_{j - 1} ) } }  
,~\frac{1}{d} \|  \bmu^*_{N_{j} + 1 } -  \bmu^*_{N_{j} } \|_1 \leq r_j ,~
\forall j \in [J] 
\bigg\} .
\end{align*}
There is a universal constant $C>0$ such that
	\begin{align*}
	& 	\inf_{\cA}  \sup_{ ( \cP_1,\cdots, \cP_N ) \in \mathscr{P} ( \bN , \br )  } 
	\EE \bigg[
	\sum_{n=1}^{N} \left(F_{n} ( \btheta_n ) - \inf_{ \btheta_n' \in \Omega } F_{n} ( \btheta_n' ) \right) \bigg] 
		\\ & 
	\geq C
	\bigg(
	1 +
	\sum_{ j=1 }^J
	 \min \bigg\{ \sqrt{ \frac{d  ( N_j - N_{j-1}  )  }{ B } } ,~
	(N_j - N_{j-1}- 2)_+ \bigg\}   
	+ \sum_{ j=1 }^J r_j
	\bigg) .
	\end{align*}
The infimum is taken over all online algorithms $\cA$ for Problem \ref{problem-online}, and $\{\btheta_n\}_{n=1}^N$ is the output of $\cA$. 
\end{theorem}

To see that \Cref{thm-lower-Lip} is a special case of \Cref{thm-lower-Lip-0}, fix $J\in[N-1]$ that divides $N-1$. For $j\in[J]$, let $N_j = j(N-1)/J$ and $r_j = 1$. Then $\mathscr{P} ( \bN , \br ) \subseteq \mathscr{P}(J)$, so
\begin{align*}
& \inf_{\cA}  \sup_{ ( \cP_1,\cdots, \cP_N ) \in \mathscr{P} ( J )  } 
\EE \left[ \sum_{n=1}^{N} \left(F_{n} ( \btheta_n ) - \inf_{ \btheta_n' \in \Omega } F_{n} ( \btheta_n' ) \right) \right] \\[4pt]
&\ge 
\inf_{\cA}  \sup_{ ( \cP_1,\cdots, \cP_N ) \in \mathscr{P} ( \bN , \br )  } 
\EE \left[ \sum_{n=1}^{N} \left(F_{n} ( \btheta_n ) - \inf_{ \btheta_n' \in \Omega } F_{n} ( \btheta_n' ) \right) \right] \\[4pt]
&\ge 
C\bigg(1 + \sum_{ j=1 }^J \min \bigg\{ \sqrt{ \frac{d }{ B } \frac{N-1}{J} },~ \left(\frac{N-1}{J}- 2 \right)_+ \bigg\}  + J \bigg) \\[4pt]
&\ge 
\frac{C}{2} \min \left\{ J + \sqrt{\frac{J(N-1)d}{B}},~ N-1 \right\} \\[4pt]
&\ge 
\frac{C}{4} \min \left\{ J + \sqrt{\frac{JNd}{B}},~ N \right\}.
\end{align*}

We now prove \Cref{thm-lower-Lip-0}. Let
\[
\mathfrak{M}_P ( \bN, \br ) = \inf_{\cA}  \sup_{ ( \cP_1,\cdots, \cP_N ) \in \mathscr{P} ( \bN , \br )  } 
\EE \left[ \sum_{n=1}^{N} \left(F_{n} ( \btheta_n ) - \inf_{ \btheta_n' \in \Omega } F_{n} ( \btheta_n' ) \right) \right].
\] 
$\mathfrak{M}_P ( \bN, \br )$ and $\mathfrak{M}_Q ( V )$ the quantities on the left-hand sides of the minimax lower bounds.

Since $\btheta_1$ is agnostic to $\bmu^*$, we have $\mathfrak{M}_P ( \bN , \br) \gtrsim 1$. 
Below we prove two separate bounds
\begin{align}
& \mathfrak{M}_P ( \bN, \br ) \gtrsim \sum_{ j=1 }^J r_j , \label{eqn-proof-lower-2} \\
& \mathfrak{M}_P ( \bN, \br ) \gtrsim 
\sum_{ j=1 }^J
\min \bigg\{ \sqrt{ \frac{d ( N_j - N_{j-1} - 1 ) }{ B } } ,
N_j - N_{j-1} - 1 \bigg\} . \label{eqn-proof-lower-1} 
\end{align}

We start from \eqref{eqn-proof-lower-2}. Define a sub-class of $\mathscr{P} (\bN, \br)$:
\begin{align*}
& \mathscr{P}_{1} (\bN, \br) =  \bigg\{ ( \cP_1 , \cdots, \cP_N ) :~ \cP_n = \cP (\bmu_n^*) ,~
\| \bmu^*_{N_{j} + 1} \|_1 / d \leq r_j \text{ if }  j \text{ is odd},~ \bmu^*_n = \bm{0} \text{ for other } n
\bigg\}  .
\end{align*}
Denote by $ \mathfrak{M}_{P_1} ( \bN, \br ) $ the minimax lower bound over $\mathscr{P}_1$. Then, it is easily seen that
\[
 \mathfrak{M}_{P_1} ( \bN, \br ) \gtrsim \sum_{ j \text{ is odd } } r_j .
\]
We can define another sub-class of $\mathscr{P} (\bN, \br)$ that involves even $j$'s and leads to a similar lower bound. The inequality \eqref{eqn-proof-lower-2} follows by combining them together.

It remains to prove \eqref{eqn-proof-lower-1}. Choose any algorithm $\cA$ and denote by $\{ \btheta_n \}_{n=1}^N$ the output. 
For any probability distribution $\cQ$ over $\mathscr{P} ( \bN , \br )$, we have
\begin{align}
	& \mathfrak{M}_P ( \bN , \br )
	\geq \sum_{n=1}^{N} 
	\underbrace{
		\EE_{  ( \cP_1,\cdots, \cP_N ) \sim \cQ  } \bigg[
		\EE_{\cA} \bigg( F_{n} ( \btheta_n ) - \inf_{ \btheta \in \Omega } F_{n} ( \btheta )
		\bigg| \cP_1,\cdots,\cP_{n-1}
		\bigg)
		\bigg]
	}_{ R(n) } 
 =  \sum_{ j=0 }^{J - 1} \sum_{ n= N_{j} + 1 }^{ N_{j+1} } R( n ) .
	\label{eqn-thm-lower-Lip-0}
\end{align}
We now invoke a useful lemma, which directly follows from the proof of Theorem 1 in \cite{AWB09}.
\begin{lemma}\label{lem-lower-Lip}
	Suppose there is an algorithm that analyzes $n$ samples from any unknown distribution $\cP ( \bmu )$ and returns $\widehat\btheta$ as an estimated minimizer of $F_{\bmu}$. 
	There exists a universal constant $C>0$ and a random vector $\bnu$ distributed in $\{ \pm 1 \}^d$ such that
	\begin{align*}
		\EE  \bigg[ F_{ r \bnu } (\widehat\btheta) - 
		\inf_{ \btheta \in \Omega } F_{ r \bnu } (\btheta) 
		\bigg] \geq C \min \bigg\{ r ,~ \sqrt{ \frac{d}{n} } \bigg\} , \qquad \forall r \in [0, 1/2].
	\end{align*}
	Here the expectation is taken over the randomness of both $\bnu$ and $\widehat\btheta$.
\end{lemma}

Let $N_0 = 0$ and $N_{J+1} = N$. We draw i.i.d.~copies $\{ \bnu^*_{j} \}_{j=1}^J$ of the random vector $\bnu$ in \Cref{lem-lower-Lip}, and generate $\{ \bmu^*_n \}_{n=1}^N$ through the following procedure:  For $j = 0, 1, \cdots, J-1$,
\begin{itemize}
	\item Let $\bmu^*_{N_j } = \bm{0}$;
	\item If $N_{j+1} - N_j = 2$, let $\bmu^*_{N_j + 1} = \bm{0}$;
	\item If $N_{j+1} - N_{j} \geq 3$, let $\bmu^*_{N_j + 1} =  \bm{0}$ and $\bmu_{N_j+2}^* = \cdots = \bmu^*_{N_{j+1} - 1 } = \frac{1}{2} \min \{ \sqrt{ \frac{d}{ B ( N_j - N_{j-1} - 1) } } , 1 \} \bnu^*_j$.
\end{itemize}
The problem instance $( \cP_1, \cdots, \cP_N )$ induced by $\{ \bmu^*_n \}_{n=1}^N$ clearly belongs to the class $\mathscr{P} ( \bN  , \br)$. Hence, we get a probability distribution $\cQ $ over $\mathscr{P} ( \bN , \br )$. 

Choose any $j \in \{ 0, 1, \cdots, J - 1 \}$ with $N_{j + 1} \geq N_j + 3$. For any $n \in \{ N_{j} + 3 , \cdots, N_{j+1}  \}$, \Cref{lem-lower-Lip} implies that
\[
R(n) \gtrsim \min \bigg\{
\min \bigg\{ \sqrt{ \frac{d}{ B ( N_{j+1} - N_{j} - 2 ) } } , 1 \bigg\}
,
\sqrt{ \frac{d}{B ( n - N_{j}  )  } }
\bigg\}
= \min \bigg\{ \sqrt{ \frac{d}{ B ( N_{j+1} - N_{j} - 2 ) } } , 1 \bigg\} .
\]
Hence,
\begin{align*}
	\sum_{n = N_j + 1}^{N_{j+1} } R(n) \ge \sum_{n = N_j + 3}^{N_{j+1} } R(n) 
	&\gtrsim 
	\min \bigg\{ \sqrt{ \frac{d ( N_{j+1} - N_{j} - 2 ) }{ B } } ,~ N_{j+1} - N_{j} - 2 \bigg\} \\
	&\gtrsim
	\min \bigg\{ \sqrt{ \frac{d ( N_{j+1} - N_{j}) }{ B } } ,~ (N_{j+1} - N_{j} - 2)_+ \bigg\}.
\end{align*}
The above bound trivially holds when $N_{j + 1} \in \{ N_j + 1, N_j + 2 \} $. Plugging it into \eqref{eqn-thm-lower-Lip-0} yields \eqref{eqn-proof-lower-1}.

\subsection{Proof of \Cref{cor-lower-Lip}}\label{sec-cor-lower-Lip-proof}

Let
\[
\mathfrak{M}_Q (V) = \inf_{\cA} \sup_{ ( \cP_1,\cdots, \cP_N ) \in \mathscr{Q} ( V )  } 
	\EE \left[
	\sum_{n=1}^{N} \left(F_{n} ( \btheta_n ) - \inf_{ \btheta_n' \in \Omega } F_{n} ( \btheta_n' ) \right) \right] 
\]
and $\cL = 1 +  \sqrt{ d N / B } + N^{2/3}  ( V d / B  )^{1/3}$. \Cref{lem-lower-Lip} and the assumption $N \geq d / B$ yield $\mathfrak{M}_Q ( 0 )
\gtrsim \sqrt{ \frac{d N }{ B } } $. Moreover, since $\btheta_1$ is agnostic to $\bmu^*$, we have $
\mathfrak{M}_Q ( 0 )
\gtrsim
1 $. The above results yields
\[
\mathfrak{M}_Q (V) \geq \mathfrak{M}_Q (0)  \gtrsim
1 +  \sqrt{ \frac{d N }{ B } } .
\]
When $V \leq \sqrt{ \frac{8d}{BN} } $, we have $N^{2/3} ( V d / B )^{1/3} \lesssim \sqrt{N d / B}$ and thus $\mathfrak{M}_Q (V) \geq \mathfrak{M}_Q (0)  \gtrsim \cL$.

When $\sqrt{ \frac{8 d}{BN} } \leq V \leq N \min \{ \sqrt{d / B} , B / d \} /6$, we let $J =  \lfloor  V^{2/3} ( B N / d  )^{1/3}\rfloor$. Then $2\le J\le N/3\le N-1$.
Define $Q =  \lceil N/J \rceil \ge 3$ and
\begin{align*}
	\mathscr{R}  & = \bigg\{ ( \cP_1 , \cdots, \cP_N ) :~ \cP_n = \cP (\bmu_n^*) \text{ and }  \bmu_n^* \in B_{\infty} (\bm{0}, 1/2)  ,~\forall n \in [N] ; \\
	& \qquad 
	\sum_{n = (j-1) Q +1 }^{ j Q }
	\| \bmu^*_{n+1} - \bmu^*_n \|_1
	\leq 
	d \sqrt{  \frac{ d}{B Q} }  
	,~
	\forall j \in [J - 1] ;~~
	\bmu^*_{ (J-1) Q + 1} = \cdots = \bmu^*_N = \bm{0}
	\bigg\} .
\end{align*}
It can be readily checked that $\mathscr{R} \subseteq \mathscr{Q} (V)$. By our lower bound over $\mathscr{P}(\bN,\br)$ with $N_j=jQ$ and $r_j=1$ for $j\in[J-1]$,
\begin{align*}
	\mathfrak{M}_Q (V) 
	&\geq 
	\sup_{ ( \cP_1,\cdots, \cP_N ) \in \mathscr{R}  } 
	\EE \bigg[
	\sum_{n=1}^{N} \left(F_{n} ( \btheta_n ) - \inf_{ \btheta_n' \in \Omega } F_{n} ( \btheta_n' ) \right) \bigg] 
	\gtrsim
	1 +
	(J - 1) \min \bigg\{ \sqrt{ \frac{d Q }{ B } } ,
	Q - 2 \bigg\} \\
	& \asymp 1 +  \min \bigg\{ \sqrt{ \frac{d N J }{ B } } ,
	N \bigg\} 
	= 1 + N \min \bigg\{ \sqrt{ \frac{d J }{ N B } } ,
	1 \bigg\} .
\end{align*}
Since $J \asymp V^{2/3} ( B N / d  )^{1/3}$ and $V \leq N B / d$,
\[
\sqrt{ \frac{d J }{ N B } } \asymp \bigg( \frac{ V d}{N B} \bigg)^{1/3} \leq 1.
\]
As a result,
\begin{align*}
	\mathfrak{M}_Q (V) 
	&\gtrsim  1 + N \bigg( \frac{ V d}{N B} \bigg)^{1/3} = 1 + N^{2/3} \bigg( \frac{ V d}{ B} \bigg)^{1/3} .
\end{align*}
Finally, note that our assumption $V \geq \sqrt{ \frac{8 d}{NB} }$ implies
\[
N^{2/3} \bigg( \frac{ V d}{ B} \bigg)^{1/3} = N  \bigg( \frac{ V d}{ N B} \bigg)^{1/3} 
\gtrsim N  \bigg(  \sqrt{ \frac{ d}{NB} } \cdot \frac{ d}{ N B} \bigg)^{1/3} =\sqrt{ \frac{d N}{B}}.
\]
Hence, $\mathfrak{M}_Q (V) \gtrsim \cL$.

%% file: appendix_technical.tex
\section{Technical Lemmas}

\begin{lemma}\label{lem-ball}
If $\btheta \in B(\bm{0}, 1 )$ and $r \leq 1$, then
\[
B \bigg( \btheta - \frac{r}{2 \| \btheta \|_2 } \btheta , \frac{r}{2} \bigg) \subseteq B ( \bm{0} , 1 ) \cap B (\btheta, r).
\]
Here we adopt the convention that $\bm{0} / 0 = \bm{0}$.
\end{lemma}

\begin{proof}[\bf Proof of \Cref{lem-ball}]
The result is trivial when $\btheta = \bm{0}$. Now, suppose that $\btheta \neq \bm{0}$ and let $\bar\btheta = \btheta - \frac{r}{2 \| \btheta \|_2 } \btheta$. We have $\| \bar\btheta - \btheta \|_2 = r/2$. Hence, $ B(\bar\btheta , r/2) \subseteq B (\btheta, r)$. It remains to show that $B(\bar\btheta , r/2)  \subseteq B(\bm{0}, 1)$, which is equivalent to $\| \bar\btheta \|_2 + r/2 \leq 1$.
\begin{itemize}
\item If $0 < \| \btheta \|_2 \leq r/2$, then $\| \bar\btheta \|_2 = r/2 - \| \btheta \|_2$ and thus $\| \bar\btheta \|_2 + r/2 \leq r - \| \btheta \|_2 \leq r \leq 1$.
\item If $\| \btheta \|_2 > r/2$, then $\| \bar\btheta \|_2 = \| \btheta \|_2 - r/2$ and thus $\| \bar\btheta \|_2 + r/2 \leq  \| \btheta \|_2 \leq r \leq 1$.
\end{itemize} 
This finishes the proof. 
\end{proof}

\begin{lemma}\label{lem-subg-norm}
Let $\{ \bv_i \}_{i=1}^n \subseteq \RR^d$ be independent random vectors with $\EE \bv_i = \bm{0}$ and $\| \bv_i \|_{\psi_1} \leq \sigma$, $\forall i \in [n]$. There exists a universal constant $C > 0$ such that
\[
\PP \bigg[
\bigg\| \frac{1}{n} \sum_{i=1}^{n} \bv_i \bigg\|_2 \geq \sigma s
\bigg]
\leq \exp (  d \log 5  - C n \min \{ s^2, s \} ) , \qquad \forall t \geq 0.
\]
\end{lemma}

\begin{proof}[\bf Proof of \Cref{lem-subg-norm}]
Let $\overline{\bv}=(1/n)\sum_{i=1}^n\bv_i$. There exists a $1/2$-net $\mathcal{N}$ of $\mathbb{S}^{d-1}$ such that $\mathcal{N}\subset\mathbb{S}^{d-1}$ and $|\mathcal{N}|\le 5^d$ (Lemma 5.2 in \cite{Ver10}). For every $\bu\in\mathbb{S}^{d-1}$, there exists $\pi(\bu)\in\mathcal{N}$ such that $\|\bu-\pi(\bu)\|_2\le 1/2$, so
\[
\|\overline{\bv}\|_2
=
\max_{\bu\in\mathbb{S}^{d-1}}\langle\overline{\bv},\bu\rangle
=
\max_{\bu\in\mathbb{S}^{d-1}}(\langle\overline{\bv},\bu-\pi(\bu)\rangle+\langle\overline{\bv},\pi(\bu)\rangle)
\le 
\frac{1}{2}\|\overline{\bv}\|+\max_{\bu\in\mathcal{N}}\langle\overline{\bv},\bu\rangle,
\]
which implies $\|\overline{\bv}\|_2\le 2\max_{\bu\in\mathcal{N}}\langle\overline{\bv},\bu\rangle$. Then for every $s\ge 0$,
\[
\PP\left(\|\overline{\bv}\| \ge \sigma s\right)
\le 
\PP\left(\max_{\bu\in\mathcal{N}}\langle\overline{\bv},\bu\rangle \ge \frac{\sigma s}{2}\right)
\le 
\sum_{\bu\in\mathcal{N}}\PP\left(\frac{1}{n}\sum_{i=1}^n\langle\bv_i,\bu\rangle \ge \frac{\sigma s}{2}\right).
\]
Since $\|\bv_i\|_{\psi_1}\le K$, then by a Bernstein-type inequality (Proposition 5.16 in \cite{Ver10}), there exists an absolute constant $C>0$ such that for every $\bu\in\mathcal{N}$,
\[
\PP\left(\frac{1}{n}\sum_{i=1}^n\langle\bv_i,\bu\rangle \ge \frac{\sigma s}{2}\right)
\le 
\exp\left(-Cn\min\{s,s^2\}\right),\quad\forall\,s\ge 0.
\]
Thus, for all $s\ge 0$,
\[
\PP\left(\|\overline{\bv}\| \ge \sigma s\right)
\le 
5^d\exp\left(-Cn\min\{s,s^2\}\right)
=
\exp\left(d\log 5-Cn\min\{s,s^2\}\right).
\]
This completes the proof.
\end{proof}

%% file: appendix_experiments.tex
\section{Non-Stationarity Patterns in Real-Data Experiments}\label{sec-patterns}

\begin{figure}[h]
    \centering
     \includegraphics[scale=0.52]{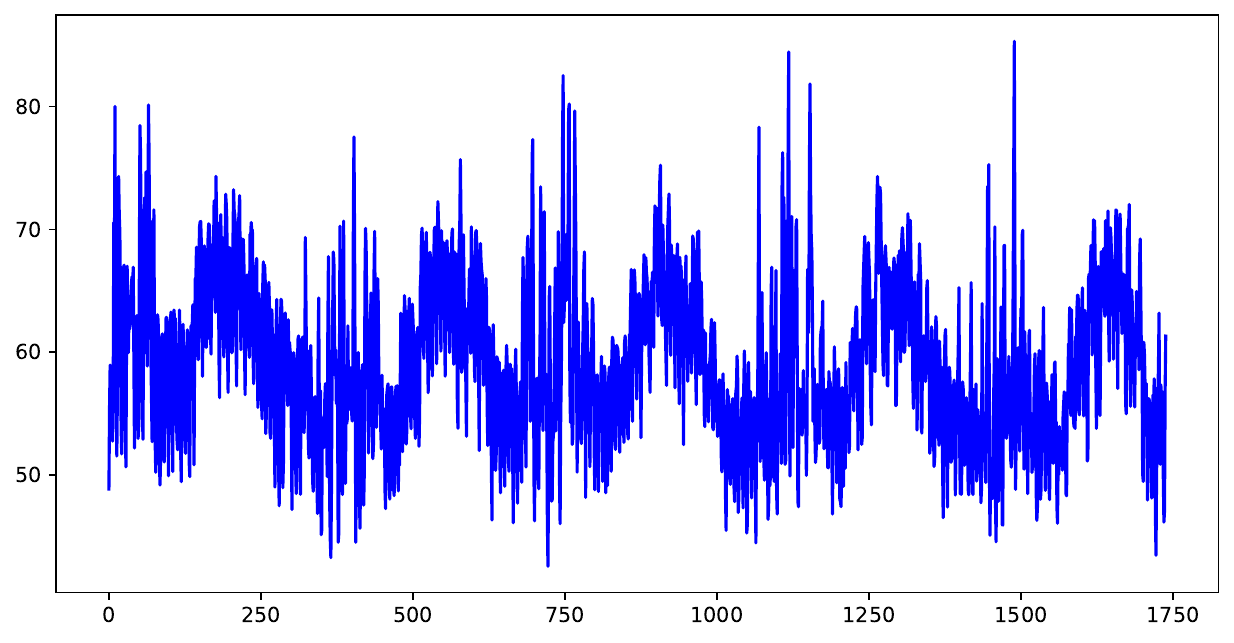}
	\caption{Daily electricity demand in Victoria, Australia from January 1st, 2016 to October 6th, 2020. Horizontal axis: time period $n$. Vertical axis: electricity demand $y_n$ (unit: megawatt-hour), scaled by $5\times 10^{-4}$. \label{fig-electricity-pattern}}
\end{figure}

\begin{figure}[h]
    \centering
    \includegraphics[scale=0.52]{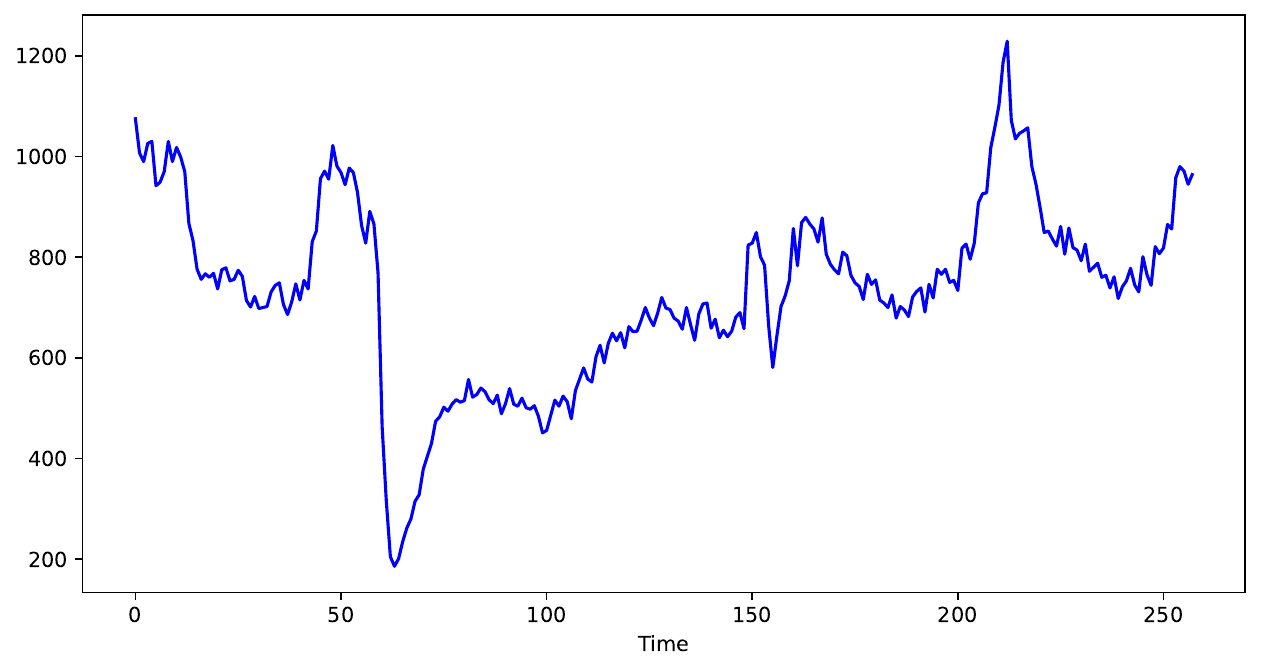}
	\caption{Weekly emergency department (ED) visit counts for vomiting, from January 7th, 2019 to December 31st, 2023. Horizontal axis: time period $n$. Vertical axis: ED visit counts. \label{fig-ED-pattern}}
\end{figure}

In this section, we provide plots to visualize the non-stationarity patterns in the real data experiments. For the electricity demand prediction problem in \Cref{sec-experiments-electricity}, \Cref{fig-electricity-pattern} plots the electricity demand from January 1st, 2016 to October 6th, 2020. For the nurse staffing problem in \Cref{sec-experiments-hospital}, \Cref{fig-ED-pattern} plots the weekly ED visit counts for vomiting from January 7th, 2019 to December 31st, 2023.

